\documentclass[lettersize,journal, compsoc]{IEEEtran}
\usepackage{amsmath,amsfonts}
\usepackage{algorithmic}
\usepackage{array}
\usepackage[caption=false,font=normalsize,labelfont=sf,textfont=sf]{subfig}
\usepackage{textcomp}
\usepackage{stfloats}
\usepackage{url}
\usepackage{verbatim}
\usepackage{graphicx}
\usepackage{hyperref}
\usepackage{url}
\usepackage{enumitem}
\usepackage{natbib}
\usepackage{booktabs}       %
\usepackage{adjustbox}
\usepackage{xspace}
\usepackage{placeins}
\usepackage{afterpage}
\usepackage{multirow}
\usepackage{colortbl}
\usepackage{xcolor}
\usepackage{pifont}
\usepackage{algorithm}
\def\BibTeX{{\rm B\kern-.05em{\sc i\kern-.025em b}\kern-.08em
    T\kern-.1667em\lower.7ex\hbox{E}\kern-.125emX}}
\usepackage{balance}
\begin{document}

\newcommand{\Perp}{\perp\!\!\! \perp}
\newcommand{\bK}{\mathbf{K}}
\newcommand{\bX}{\mathbf{X}}
\newcommand{\bY}{\mathbf{Y}}
\newcommand{\bk}{\mathbf{k}}
\newcommand{\bx}{\mathbf{x}}
\newcommand{\by}{\mathbf{y}}
\newcommand{\bhy}{\hat{\mathbf{y}}}
\newcommand{\bty}{\tilde{\mathbf{y}}}
\newcommand{\bG}{\mathbf{G}}
\newcommand{\bI}{\mathbf{I}}
\newcommand{\bg}{\mathbf{g}}
\newcommand{\bS}{\mathbf{S}}
\newcommand{\bs}{\mathbf{s}}
\newcommand{\bw}{\mathbf{w}}
\newcommand{\eye}{\mathbf{I}}
\newcommand{\bU}{\mathbf{U}}
\newcommand{\bV}{\mathbf{V}}
\newcommand{\bW}{\mathbf{W}}
\newcommand{\bn}{\mathbf{n}}
\newcommand{\bv}{\mathbf{v}}
\newcommand{\bwv}{\mathbf{wv}}
\newcommand{\bq}{\mathbf{q}}
\newcommand{\bR}{\mathbf{R}}
\newcommand{\bi}{\mathbf{i}}
\newcommand{\bj}{\mathbf{j}}
\newcommand{\bp}{\mathbf{p}}
\newcommand{\bt}{\mathbf{t}}
\newcommand{\bJ}{\mathbf{J}}
\newcommand{\bu}{\mathbf{u}}
\newcommand{\bB}{\mathbf{B}}
\newcommand{\bD}{\mathbf{D}}
\newcommand{\bz}{\mathbf{z}}
\newcommand{\bP}{\mathbf{P}}
\newcommand{\bC}{\mathbf{C}}
\newcommand{\bA}{\mathbf{A}}
\newcommand{\bZ}{\mathbf{Z}}
\newcommand{\bff}{\mathbf{f}}
\newcommand{\bF}{\mathbf{F}}
\newcommand{\bo}{\mathbf{o}}
\newcommand{\bO}{\mathbf{O}}
\newcommand{\bc}{\mathbf{c}}
\newcommand{\bT}{\mathbf{T}}
\newcommand{\bQ}{\mathbf{Q}}
\newcommand{\bL}{\mathbf{L}}
\newcommand{\bl}{\mathbf{l}}
\newcommand{\ba}{\mathbf{a}}
\newcommand{\bE}{\mathbf{E}}
\newcommand{\bH}{\mathbf{H}}
\newcommand{\bd}{\mathbf{d}}
\newcommand{\br}{\mathbf{r}}
\newcommand{\be}{\mathbf{e}}
\newcommand{\bb}{\mathbf{b}}
\newcommand{\bh}{\mathbf{h}}
\newcommand{\bhh}{\hat{\mathbf{h}}}
\newcommand{\btheta}{\boldsymbol{\theta}}
\newcommand{\bTheta}{\boldsymbol{\Theta}}
\newcommand{\bpi}{\boldsymbol{\pi}}
\newcommand{\bphi}{\boldsymbol{\phi}}
\newcommand{\bpsi}{\boldsymbol{\psi}}
\newcommand{\bPhi}{\boldsymbol{\Phi}}
\newcommand{\bmu}{\boldsymbol{\mu}}
\newcommand{\bsigma}{\boldsymbol{\sigma}}
\newcommand{\bSigma}{\boldsymbol{\Sigma}}
\newcommand{\bGamma}{\boldsymbol{\Gamma}}
\newcommand{\bbeta}{\boldsymbol{\beta}}
\newcommand{\bomega}{\boldsymbol{\omega}}
\newcommand{\blambda}{\boldsymbol{\lambda}}
\newcommand{\bLambda}{\boldsymbol{\Lambda}}
\newcommand{\bkappa}{\boldsymbol{\kappa}}
\newcommand{\btau}{\boldsymbol{\tau}}
\newcommand{\balpha}{\boldsymbol{\alpha}}
\newcommand{\nR}{\mathbb{R}}
\newcommand{\nN}{\mathbb{N}}
\newcommand{\nL}{\mathbb{L}}
\newcommand{\cN}{\mathcal{N}}
\newcommand{\cA}{\mathcal{A}}
\newcommand{\cM}{\mathcal{M}}
\newcommand{\cR}{\mathcal{R}}
\newcommand{\cB}{\mathcal{B}}
\newcommand{\cG}{\mathcal{G}}
\newcommand{\cL}{\mathcal{L}}
\newcommand{\cH}{\mathcal{H}}
\newcommand{\cS}{\mathcal{S}}
\newcommand{\cT}{\mathcal{T}}
\newcommand{\cO}{\mathcal{O}}
\newcommand{\cC}{\mathcal{C}}
\newcommand{\cP}{\mathcal{P}}
\newcommand{\cE}{\mathcal{E}}
\newcommand{\cI}{\mathcal{I}}
\newcommand{\cF}{\mathcal{F}}
\newcommand{\cK}{\mathcal{K}}
\newcommand{\cV}{\mathcal{V}}
\newcommand{\cY}{\mathcal{Y}}
\newcommand{\cX}{\mathcal{X}}
\newcommand{\cZ}{\mathcal{Z}}
\def\bgamma{\boldsymbol\gamma}

\newcommand{\specialcell}[2][c]{%
  \begin{tabular}[#1]{@{}c@{}}#2\end{tabular}}

\newcommand{\figref}[1]{\Fig~\ref{#1}}
\newcommand{\secref}[1]{Section~\ref{#1}}
\newcommand{\algref}[1]{Algorithm~\ref{#1}}
\newcommand{\eqnref}[1]{Eq.~\ref{#1}}
\newcommand{\tabref}[1]{Table~\ref{#1}}

\newcommand{\rulesep}{\unskip\ \vrule\ }

\newcommand{\KLD}[2]{D_{\mathrm{KL}} \Big(#1 \mid\mid #2 \Big)}

\renewcommand{\b}{\ensuremath{\mathbf}}

\def\mc{\mathcal}
\def\mb{\mathbf}

\newcommand{\T}{^{\raisemath{-1pt}{\mathsf{T}}}}

\makeatletter
\DeclareRobustCommand\onedot{\futurelet\@let@token\@onedot}
\def\@onedot{\ifx\@let@token.\else.\null\fi\xspace}
\def\eg{{\em e.g}\onedot} \def\Eg{E.g\onedot}
\def\ie{{\em i.e}\onedot} \def\Ie{I.e\onedot}
\def\cf{cf\onedot} \def\Cf{Cf\onedot}
\def\etc{{\em etc}\onedot} \def\vs{vs\onedot}
\def\wrt{wrt\onedot}
\def\dof{d.o.f\onedot}
\def\etal{et~al\onedot} \def\iid{i.i.d\onedot}
\def\Fig{Fig\onedot} \def\Eqn{Eqn\onedot} \def\Sec{Sec\onedot} \def\Alg{Alg\onedot}
\makeatother

\newcommand{\xdownarrow}[1]{%
  {\left\downarrow\vbox to #1{}\right.\kern-\nulldelimiterspace}
}

\newcommand{\xuparrow}[1]{%
  {\left\uparrow\vbox to #1{}\right.\kern-\nulldelimiterspace}
}

\renewcommand\UrlFont{\color{blue}\rmfamily}

\newcommand*\rot{\rotatebox{90}}
\newcommand{\boldparagraph}[1]{\noindent{\bf #1:} }
\newcommand{\boldquestion}[1]{\noindent{\bf #1} }

\newcommand{\cmark}{\ding{51}}%
\newcommand{\xmark}{\ding{55}}%

\title{Compositional Video Synthesis by Temporal Object-Centric Learning}
\author{Adil Kaan Akan, Yucel Yemez
\thanks{$\bullet$ Adil Kaan Akan and Yucel Yemez are with Department of Computer Engineering, Koc University.
Yucel Yemez is with KUIS AI Center, Koc University. \indent \indent E-mail: kakan20@ku.edu.tr, yyemez@ku.edu.tr}
\thanks{This work has been submitted to the IEEE for possible publication. Copyright may be transferred without notice, after which this version may no longer be accessible.}}

\markboth{Journal of \LaTeX\ Class Files,~Vol.~18, No.~9, September~2020}%
{How to Use the IEEEtran \LaTeX \ Templates}

\IEEEtitleabstractindextext{%
We present a novel framework for compositional video synthesis that leverages temporally consistent object-centric representations, extending our previous work, SlotAdapt, from images to video. While existing object-centric approaches either lack generative capabilities entirely or treat video sequences holistically, thus neglecting explicit object-level structure, our approach explicitly captures temporal dynamics by learning pose invariant object-centric slots and conditioning them on pretrained diffusion models. This design enables high-quality, pixel-level video synthesis with superior temporal coherence, and offers intuitive compositional editing capabilities such as object insertion, deletion, or replacement, maintaining consistent object identities across frames. Extensive experiments demonstrate that our method sets new benchmarks in video generation quality and temporal consistency, outperforming previous object-centric generative methods. Although our segmentation performance closely matches state-of-the-art methods, our approach uniquely integrates this capability with robust generative performance, significantly advancing interactive and controllable video generation and opening new possibilities for advanced content creation, semantic editing, and dynamic scene understanding.
}

\maketitle
\IEEEdisplaynontitleabstractindextext

\begin{IEEEkeywords}
Object-centric learning, compositional video generation and editing, unsupervised video object segmentation, slot diffusion, invariant slot attention \vspace{-0.5cm}
\end{IEEEkeywords}

\section{Introduction}
\label{sec:intro}
The real world is inherently compositional, made up of distinct entities that can be flexibly combined and reconfigured into richer structures. This property underlies core cognitive abilities such as abstraction, causal reasoning, and systematic generalization~\cite{fodor88, lake2017building, bahdanau2018systematic}. However, beyond static structure, real-world environments are also deeply dynamic: objects move, interact, appear, and disappear over time. Humans naturally parse these dynamics into persistent, interacting entities, forming temporally coherent mental models that support understanding and prediction~\cite{spelke2007core, ullman2017mind}. Modeling such temporal compositionality remains a major challenge for artificial systems. While object-centric learning has shown promise in uncovering latent structure in images~\cite{greff2020binding}, extending these ideas to video requires capturing not only spatial grouping, but also object continuity, interaction, and transformation across time. Although recent text-to-video models have achieved impressive synthesis quality, they generally lack compositional structure and offer limited control over object-level content. Bridging this gap is essential for building video models that can reason, generalize, and interact with dynamic environments in a human-like, compositional manner.

Recent progress in object-centric learning (OCL) has been especially notable in static image domains, with models such as SlotDiffusion~\cite{wu2023slotdiffusion}, Latent Slot Diffusion (LSD)~\cite{jiang2023lsd}, and our prior work SlotAdapt~\cite{akan2025slotadapt} achieving strong results in unsupervised object discovery, segmentation, and image generation. These methods decompose scenes into discrete, interpretable object representations (“slots”)~\cite{lake2017building, scholkopf2021toward}, enabling structured understanding and high-fidelity synthesis. Importantly, they offer a pathway toward models that can generalize compositionally by reasoning over object-level primitives.

In our previous work, SlotAdapt~\cite{akan2025slotadapt}, we introduced a framework that leverages pretrained diffusion models conditioned with slot-based representations via adapter layers. It outperformed earlier slot-based approaches in both segmentation and image generation and enabled compositional editing of real-world images—an ability that prior models lacked. However, generative modeling from object-centric representations in video remains largely unexplored.

The broader OCL field has gradually progressed from synthetic datasets~\cite{johnson2017clevr, karazija2021clevrtex} to real-world images~\cite{everingham2010pascalvoc, lin2014coco} and videos~\cite{perazzi2016benchmark, Xu2018ECCV, Yang2019CVPRb, Li2013CVPR, Ochs2013PAMI}, typically within an autoencoding framework~\cite{greff2019iodine, locatello2020object}. These models aim to uncover object structure by reconstructing input frames using architectural or data-driven priors, often guided by static cues like color, shape, or pretrained features.

Extending object-centric generative modeling to the video domain introduces unique challenges. Models must capture temporal continuity, dynamic object transformations, and multi-object interactions across frames. Most existing approaches focus on segmentation or tracking and enforce temporal consistency through auxiliary signals such as optical flow~\cite{kipf2021savi} or depth maps~\cite{elsayed2022savi++}, or by incorporating architectural biases that promote slot stability~\cite{singh2022steve, wu2023slotformer} such as slot-aligned cross-frame attention and slot-specific recurrent updates. While such cues can aid learning, they introduce substantial computational overhead and are fragile in the presence of motion blur, deformation, or occlusion. Recent works in unsupervised temporal OCL, such as SOLV~\cite{Aydemir2023NeurIPS}, TC-Slot~\cite{manasyan2024temporally}, and others~\cite{ding2024betrayed, qian2024rethinking}, explore learning temporally consistent slots using pretrained vision encoders (e.g., DINO~\cite{caron2021dino, seitzer2023dinosaur} or CLIP~\cite{pmlr-v139-radford21a}). While effective in producing stable representations, these methods lack generative capabilities and compositional control, relying on feature-level decoding instead of pixel-level synthesis. As a result, they cannot support interactive editing or video generation.

Meanwhile, diffusion-based video generation models~\cite{ho2022imagenvideo, singer2022makeavideo, blattmann2023align} have demonstrated high-quality synthesis using large-scale text-video datasets. However, these models treat scenes as monolithic visual fields and do not incorporate object-centric structure. This limits compositional controllability, semantic disentanglement, and consistent object identity over time. Some recent efforts~\cite{wang2023videocomposer, bar2024lumiere, zhang2023controlvideo} introduce control mechanisms through keyframes, masks, or motion guidance, but they remain orthogonal to our approach and do not utilize slot-level object representations.

In this work, we propose a fully self-supervised framework for generative video modeling that combines object-centric representation learning with high-quality synthesis capabilities. Building on SlotAdapt~\cite{akan2025slotadapt}, we extend object-centric generation into the temporal domain by learning temporally consistent slots that encode object identity, motion, and interactions—without relying on external signals like optical flow or depth. By conditioning a pretrained diffusion model on these slot-based temporal features, we achieve video reconstructions that are both temporally coherent and semantically grounded through pixel-level diffusion synthesis.

A key methodological distinction lies in our model's ability to discover temporal structure directly from raw video data. Rather than relying on architectural constraints such as STEVE's deterministic slot transitions~\cite{singh2022steve} or SlotFormer's slot-aligned cross-frame attention~\cite{wu2023slotformer} or hand-designed temporal signals like optical flow guidance in G-SWM and STOVE~\cite{lin2020gswm, kossen2020stove}, our approach learns dynamic object relationships through self-supervised conditioning of a pretrained diffusion model with slot-based temporal features. This design enables object-centric video generation with compositional editing control, capabilities absent in existing paradigms.

Unlike conventional text-to-video models~\cite{ho2022imagenvideo, singer2022makeavideo} that operate holistically without object-level structure, or feature-decoding temporal OCL methods~\cite{Aydemir2023NeurIPS, Yang2019CVPRb} that lack generative capabilities, our slot-based method supports flexible compositional control, allowing users to insert, remove, or modify individual objects while maintaining temporal coherence and scene consistency.
Empirically, our model achieves competitive segmentation performance compared to existing unsupervised methods such as SOLV~\cite{Aydemir2023NeurIPS} or OCLR~\cite{xie2022oclr}, while also enabling full-resolution video synthesis. Compared to image-based OCL models including LSD~\cite{jiang2023lsd}, SlotDiffusion~\cite{wu2023slotdiffusion}, and SlotAdapt~\cite{akan2025slotadapt}, our approach establishes new benchmarks for video synthesis with substantial improvements in temporal consistency and visual fidelity. To the best of our knowledge, our method is the first to combine object-centric representation learning with high-quality video generation on real-world videos, enabling compositional editing where individual objects can be inserted, removed, or modified while maintaining temporal coherence, in a self-supervised framework.

\section{Related Work}
\label{sec:related_work}

\textbf{Unsupervised Object-Centric Learning (OCL).}  
Unsupervised object-centric learning aims to decompose visual scenes into discrete and semantically meaningful representations, typically referred to as ``slots”, without explicit supervision. Early works such as Attend-Infer-Repeat (AIR)~\cite{eslami2016air} and SQAIR~\cite{kosiorek2018sqair} used iterative inference and variational decoders but were limited to simple, low-complexity settings. Slot Attention~\cite{locatello2020object} introduced a permutation-invariant attention mechanism that enabled effective object discovery and segmentation on synthetic datasets. This approach has since been extended with autoregressive transformers (e.g., SLATE~\cite{singh2021slate}, discrete tokenization~\cite{singh2022steve}, and self-supervised objectives in feature space (e.g., DINOSAUR~\cite{seitzer2023dinosaur}). More recently, hierarchical approaches have been explored, such as COCA-Net~\cite{kucuksozen2025cvpr}, which introduces a hierarchical clustering strategy with spatial broadcast decoding to achieve superior segmentation performance on synthetic datasets.

Several methods have explored temporal extensions of object-centric modeling by encouraging slot consistency across frames. SAVi~\cite{kipf2021savi} and SAVi++~\cite{elsayed2022savi++} incorporate predictor/corrector architecture which relies on auxiliary signals such as optical flow and depth to guide temporal slot alignment. However, such additional cues are prone to failure under deformation, occlusion, or motion blur. To address these limitations, self-supervised alternatives have been proposed: TC-Slot~\cite{manasyan2024temporally} employs contrastive learning to enforce cross-frame slot consistency; Betrayed-by-Attention~\cite{ding2024betrayed} introduces a combination of hierarchical clustering and consistency objective to stabilize slot attention; RIV~\cite{qian2024rethinking} reconsiders image-to-video transfer from an object-centric lens; and SOLV~\cite{Aydemir2023NeurIPS} applies Invariant Slot Attention~\cite{biza2023invariant} to cluster DINO features across time for coherent unsupervised slot assignments. Early work in future prediction and scene understanding \cite{akan2021slamp,akan2022stretchbev} also highlighted the relevance of object-centric temporal modeling, motivating subsequent developments in this direction.

Despite these advances in segmentation and tracking, most temporal OCL methods rely on feature-level decoding and thus lack pixel-level generative capabilities—crucial for photorealistic synthesis and fine-grained control. Moreover, they do not support structured or compositional manipulation of real-world content. In contrast, our method significantly advances this direction by enabling pixel-level, temporally consistent video generation and slot-based compositional editing in complex, real-world scenarios. \\
\textbf{Diffusion Models for Image Generation.} Diffusion models~\cite{sohl2015dm, ho2020ddpm} have rapidly become the state-of-the-art for high-fidelity image generation, with models like ADM~\cite{dhariwal2021adm}, Imagen~\cite{saharia2022imagen}, and DALLE-2~\cite{ramesh2022dalle2} showcasing controllable, text-conditioned synthesis. Latent Diffusion Models (LDMs)~\cite{rombach2022ldm} reduce the computational cost by operating in compressed latent space and offer flexible conditioning through cross-attention. Several recent works have aimed to scale or improve the underlying architecture~\cite{flux2024, esser2024scaling, podell2023sdxl, chen2023pixart}. Recent extensions like T2I-Adapters~\cite{mou2024t2i} allow lightweight adaptation of pretrained diffusion models to new tasks without retraining the core model. \\
\textbf{Object-Centric Diffusion Models.}  
Several recent works combine object-centric learning with diffusion-based generation to improve compositionality and controllability in images. LSD~\cite{jiang2023lsd}, SlotDiffusion~\cite{wu2023slotdiffusion}, GLASS~\cite{singh2024guidedsa} and our prior work SlotAdapt~\cite{akan2025slotadapt} use slot-based representations to condition diffusion decoders, enabling unsupervised object discovery and structured image generation. However, LSD and GLASS~\cite{singh2024guidedsa} rely on pretrained diffusion models, and often suffer from a mismatch between learned slots and the pretrained attention layers, leading to degraded generative quality. SlotDiffusion addresses this by training the diffusion model from scratch, but this approach requires extensive compute and lacks generalization.

 In our previous work, SlotAdapt~\cite{akan2025slotadapt}, we introduce adapter layers specifically designed to better align slot semantics with pretrained diffusion priors, achieving superior segmentation and image generation performance compared to prior slot-based methods. Furthermore, SlotAdapt was the first to successfully demonstrate compositional generation and editing capabilities on challenging real-world image datasets. Despite these advancements, existing object-centric diffusion models—including SlotAdapt—primarily focus on static images, leaving the challenge of modeling temporal object dynamics largely unaddressed. Recognizing this crucial gap, our current work explicitly extends SlotAdapt's foundational ideas into the temporal domain, enabling coherent object-centric video synthesis and editing. We significantly advance beyond previous image-centric methods by incorporating mechanisms to model object continuity and interaction over time. \\
 \textbf{Video Generation with Diffusion Models.}  
Recent advances in video synthesis have been driven by diffusion models capable of producing high-resolution, temporally consistent outputs. Ho et al.~\cite{ho2022video} introduced Video Diffusion Models (VDMs), extending denoising diffusion to sequential data. Follow-up models such as Imagen Video~\cite{ho2022imagenvideo}, Make-A-Video~\cite{singer2022makeavideo}, Phenaki~\cite{villegas2022phenaki}, and Align Your Latents~\cite{blattmann2023align} further advanced text-to-video generation using large-scale paired datasets. Despite their impressive results, these models operate on holistic, entangled scene representations and lack explicit object-level structure. This limits their ability to support semantic disentanglement, compositional reasoning, or precise object control.

While there have been early steps toward compositionality in video—e.g., Tune-A-Video~\cite{wu2023tune}, VideoComposer~\cite{wang2023videocomposer}, and ControlVideo~\cite{zhang2023controlvideo}—these methods rely on fine-tuning or auxiliary conditioning (e.g., masks, keyframes, or motion) rather than on disentangled object representations. As such, they lack an object-centric inductive bias and cannot perform structured manipulations such as adding, removing, or replacing individual entities in a scene.

In contrast, our work addresses this fundamental limitation by learning object-centric temporal representations that enable direct generative control. Unlike existing approaches that operate post-hoc on pretrained models~\cite{wu2023tune, wang2023videocomposer, zhang2023controlvideo, mou2024t2i}, we develop an end-to-end framework that jointly learns object discovery and temporally consistent synthesis from raw video data.
\vspace{-0.75cm}

\section{Preliminaries}
\label{sec:preliminaries}\vspace{-0.2cm}
\subsection{Slot Attention}
\label{sec:slot_attention}
Slot Attention \citep{locatello2020object} provides a framework to decompose visual scenes into discrete, interpretable components termed \textit{slots}. Slots are initialized randomly and iteratively refined to represent distinct objects or entities within the input data through an attention mechanism. Formally, the slot update rule can be described as:
\begin{align}
\bU^{(m)} &= \text{Attention}\left(q(\bS^{(m)}), k(\bF), v(\bF)\right), \\
\bS^{(m+1)} &= \text{GRU}(\bS^{(m)}, \bU^{(m)}),
\end{align}
where $q$, $k$, and $v$ represent learnable linear transformations corresponding to queries, keys, and values, respectively; $\bF$ denotes extracted image features; $\bS$ are the slot vectors; $\bU$ represents the update generated by the attention operation; $m$ is the iteration index. Slots compete for pixels through attention over the slot dimension, encouraging each slot to bind to a distinct region or object in the scene. The effectiveness of Slot Attention lies in its capacity to disentangle and encode complex scenes into structured representations without supervision. This is achieved through the ability of the mechanism to assign representational capacity where needed.\vspace{-0.4cm}
\vspace{-0.3cm}
\subsection{Diffusion Models}
\label{sec:diffusion_models}

Diffusion models \citep{sohl2015dm, ho2020ddpm} have recently emerged as powerful generative frameworks, producing high-quality samples by modeling the reverse process of a progressive noise addition. Given an input image $\mathbf{X}$, the model learns to reverse a noisy image $\mathbf{X}_\tau$ at timestep $\tau$\footnote{We use the symbol $\tau$ for diffusion timesteps to distinguish from $t$ which denotes video frame index throughout this paper.} back to the original data distribution. 

The underlying generative process consists of a series of reverse diffusion steps that transform samples from a noise distribution to the data distribution $p(\mathbf{X})$, where each step is defined by conditional probabilities $p(\mathbf{X}_{\tau-1}|\mathbf{X}_\tau)$. During training, a loss function $\mathcal{L}$ is optimized that penalizes the expected prediction error across random timesteps $\tau$, effectively teaching the model to denoise progressively corrupted inputs and reconstruct the original data distribution via:\vspace{-0.4cm}
\begin{equation}
\label{eq:diffusion_loss}
\mathcal{L{(\btheta)}} = \mathbb{E}_{\mathbf{X} \sim p(\mathbf{X}), \mathbf{\epsilon}_\tau \sim \mathcal{N}(0,1), \tau \sim \mathcal{U}(1, T)} \left[\Vert \mathbf{\epsilon}_\tau - \mathbf{\epsilon}_{{\btheta}}(\mathbf{X}_\tau, \tau, y) \Vert_{2}^2\right],
\end{equation} where $\epsilon_{\btheta}$ denotes the noise prediction network parameterized by $\btheta$, and $y$ represents optional conditioning information. 
For high-resolution images, Latent Diffusion Models~(LDM)~\cite{rombach2022ldm} are proposed to perform the training and sampling in a low-dimensional latent space, obtained through a pretrained variational autoencoder~(VAE)~\cite{rombach2022ldm}. This approach improves computational efficiency while preserving generative fidelity by decoupling image compression and synthesis.

\subsection{SlotAdapt}
\label{sec:slotadapt_method}
In our prior work, SlotAdapt \citep{akan2025slotadapt}, we introduced a robust methodology that integrates the strengths of Slot Attention and pretrained diffusion models to achieve effective object-centric image generation and compositional editing. Given an input image $\bX \in \mathbb{R}^{H \times W \times 3}$, SlotAdapt first employs a visual backbone, typically based on pretrained vision transformers like DINO~\cite{caron2021dino,oquab2023dinov2}, to extract a compact set of visual features represented by $\bF \in \mathbb{R}^{h \cdot w \times d}$, where $h, w$ indicate reduced spatial dimensions and $d$ is the feature dimension. Slot Attention is then applied to these features to dynamically allocate object-centric representations into discrete slots $\bS \in \mathbb{R}^{K \times D_{\text{slot}}}$, with each slot ideally corresponding to a separate object or distinct entity in the scene.

These learned slots condition a pretrained Stable Diffusion decoder, which we augment with specialized adapter layers. Specifically, these adapter layers, implemented as additional cross-attention modules inserted into each downsampling and upsampling block of the pretrained U-Net architecture, enable explicit conditioning on slot representations. This design significantly differs from prior approaches such as SlotDiffusion~\cite{wu2023slotdiffusion} and LSD~\cite{jiang2023lsd}, which condition on slots via cross-attention layers originally trained for text embeddings—leading to potential misalignment between slot semantics and the pretrained diffusion attention pathways. By separating the adapter conditioning from the text-based conditioning modules of the pretrained diffusion model, we allow slots to focus exclusively on object-level semantics, independent from the original textual embedding space.

To further enhance the conditioning mechanism, SlotAdapt introduces a dedicated register token, $\br$, computed by mean pooling the slot representations or alternatively, the visual backbone features, following~\cite{darcet2024register}. This register token captures overall contextual scene information and is conditioned via the original text-based cross-attention layers of the diffusion model, enabling slots to remain focused on specific objects without being diluted by global scene context.

During training, SlotAdapt utilizes a reconstruction-based loss framed as a noise prediction objective within the diffusion framework. Formally, given a noisy latent image representation $\bX_\tau$ obtained at diffusion timestep $\tau$, SlotAdapt aims to predict the noise $\mathbf{\epsilon}_\tau$ using slot $\bS$ and register $\br$ conditioning:\vspace{-0.2cm}
\begin{equation}
\mathcal{L}(\btheta) = \mathbb{E}_{\mathbf{X} \sim p(\mathbf{X}), \boldsymbol{\epsilon}_\tau \sim \mathcal{N}(0, \mathbf{I}), \tau \sim \mathcal{U}(1, T)} \left[ \left\| \boldsymbol{\epsilon}_\tau - \epsilon_{\btheta}(\mathbf{X}_\tau, \tau, \mathbf{S}, \mathbf{r}) \right\|_2^2 \right], \vspace{-0.05cm} 
\end{equation} where $\epsilon_{\btheta}$ denotes the diffusion-based noise prediction model parameterized by $\btheta$.

Importantly, SlotAdapt freezes the pretrained diffusion model parameters throughout training, updating only the adapter layers and the slot attention mechanism. This strategy allows the method to efficiently leverage the powerful generative prior encoded in pretrained diffusion models, significantly enhancing generative performance and training efficiency. Additionally, SlotAdapt introduces a guidance loss that enforces alignment between slot attention masks and the diffusion model's cross-attention maps, leveraging the prior knowledge residing in the frozen diffusion layers to improve slot-object correspondence during training.

SlotAdapt achieves state-of-the-art results on object-centric image generation and compositional editing, particularly on complex real-world datasets. Its robust methodology, which combines structured object representations with pretrained diffusion, provides a strong foundation for this work, where we extend these ideas into the temporal domain for object-centric video generation and editing.\vspace{-0.33cm}

\begin{figure*}[!ht]
\centering
\includegraphics[width=0.95\textwidth]{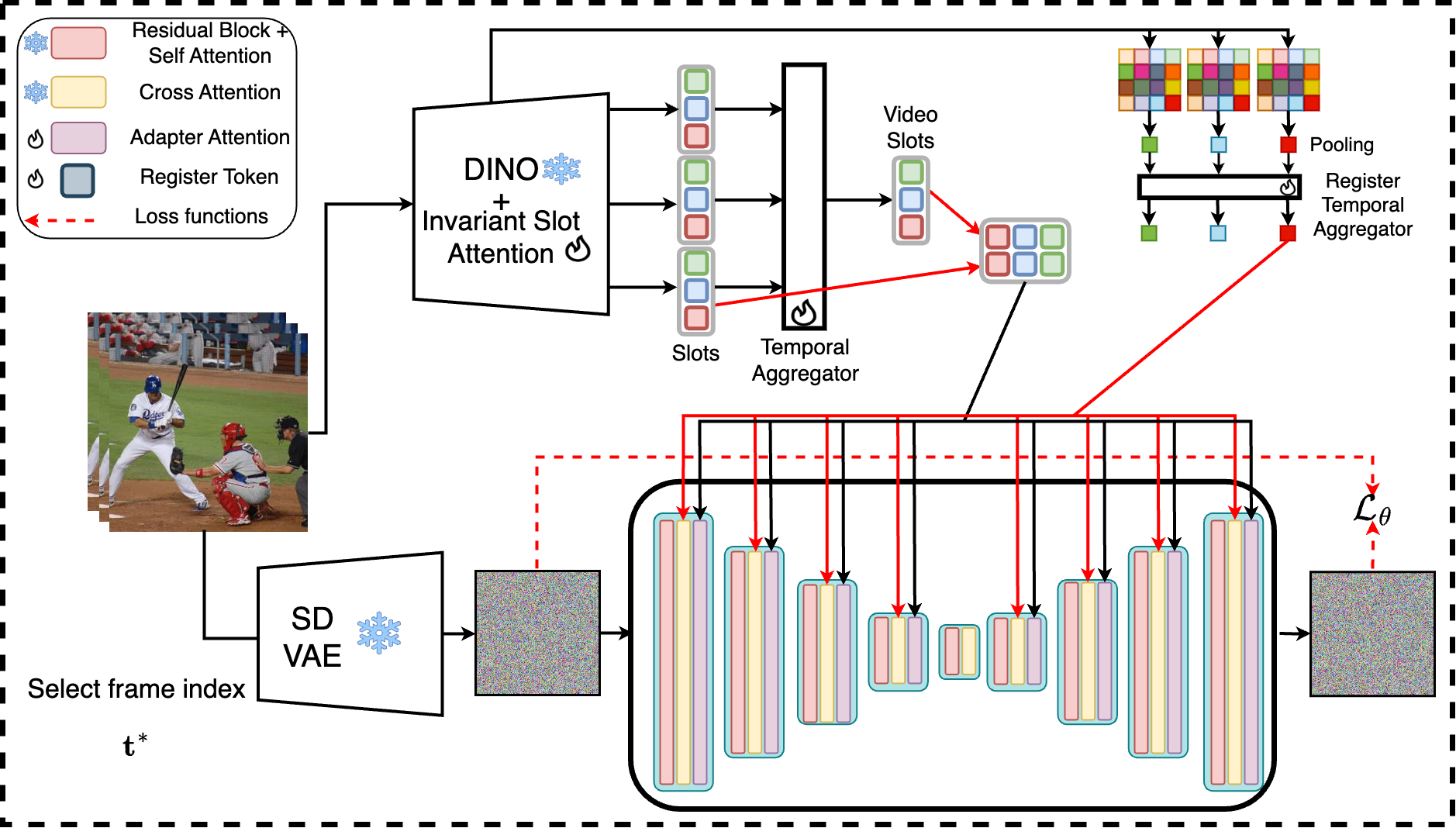}
\vskip -0.1in
\caption{\textbf{Architecture Block Diagram}
We extract object-centric and temporally consistent information from input video frames using a visual backbone composed of DINOv2 and Invariant Slot Attention (ISA). The ISA mechanism generates slots for each frame, which are then aggregated temporally using a Transformer-based temporal aggregator to produce enriched, temporally-aware video slots. Concurrently, global context information is summarized by average pooling frame-level features and further processed through a separate temporal aggregator to produce global scene tokens. A pretrained Stable Diffusion Variational Autoencoder (VAE) encodes a randomly selected frame into latent space, and Gaussian noise is subsequently added. The diffusion model is explicitly conditioned on both the temporally aggregated video slots and the slots from the selected frame (shown here as the last frame for visualization purposes, though in practice this could be any frame) via additional adapter attention layers, and on the global scene token using the diffusion model's native cross-attention layers. During training, the model learns to predict the added noise, with the diffusion loss ($\mathcal{L}_{\theta}$) measuring the prediction error. This training strategy ensures temporally coherent, object-centric video synthesis and intuitive compositional editing capabilities across video frames.
}
\vspace{-0.6cm}
\label{fig:method}
\end{figure*}
\section{Methodology}
\label{sec:method}

Our proposed framework extends SlotAdapt~\citep{akan2025slotadapt} to generate high-quality, temporally coherent videos from object-centric representations. It comprises two core components: a temporal object-centric encoder that captures dynamics and interactions across video frames, and a slot-conditioned diffusion decoder that synthesizes photorealistic frames. The entire architecture is trained in a fully self-supervised manner, without reliance on auxiliary cues such as optical flow or depth.\vspace{-0.35cm}
\subsection{Object-Centric Temporal Encoding}
\label{sec:encoder}
Given an input video composed of $L$ frames, our model first extracts visual features from each frame using a frozen DINOv2 backbone \citep{oquab2023dinov2}. Specifically, each frame at timestep $t$ is transformed into patch-level features $\mathbf{F}_t \in \mathbb{R}^{N \times d}$, where each patch corresponds to a distinct spatial region of the frame, $N$ is the number of total patches and $d$ is the feature dimension.

To achieve temporally consistent object-centric representations, we utilize Invariant Slot Attention (ISA)~\citep{biza2023invariant}. While original ISA enforces spatial invariance by decoupling object identity from spatial information, our approach retains ISA's slot attention mechanism but replaces the spatial broadcast decoder with a diffusion decoder that is directly conditioned on pose-invariant slots, with register tokens providing the necessary spatial context for coherent generation (discussed in Section~\ref{subsec:isa_adaptation}).

Formally, ISA decomposes each frame into $K$ slot vectors, each slot vector $\mathbf{z}_t^j \in \mathbb{R}^{D_{\text{slot}}}$ attending to frame features as follows:\vspace{-0.2cm}
\begin{align}
\label{eq:inv_slot_att}
    &\mathbf{A}_t := \underset{j=1,\dots,K}{\mathrm{softmax}} \left(\mathbf{M}_t\right) \in \mathbb{R}^{K\times N}, \\
    &\mathbf{m}_t^j := \dfrac{1}{\sqrt{d}}\ \  p\left(k\left(\mathbf{F}_t\right) + g(\mathbf{G}^j_{\text{rel}, t})\right) q(\mathbf{z}^j_t) \in \mathbb{R}^{N}
\end{align} where $q: \mathbb{R}^{D_{\text{slot}}} \to \mathbb{R}^{D_{\text{slot}}}$, $k: \mathbb{R}^{d} \to \mathbb{R}^{D_{\text{slot}}}$, $p: \mathbb{R}^{D_{\text{slot}}} \to \mathbb{R}^{D_{\text{slot}}}$, and $g: \mathbb{R}^{2} \to \mathbb{R}^{D_{\text{slot}}}$ are learnable linear projections applied to each patch and slot vector independently and $\mathbf{G}_{\text{rel},t}^j \in \mathbb{R}^{N \times 2}$ encodes the relative spatial position of each patch with respect to slot $j$ using a learnable scale and shift transformation. The unnormalized attention scores $\mathbf{m}_t^j \in \mathbb{R}^{N}$ represent the $j$-th row of the matrix $\mathbf{M}_t \in \mathbb{R}^{K \times N}$, computed for all $K$ slots. The softmax is then applied column-wise (over slots) to obtain the attention matrix $\mathbf{A}_t \in \mathbb{R}^{K \times N}$. The attention matrix is computed per frame, where each row represents the contribution of each patch to the update of slot $\mathbf{z}_t^j$ at time $t$. Please refer to the supplementary material for details on how we adapted ISA to our case.

These slot vectors are iteratively refined using a Gated Recurrent Unit (GRU) and residual Multi-Layer Perceptron (MLP), following the Slot Attention mechanism~\cite{locatello2020object}, to reinforce their object binding consistency across frames. The output per frame is a set of $K$ slots, denoted by $\mathbf{S}_t = \{\mathbf{z}_t^1, \dots, \mathbf{z}_t^K\}$, accompanied by corresponding soft attention masks $\mathbf{A}_t$. \\
\textbf{Temporal Context Aggregation.} To capture broader temporal context, we concatenate slot sequences from all frames within the video segment and pass them through a Transformer encoder augmented with learnable temporal positional embeddings:\vspace{-0.1cm}
\begin{equation}
\tilde{\mathbf{S}}_{1:L} = \mathrm{Transformer}(\mathbf{S}_{1:L}), \quad \mathbf{S}_{1:L} = \mathrm{concat}(\mathbf{S}_1, \dots, \mathbf{S}_L).
\end{equation}\vspace{-0.05cm}
The resulting temporally aggregated slots $\tilde{\mathbf{S}}_{1:T}$ are reshaped and concatenated back with original frame-wise slots, creating augmented slots for each frame:\vspace{-0.1cm}
\begin{equation}
\label{eq:augmented_slot}
\tilde{\mathbf{S}}_t^{\mathbf{+}} = \mathrm{concat}(\mathbf{S}_t, \tilde{\mathbf{S}}_t).
\end{equation}
Additionally, we compute global scene-level context vectors by average-pooling the DINO features of each frame. These pooled vectors undergo further temporal encoding via another Transformer encoder to yield temporally-aware global tokens $\tilde{\mathbf{r}}_t \in \mathbb{R}^d$:\vspace{-0.1cm}
\begin{equation}
\label{eq:register_aggregator}
\mathbf{r}_t = \frac{1}{N} \sum_{i=1}^{N} \mathbf{F}_{t,i}, \quad \tilde{\mathbf{r}}_t = \mathrm{Transformer}(\mathbf{r}_1, \dots, \mathbf{r}_L)_t. \vspace{-0.05cm}
\end{equation} These global tokens summarize high-level dynamics and semantic context, complementing the slot-based object representations.

During training, we adopt a 1-frame training strategy: we randomly select a single frame index $t^\star$ at each iteration and retain only its temporally enriched slot set $\tilde{\mathbf{S}}_{t^\star}^{\mathbf{+}}$ and global context token $\tilde{\mathbf{r}}_{t^\star}$ for decoding. This 1-frame training allows efficient supervision using a pretrained image-based diffusion model.

\vspace{-0.3cm}

\subsection{Slot-Conditioned Diffusion Decoding}
\label{sec:decoder}

We decode the selected video frame using a pretrained Stable Diffusion model \citep{rombach2022ldm}. The chosen frame $\mathbf{V}_{t^\star}$ is encoded into a latent representation $\mathbf{X}_{t^\star} \in \mathbb{R}^{h \times w \times c}$ via the VAE encoder of the diffusion model. The latent representation is then perturbed by Gaussian noise according to the diffusion schedule, resulting in a noisy latent $\mathbf{X}_\tau$ at timestep $\tau$.

To condition the diffusion process explicitly on object-level semantics, we inject lightweight adapter-based cross-attention layers into each residual block of the U-Net decoder, following the SlotAdapt architecture~\cite{akan2025slotadapt}. Adapter layers are used to condition the diffusion U-Net with the augmented slots, $\tilde{\mathbf{S}}_{t^\star}^{\mathbf{+}}$. Concurrently, the native cross-attention layers of the U-Net, originally designed for textual embeddings, are used to condition the global scene token $\tilde{\mathbf{r}}_{t^\star}$, which effectively summarizes global contextual information.

The diffusion training objective aims to predict the added noise:\vspace{-0.1cm}
\begin{equation}
\mathcal{L}_{\text{diff}} = \left\| \boldsymbol{\epsilon} - \epsilon_\theta\left(\mathbf{X}_\tau, \tau, \tilde{\mathbf{S}}_{t^\star}^{\mathbf{+}}, \tilde{\mathbf{r}}_{t^\star}\right) \right\|^2, \  \mathrm{where} \  \boldsymbol{\epsilon}\sim\mathcal{N}(0,\mathbf{I}).
\end{equation}\vspace{-0.05cm}
Throughout training, we freeze the pretrained diffusion model parameters and optimize only the temporal slot encoder and the adapter layers. This ensures that the model benefits from the robust generative prior inherent in the pretrained diffusion backbone without requiring large-scale retraining.

Unlike SlotAdapt, we omit auxiliary attention-guidance losses. Since we employ 1-frame training (randomly selecting one frame per iteration for efficient decoding), applying guidance loss would only align slot attention masks with diffusion cross-attention masks for the selected frame, while other frames in the temporal window receive no such alignment signal. This inconsistent gradient flow across frames disrupts temporal coherence. Instead, we directly utilize encoder-derived attention masks without additional processing or merging, ensuring simplicity and stability.\\
\textbf{Inference.} At test time, we apply a sliding-window strategy, decoding the central frame within each overlapping window independently, following SOLV~\cite{Aydemir2023NeurIPS}. We align slot identities across frames using Hungarian matching based on slot similarity. This design supports temporally coherent video synthesis and provides intuitive compositional editing capabilities: objects can be explicitly manipulated across frames by modifying corresponding slots directly. In contrast to prior methods, which were limited to synthetic videos or static images, our approach successfully handles real-world dynamic video scenes, effectively bridging segmentation accuracy and high-quality generative performance. \vspace{-0.2cm}

\subsection{Invariant Slot Attention Adaptation for Diffusion Conditioning}
\label{subsec:isa_adaptation}

The effectiveness of our object-centric video generation framework relies on the assumption that the slot representations are temporally consistent and invariant to pose changes induced by motion. To achieve this, we employ ISA on the encoder side, which estimates slot-specific pose parameters and incorporates them into relative position encodings during slot formation. In the original ISA architecture, these pose estimates are also used during decoding via a spatial broadcast mechanism, enabling spatially accurate reconstructions from pose-invariant slots.

In our model, while the slot attention computation remains identical to that of the original ISA, we replace the spatial broadcast decoder with a pretrained diffusion decoder to enable high-quality generative modeling. This decoder is directly conditioned on the pose-invariant slots, which by design lack explicit spatial information. To compensate for this, we leverage the register tokens (introduced in Sections~\ref{sec:slotadapt_method} and \ref{sec:encoder}), which provide global pose and background context to the diffusion decoder. The slots and register tokens together maintain a disentangled representation of object identity and spatial attributes, enabling spatially coherent video generation within a diffusion-based framework.

\begin{figure}
    \centering
    \includegraphics[width=\linewidth]{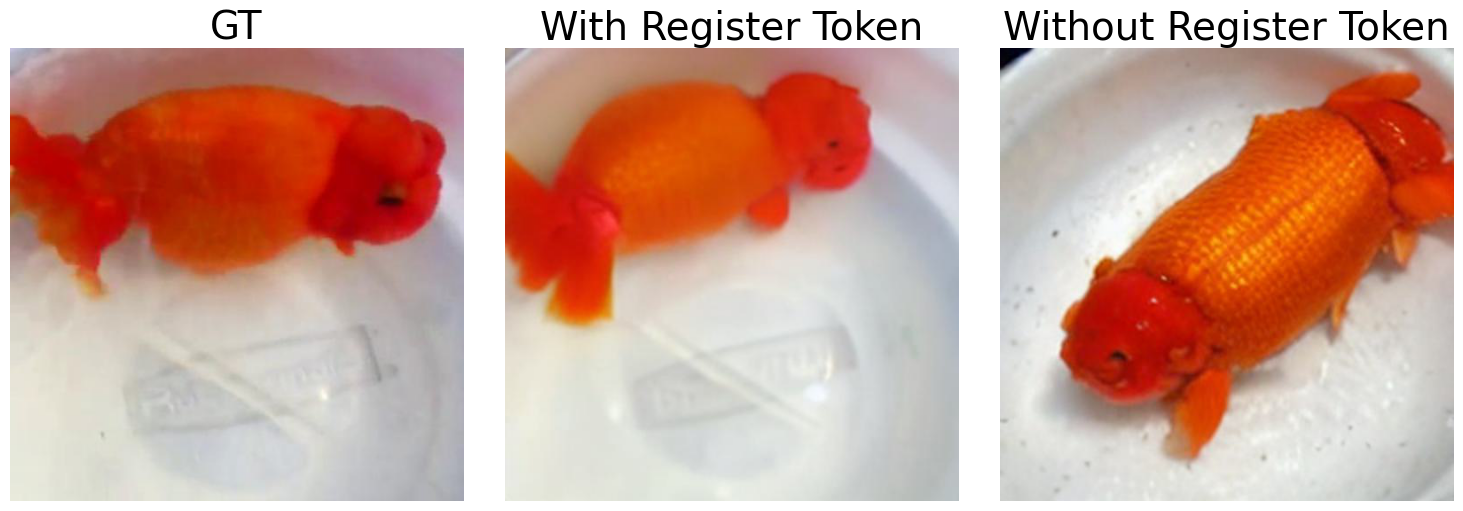}
    \includegraphics[width=\linewidth]{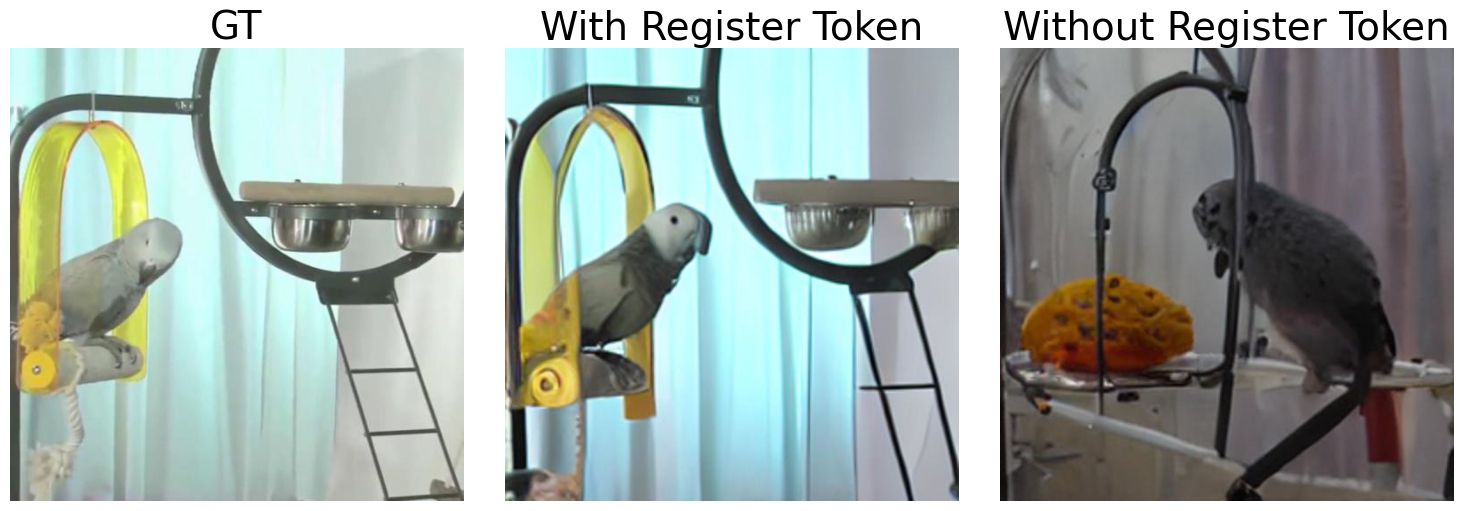}\vspace{-0.85cm}
    \caption{\textbf{Pose Invariance in Diffusion Conditioning.} Comparison of video frame generation with and without register tokens on YTVIS dataset. \textbf{Without register tokens} (middle), objects appear in incorrect positions and backgrounds deviate from ground truth. \textbf{With register tokens} (right), generations accurately match ground truth (left), confirming that register tokens handle pose information while slots maintain object identity focus. Full temporal sequences are provided in the Supplementary Material.}
    \label{fig:reg_token}\vspace{-0.62cm}
\end{figure}

To empirically verify the role of register tokens,  we generate images with and without register tokens (Fig.~\ref{fig:reg_token}) using a model trained with both components. When register tokens are omitted (replaced with zero vectors), the generated objects appear in incorrect positions, with incorrect scales and, in some cases, altered orientations compared to the ground truth. The backgrounds also deviate significantly from the original scenes. In contrast, when register tokens are included, the generated object positions, scales, orientations, and backgrounds closely match the ground truth. These results demonstrate that the register tokens effectively capture the pose and spatial context information that would otherwise be handled by the spatial broadcast decoder in the original ISA architecture. Please refer to the Supplementary Material for additional results on the role of register tokens in capturing spatial context.\vspace{-0.36cm}

\section{Experiments}
\label{sec:experiments}

We comprehensively evaluate our proposed framework against state-of-the-art object-centric learning methods. We focus on two core tasks: unsupervised video object segmentation and temporally consistent video generation. Experiments are conducted on two real-world datasets using widely adopted metrics to ensure rigorous and meaningful evaluation. \vspace{-0.4cm}
\subsection{Datasets}
We evaluate our approach on two widely used real-world video datasets: DAVIS17~\citep{perazzi2016benchmark} and YouTube-VIS 2019 (YTVIS19)~\citep{Yang2019CVPRb}. 

\textbf{DAVIS17} is a benchmark specifically tailored for video object segmentation. It contains short, high-quality video sequences annotated with precise ground-truth masks, requiring temporal consistency.

\textbf{YTVIS19} consists of diverse video sequences with complex scenes and significant variation in object appearance, pose, and background. Following prior work~\cite{Aydemir2023NeurIPS}, we evaluate our model on a subset comprising 300 videos selected from the original training set of 2,883 high-resolution videos, as YTVIS19 lacks an official validation or test set with provided ground-truth masks. 

Together, these datasets provide a challenging and realistic evaluation environment for both segmentation and generation tasks. \vspace{-0.25cm}

\subsection{Implementation Details}
We follow the implementation practices from our prior work, SlotAdapt \citep{akan2025slotadapt}, adapting and extending them to model temporal dynamics. Specifically, we use a frozen DINOv2~\citep{oquab2023dinov2} ViT-B/14 as the visual backbone to extract frame-level features. Invariant Slot Attention (ISA) is applied per frame with shared initialization across time. A transformer-based temporal aggregator enriches the slots with temporal context, using a temporal window of $L=5$ frames (2 past, 1 present, 2 future) following previous work~\cite{Aydemir2023NeurIPS}.

For decoding, we use Stable Diffusion v1.5~\citep{rombach2022ldm}, with adapters inserted as in SlotAdapt~\cite{akan2025slotadapt}. During training, we keep the Stable Diffusion model parameters fixed and optimize only the ISA, the temporal transformers, and the adapter layers. We train all models for 350K iterations on YTVIS19, then fine-tune for 50K iterations on DAVIS17. All experiments are conducted on 2× NVIDIA A40 GPUs with 48GB memory each. \vspace{-0.25cm}

\begin{table}[t]
    \caption{\textbf{Ablation study on YTVIS dataset.} We systematically evaluate the contribution of key components in our unified framework by comparing against our full model configuration. The full model uses invariant slot attention, DINO register tokens, register aggregator, and 1-frame training. We analyze the impact of removing individual components and varying training strategies. \vspace{-0.3cm}
    }
    \label{table:ablation-study}
    \centering
    \small
    \setlength{\tabcolsep}{8pt}
    \begin{tabular}{lcc}
    \toprule
    \textbf{Method Configuration} & \textbf{mIoU} & \textbf{FG-ARI} \\
    \midrule
    Full Model & 40.57 & 22.40 \\
    \midrule
    \multicolumn{3}{l}{\textit{Component Ablations}} \\
    w/o Register Aggregator & 39.22 & 20.49 \\
    w/ Slot Avg Register Tokens & 36.87 & 18.06 \\
    w/ Standard Slot Attention & 27.09 & 11.06 \\
    \midrule
    \multicolumn{3}{l}{\textit{Training Strategy Ablations}} \\
    5-frame Training & 40.67 & 22.00 \\
    5-frame Training + Guidance & 41.02 & 22.69 \\
    \bottomrule
    \end{tabular}
    \vspace{-0.65cm}
\end{table}

\subsection{Baselines}
For segmentation, we compare against SOLV~\citep{Aydemir2023NeurIPS}, a recent state-of-the-art method in unsupervised temporal object-centric learning.

For the video generation task, no prior method directly addresses object-centric video generation from unsupervised representations, as existing object-centric approaches use feature decoders such as those proposed by DINOSAUR~\cite{seitzer2023dinosaur}. Thus, we benchmark against existing object-centric image generation models: SlotDiffusion \citep{wu2023slotdiffusion}, Latent Slot Diffusion (LSD) \citep{jiang2023lsd}, and our previously developed SlotAdapt~\cite{akan2025slotadapt} method, by training them on video frames individually. This provides a rigorous baseline, highlighting our method's unique capability of generating coherent videos directly from object-centric representations that maintain temporal coherence. \vspace{-0.4cm}
\subsection{Evaluation Metrics}
\textbf{Segmentation:} We employ two complementary evaluation metrics for comprehensive segmentation assessment. We use the Foreground Adjusted Rand Index (FG-ARI) to measure the quality of clustering foreground pixels into distinct object segments. Following prior work \cite{Aydemir2023NeurIPS, karazija2021clevrtex, bao2023object}, we calculate per-frame FG-ARI and report the mean across all frames for consistency with existing approaches.

Additionally, we utilize mean Intersection-over-Union (mIoU) focusing on foreground objects, which is widely accepted in segmentation literature \cite{seitzer2023dinosaur}. To ensure temporal consistency between frames, we apply Hungarian matching between predicted and ground-truth masks following standard practice \cite{perazzi2016benchmark}.\\
\textbf{Generation:} We evaluate video generation quality through a number of complementary metrics that capture different aspects of visual fidelity and perceptual quality. We employ Peak Signal-to-Noise Ratio (PSNR) to quantify pixel-wise reconstruction accuracy between generated and ground-truth frames. To assess perceptual similarity, we utilize the Structural Similarity Index (SSIM)~\cite{Wang2004TIP}, which evaluates structural information preservation, including luminance and contrast patterns.
For deeper perceptual evaluation, we incorporate Learned Perceptual Image Patch Similarity (LPIPS)~\cite{Zhang2018CVPR}, which leverages deep features to measure perceptual differences that correlate with human judgments. To evaluate distributional quality and realism, we employ Fréchet Inception Distance (FID)~\cite{heusel2017fid}, which measures feature distribution distances between real and generated images. Additionally, we utilize Fréchet Video Distance (FVD)~\cite{Unterthiner2018arXiv} to assess temporal consistency and motion quality in generated video sequences. This multi-perspective evaluation ensures a robust understanding of generation quality in both spatial and temporal dimensions. \vspace{-0.75cm}
\begin{table}[t]
    \caption{\textbf{Unsupervised video object segmentation on real-world datasets.} We compare our method with state-of-the-art approaches on YTVIS and DAVIS datasets. For fair comparison with our encoder-based approach, we include SOLV-E (encoder attention masks) and SOLV-E + M (encoder attention masks with merging), alongside the full SOLV method which uses decoder-generated masks. Our approach demonstrates strong performance across both clustering-based (FG-ARI) and overlap-based (mIoU) evaluation metrics. \vspace{-0.3cm}
    }
    \label{table:video-seg-quan}
    \centering
    \small
    \setlength{\tabcolsep}{4pt}
    \begin{tabular}{lcccc}
    \toprule
    \multirow{2}{*}{\textbf{Method}} & \multicolumn{2}{c}{\textbf{YTVIS}} & \multicolumn{2}{c}{\textbf{DAVIS}} \\
    \cmidrule(lr){2-3} \cmidrule(lr){4-5}
    & mIoU & FG-ARI & mIoU & FG-ARI \\
    \midrule
    LSD~\cite{jiang2023lsd} & 29.55 & 14.07 & 29.55 & 14.35 \\
    SlotDiffusion~\cite{wu2023slotdiffusion} & 38.33 & 15.70 & 31.27 & 12.34 \\
    SlotAdapt~\cite{akan2025slotadapt} & 36.51 & 20.32 & 29.95 & 16.28 \\
    \midrule
    SOLV-E & 32.91 & 19.30 & 31.23 & 18.89 \\
    SOLV-E + M & 36.91 & 21.34 & 33.12 & 20.40 \\ 
    SOLV~\cite{Aydemir2023NeurIPS}\footnotemark[2] & \textbf{42.01} & 21.55 & \textbf{36.62} & 20.98 \\
    \midrule
    Ours & 40.57 & \textbf{22.40} & 34.93 & \textbf{21.60} \\
    \bottomrule
    \end{tabular}
    \vspace{-0.62cm}
\end{table}

\footnotetext[2]{The original SOLV results were reported at a higher resolution ($336\times504$) with a different aspect ratio, while our experiments are conducted at $224\times224$. This resolution change partly accounts for the observed performance differences.}

\subsection{Quantitative Results}

We first analyze the contribution of individual components through ablation studies, then compare our unified framework against specialized baselines for both segmentation and generation tasks. \\
\textbf{Ablation Studies.} We systematically investigate our framework's key components and training strategies (Table~\ref{table:ablation-study}). Removal of the register aggregator~(the version where only the corresponding frame's DINO tokens are used without any transformer encoder for temporal aggregation, $\br_t$ in Eq.~\ref{eq:register_aggregator}) or replacing register tokens with slot averaging~(the augmented slots are averaged in the slot dimension, $\frac{1}{K} \sum_{i=1}^{K} \tilde{\mathbf{S}}_{t,k}^{\mathbf{+}}$ in Eq.~\ref{eq:augmented_slot}) leads to significant performance declines, underscoring the importance of these components. Using standard slot attention drastically reduces performance, highlighting the critical role of pose invariance.%

By default, we employ 1-frame training where we randomly select one frame from each video sequence, as explained in Section~\ref{sec:encoder}. As an alternative, we evaluate 5-frame training with loss applied to all frames in the video sequences. This 5-frame training strategy yields minor performance improvements over the 1-frame approach.

As stated in Section~\ref{sec:decoder}, we omit the auxiliary attention-guidance loss in our default 1-frame training. However, when using 5-frame training, the attention guidance loss can be applied, which further improves performance but significantly increases computational costs. We therefore choose the more efficient 1-frame training as our default setting. \\
\textbf{Comparison with Baselines.} Table~\ref{table:video-seg-quan} summarizes segmentation results. 
We report results for three variants of SOLV: (i) encoder-only (SOLV-E), which evaluates invariant slot attention masks similar to our architecture; (ii) encoder + merging (SOLV-E+M), which applies merging to the invariant slot attention masks; and (iii) full decoder-based masks (SOLV), the default version that uses masks from the spatial broadcast decoder.

The comparison with SOLV variants reveals an important trade-off in existing approaches. While SOLV's decoder masks achieve the highest mIoU scores, this comes at a cost: the decoder masks cannot be used for video synthesis, making the method specialized for segmentation only. When SOLV operates in configurations comparable to our approach, using encoder attention masks, its performance drops across both metrics, falling below our unified method on all metrics. 
\begin{figure}
    \centering
    \includegraphics[width=\linewidth]{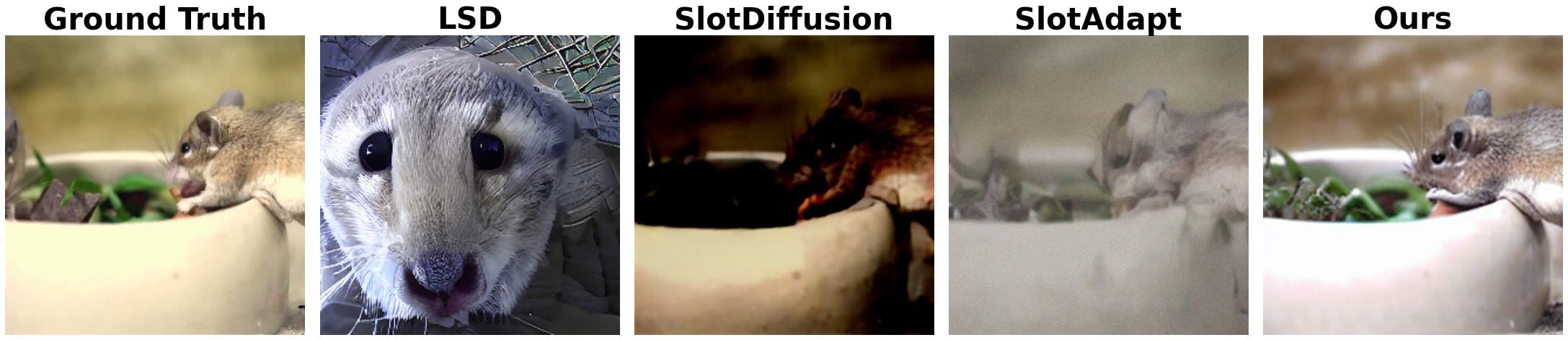}
    \includegraphics[width=\linewidth]{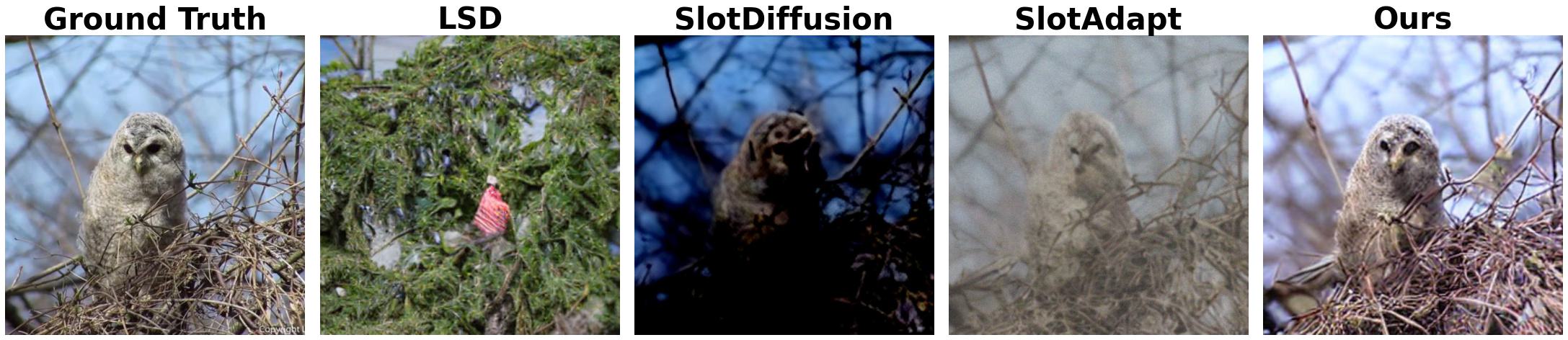}
    \includegraphics[width=\linewidth]{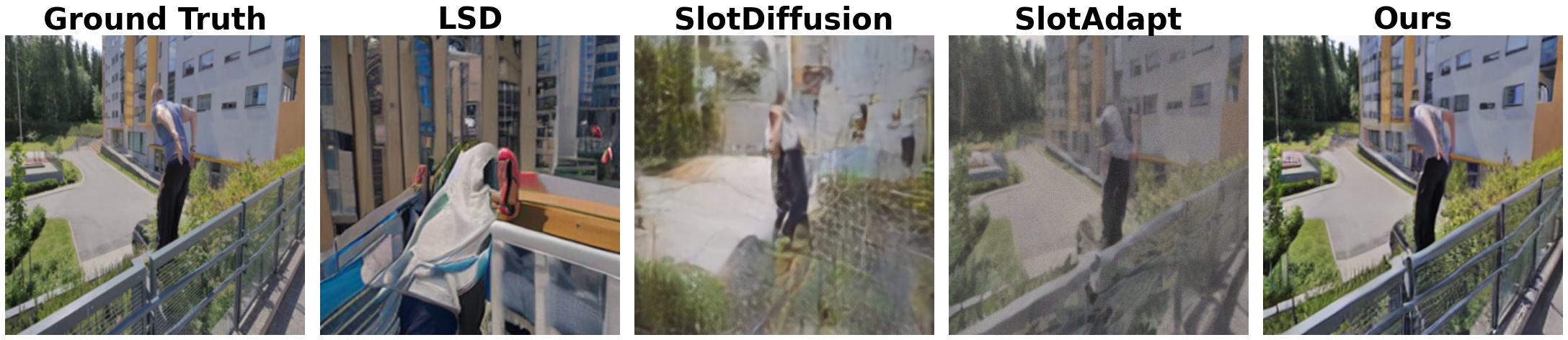}
    \includegraphics[width=\linewidth]{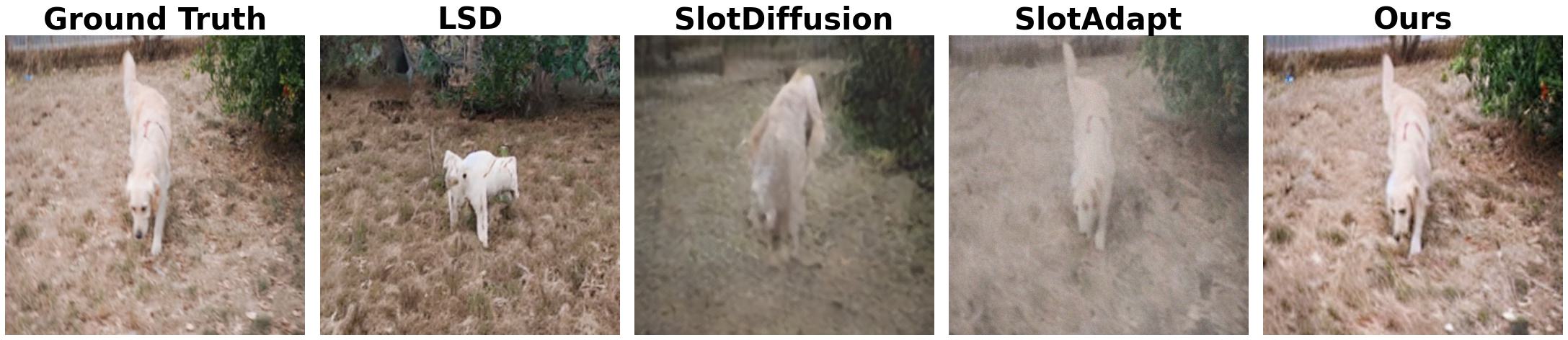}\vspace{-0.5cm}
    \caption{\textbf{Generation Results.} Visual comparison of video generation quality across methods on YTVIS (rows 1-2) and DAVIS17 (rows 3-4) datasets. Our method generates high-quality frames that closely match the ground truth, maintaining sharp object boundaries and preserving fine-grained textures. The results demonstrate faithful reconstruction of original scenes with superior detail preservation in the small animal's features (row 1), natural appearance and accurate coloring of the bird (row 2), clear structural elements in the urban scene (row 3), and realistic texture and form of the white animal (row 4). Full temporal sequences are provided in the Supplementary Material.}
    \label{fig:gen-comparison}\vspace{-0.65cm}
\end{figure}
Importantly, our 5-frame training nearly matches the mIoU of existing methods like SOLV, indicating its potential for further improvements, as shown in Table~\ref{table:ablation-study}. However, we choose the 1-frame training configuration to balance performance with computational efficiency, representing an optimal trade-off for practical applications.\\
\textbf{Video Generation Performance.} Table~\ref{table:video-gen-quan} presents comprehensive video generation results compared to baselines. Our approach establishes new state-of-the-art results on YTVIS and DAVIS17 datasets, achieving superior performance across all complementary metrics, spanning pixel-level fidelity, structural preservation, perceptual quality, and temporal consistency.
\begin{table*}[t]
    \caption{\textbf{Video generation performance on real-world datasets.} We evaluate our method against state-of-the-art approaches on YTVIS and DAVIS datasets using comprehensive generation metrics. Our approach demonstrates superior performance across both pixel-level accuracy (PSNR, SSIM), perceptual quality measures (LPIPS, FID) and temporal coherence (FVD).\vspace{-0.3cm}}
    \label{table:video-gen-quan}
    \centering
    \small
    \setlength{\tabcolsep}{6pt}
    \begin{tabular}{lccccc|ccccc}
    \toprule
    \multirow{2}{*}{\textbf{Method}} & \multicolumn{5}{c}{\textbf{YTVIS}} & \multicolumn{5}{c}{\textbf{DAVIS}} \\
    \cmidrule(lr){2-6} \cmidrule(lr){7-11}
    & PSNR$\uparrow$ & SSIM$\uparrow$ & LPIPS$\downarrow$ & FID$\downarrow$ & FVD$\downarrow$ & PSNR$\uparrow$ & SSIM$\uparrow$ & LPIPS$\downarrow$ & FID$\downarrow$ & FVD$\downarrow$ \\
    \midrule
    LSD~\cite{jiang2023lsd} & 9.64 & 0.2793 & 0.777 & 100.68 & 121.41 & 9.58 & 0.0356 & 0.6079 & 84.30 & 75.05 \\
    SlotDiffusion~\cite{wu2023slotdiffusion} & 9.18 & 0.1867 & 0.6484 & 86.38 & 123.8413 & 10.68 & 0.0340 & 0.6143 & 136.18 & 152.73 \\
    SlotAdapt~\cite{akan2025slotadapt} & 10.92 & 0.3669 & 0.6556 & 65.30 & 63.72 & 12.18 & 0.0674 & 0.2681 & 41.94 & 29.96 \\
    \midrule
    Ours & \textbf{11.37} & \textbf{0.3933} & \textbf{0.5908} & \textbf{49.51} & \textbf{51.77} & \textbf{12.38} & \textbf{0.0946} & \textbf{0.1886} & \textbf{28.43} & \textbf{16.17} \\
    \bottomrule
    \end{tabular}
    \vspace{-0.5cm}
\end{table*}

\begin{figure}
    \centering
    \includegraphics[width=\linewidth]{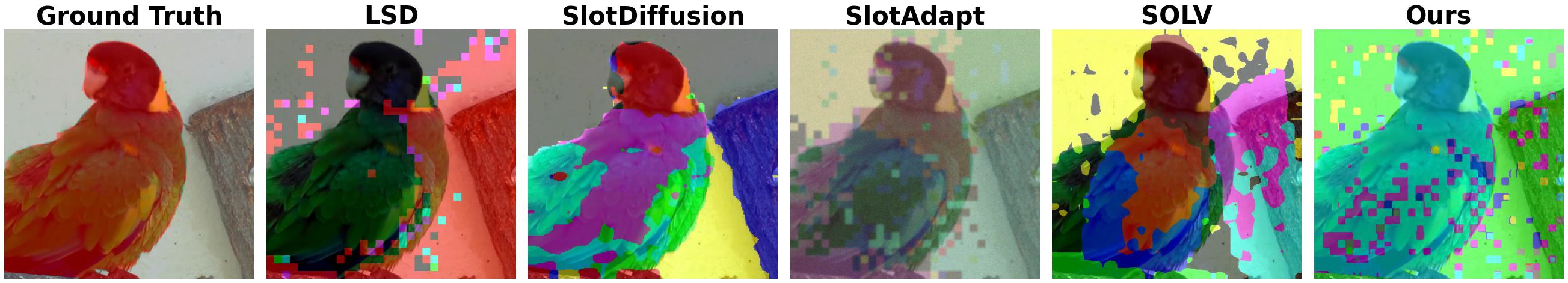}
    \includegraphics[width=\linewidth]{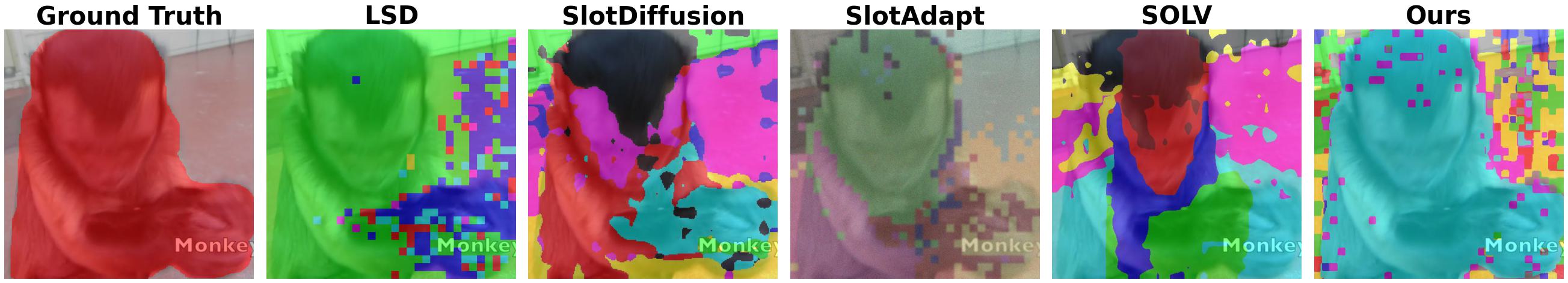}
    \includegraphics[width=\linewidth]{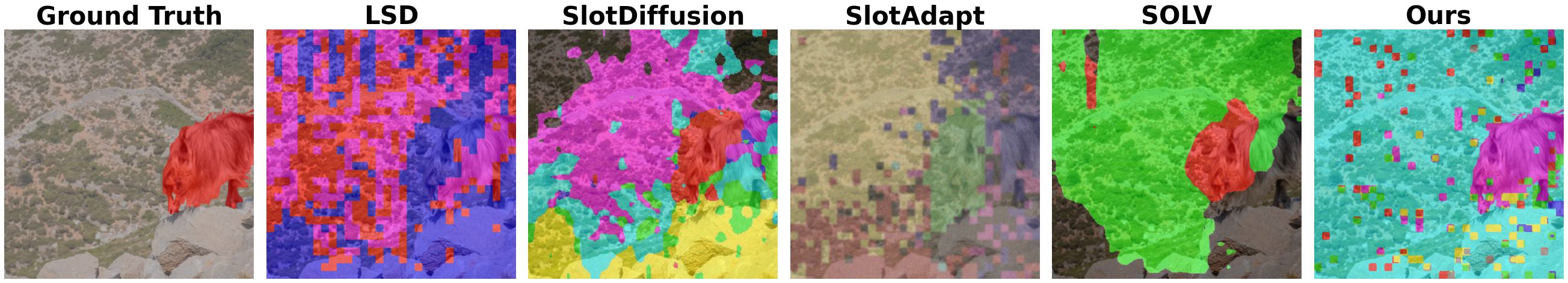}
    \includegraphics[width=\linewidth]{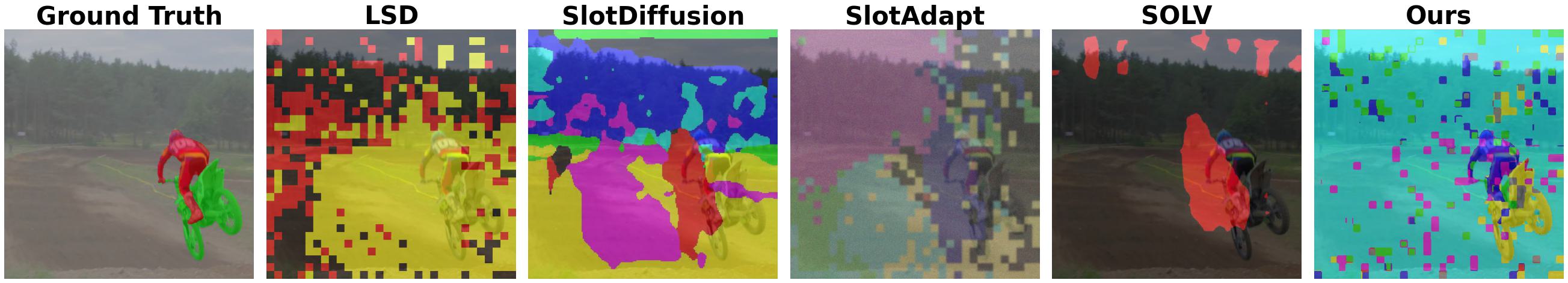}\vspace{-0.5cm}
    \caption{\textbf{Segmentation Results.} Qualitative comparison of video object segmentation on YTVIS (rows 1-2) and DAVIS17 (rows 3-4). Our method successfully delineates objects with accurate boundaries across diverse challenging scenarios. Row 1 shows segmentation of a bird with detailed boundary preservation, row 2 demonstrates segmentation of a monkey that covers most of the frame, row 3 shows segmentation of an animal against a challenging natural background, and row 4 presents segmentation of closely touching objects (motorcycle and person) that are difficult to separate. Different colors represent distinct object instances discovered by each method. Temporal consistency across frames is demonstrated in the Supplementary Material.}
    \label{fig:seg-comparison}\vspace{-0.65cm}
\end{figure}
\begin{figure*}[ht]
    \centering
    \includegraphics[width=\linewidth]{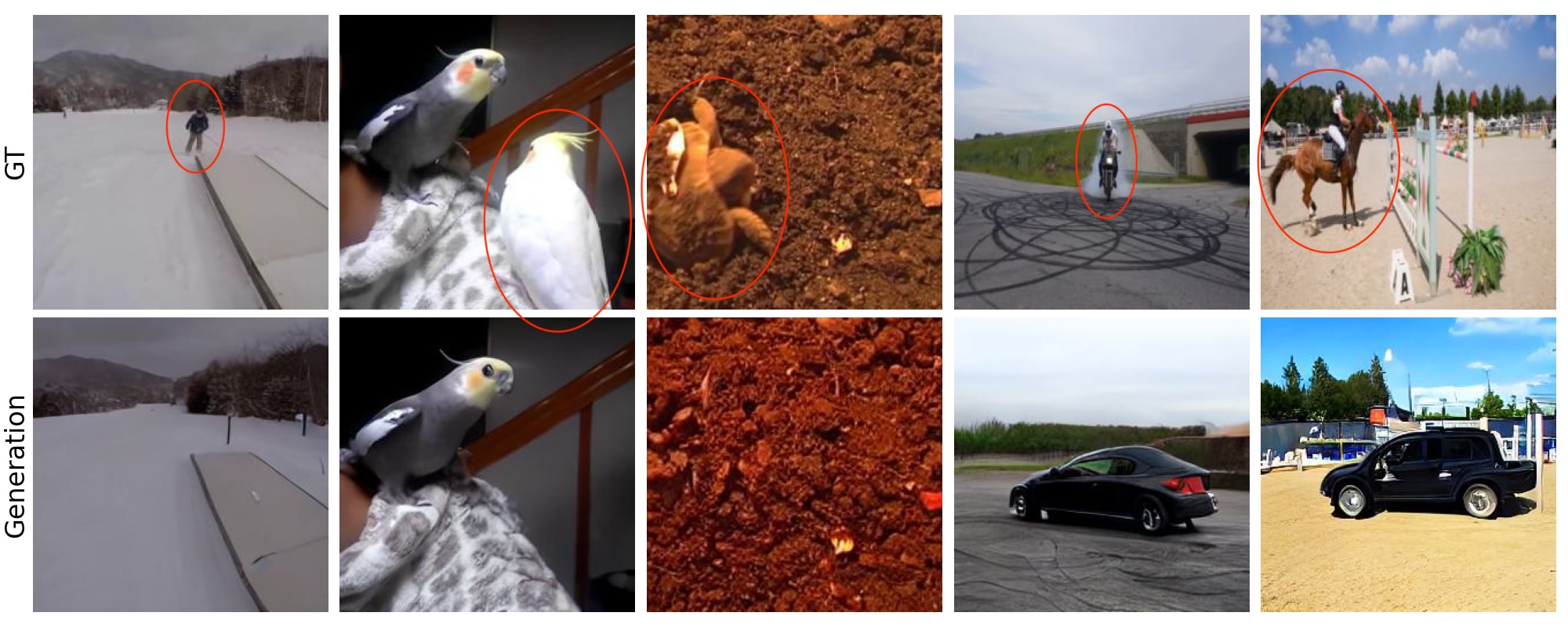}\vspace{-0.8cm}
    \caption{\textbf{Compositional Generation Results.} Demonstration of object-level editing capabilities through deletion and replacement operations on YTVIS (columns 1-3) and DAVIS17 (columns 4-5). Top rows show ground truth frames; bottom rows display edited results. Our method handles challenging scenarios including: (a) removal of closely positioned objects while preserving scene coherence (column 1, bird deletion), (b) deletion of camouflaged objects with natural background inpainting (column 3, turtle removal), and (c) semantic object replacement maintaining proper occlusion and lighting (columns 4-5). The edited videos maintain temporal consistency throughout the sequences. Full temporal sequences are provided in the Supplementary Material.}
    \label{fig:comp-comparison}\vspace{-0.65cm}
\end{figure*}

The consistent performance across traditional pixel-based metrics (PSNR, SSIM) and more recent perceptual measures (LPIPS, FID) is particularly notable, as these metrics often exhibit trade-offs in conventional approaches. Our unified framework's ability to simultaneously optimize for reconstruction accuracy and perceptual realism indicates a fundamental advancement in video generation quality.

The results on DAVIS17 show consistent performance across all metrics. Our method achieves lower LPIPS and FID scores compared to baselines, indicating improved perceptual quality and better distributional alignment with real video content. These improvements in perceptual metrics complement the gains observed in pixel-level measures, demonstrating the effectiveness of our approach across different evaluation criteria.

These results establish that high-quality segmentation and generation can be achieved within a unified architecture, demonstrating significant advantages over methods that specialize in single tasks. The consistent performance across both clustering-based and overlap-based segmentation metrics, combined with improvements across pixel-level fidelity and perceptual quality measures, validates our core hypothesis that temporally consistent slot representations can effectively condition temporally coherent video generation while maintaining competitive segmentation capabilities. \vspace{-0.32cm}
\begin{table}[t]
    \caption{\textbf{Compositional generation performance.} We evaluate compositional generation by mixing slots from different batch samples. Our method demonstrates superior performance across both datasets.\vspace{-0.3cm}}
    \label{table:comp-gen-quan}
    \centering
    \small
    \setlength{\tabcolsep}{3pt}
    \begin{tabular}{lcccc}
    \toprule
    \textbf{Method} & PSNR$\uparrow$ & SSIM$\uparrow$ & LPIPS$\downarrow$ & FID$\downarrow$ \\
    \midrule
    \multicolumn{5}{c}{\textit{YTVIS}} \\
    \midrule
    LSD & 8.05 & 0.0527 & 0.8734 & 127.49 \\
    SlotDiffusion & 7.83 & 0.0521 & 0.9486 & 129.33 \\
    SlotAdapt & 9.49 & 0.0575 & 0.7313 & 112.79 \\
    Ours & \textbf{10.07} & \textbf{0.0687} & \textbf{0.629} & \textbf{83.45} \\
    \midrule
    \multicolumn{5}{c}{\textit{DAVIS}} \\
    \midrule
    LSD & 5.22 & 0.028 & 1.0234 & 198.23 \\
    SlotDiffusion & 6.04 & 0.031 & 0.9912 & 187.13 \\
    SlotAdapt & 6.93 & 0.032 & 0.9721 & 172.39 \\
    Ours & \textbf{8.27} & \textbf{0.065} & \textbf{0.694} & \textbf{113.86} \\
    \bottomrule
    \end{tabular}
    \vspace{-0.65cm}
\end{table}

\subsection{Qualitative Results} 
\textbf{Segmentation Quality.} Figure~\ref{fig:seg-comparison} presents visual comparisons of segmentation results across different methods on challenging video sequences. In these examples, our method successfully separates different objects with clear boundary delineation. For instance, in the bottom image of Figure~\ref{fig:seg-comparison}, our approach effectively differentiates individual object instances where other methods struggle to maintain distinct segmentations.\\
\textbf{Generation Performance.} Figure~\ref{fig:gen-comparison} compares video generation results across different methods. Our method generates higher quality frames with better detail preservation and cleaner object boundaries compared to baseline approaches. The visual comparison shows our approach maintains object identity and spatial relationships effectively while producing temporally consistent video content. For multi-frame visualizations, please refer to the Supplementary Material. \\
\textbf{Video Editing Capabilities.} 
Figure~\ref{fig:comp-comparison} presents examples of our framework's compositional video editing capabilities. Thanks to a unified architecture, our method allows intuitive operations such as inserting, removing, or replacing objects, while preserving photorealistic detail. These results demonstrate the practical value of our object-centric representation for enabling flexible and interactive video editing. Additional compositional editing results are provided in the Supplementary Material.

To assess compositional generation quantitatively, we follow the setup introduced in SlotDiffusion~\cite{wu2023slotdiffusion} and extended in our previous work, SlotAdapt~\cite{akan2025slotadapt}. SlotDiffusion evaluates compositionality by randomly mixing slot representations from different images within a batch. We adapt this idea to the video domain by first aligning slot correspondences across frames, then randomly exchanging slots between videos in the batch, frame by frame.

As shown in Table~\ref{table:comp-gen-quan}, our method consistently outperforms prior approaches on this task. Taken together with the qualitative results in Figure~\ref{fig:comp-comparison}, these findings confirm that our model can effectively handle compositional video generation and editing, both in terms of visual quality and quantitative performance.\vspace{-0.27cm}

\section{Conclusion}
This work introduces the first unified framework for simultaneous video object segmentation and compositional video generation, challenging the conventional separation of these fundamental tasks. Our approach demonstrates that object-centric representations can effectively bridge perception and synthesis, achieving state-of-the-art FG-ARI performance on YTVIS and DAVIS17 datasets while establishing new benchmarks across all video generation metrics.

Our core insight is that generative modeling provides inductive structure beneficial for segmentation, while object-centric decomposition offers strong priors for temporally coherent synthesis. This synergy allows our model to outperform task-specific baselines without relying on handcrafted temporal cues or architectural constraints. Extensive experiments confirm that integrating structured slot representations with pretrained diffusion models yields consistent improvements in temporal stability and visual~fidelity.

The results demonstrate that structural commonalities between perception and generation can be exploited for mutual benefit. Strong clustering accuracy reflects improved temporal modeling, while generation metrics confirm the model’s ability to synthesize content that respects both structural and semantic~constraints.

Future work will explore replacing the image-based decoder with dedicated video diffusion models for end-to-end temporal dynamics modeling. Additional directions include scaling to higher resolutions and longer sequences, and incorporating text-based supervision for language-guided generation and editing. \vspace{-0.45cm}

{\fontsize{8.9pt}{11pt}\selectfont
\bibliographystyle{plain}
\bibliography{references}
}

\begin{IEEEbiography}[{\includegraphics[width=1in,height=1.25in,clip,keepaspectratio]{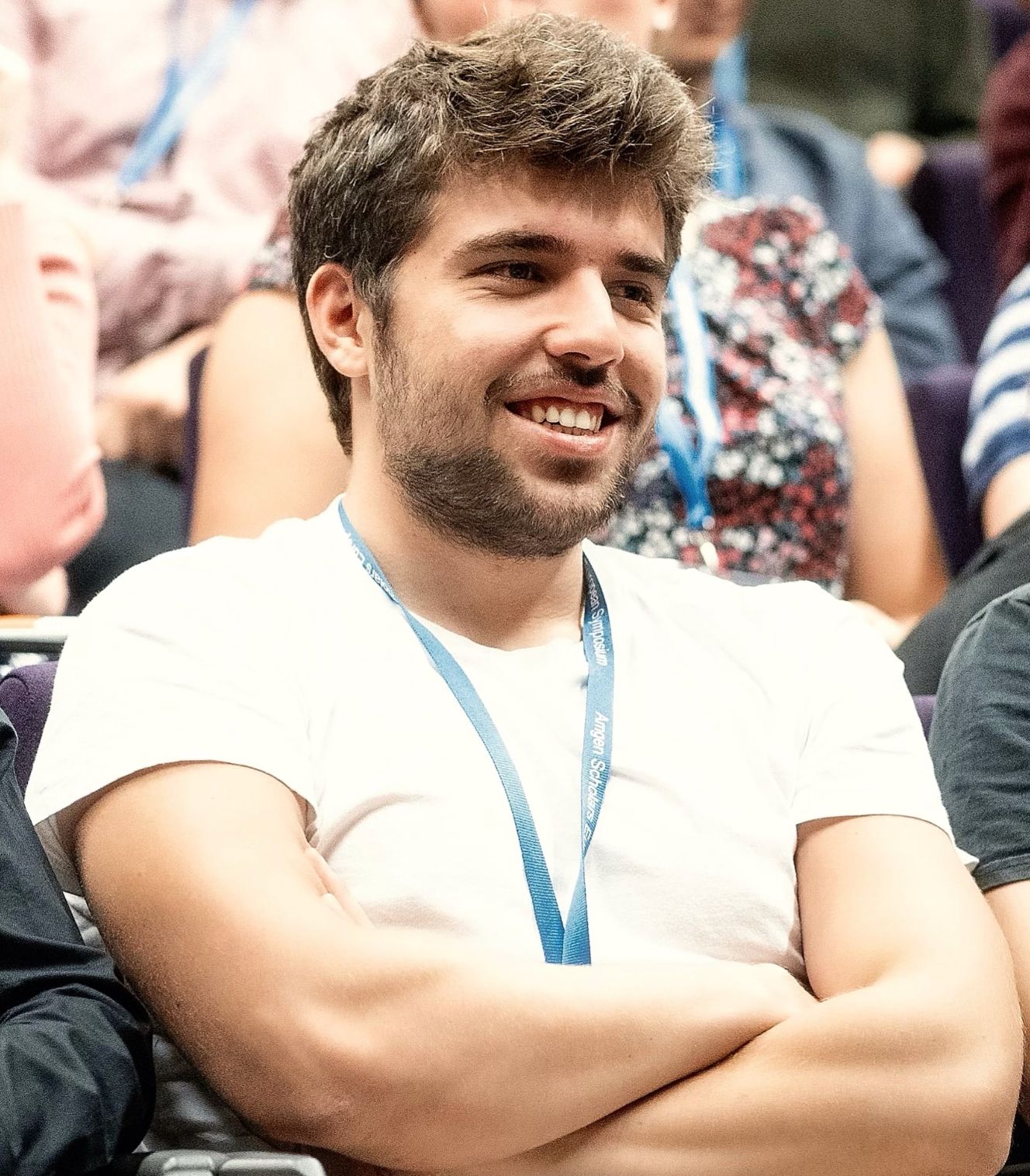}}]{Adil Kaan Akan} received the B.Sc. degree in computer engineering from Middle East Technical University, Ankara, Turkey, in 2020 and received the M.Sc. degree in computer science from Koc University, Istanbul, Turkey, in 2022, where he was awarded the Academic Excellence Award. He is currently a Ph.D. candidate at Koc University, supervised by Prof. Yucel Yemez. His research interests include object-centric learning, generative models, compositional image and video synthesis. His recent work explores the integration of object-centric representations with diffusion models for controllable visual content generation. \end{IEEEbiography}

\begin{IEEEbiography}[{\includegraphics[width=1in,height=1.25in,clip,keepaspectratio]{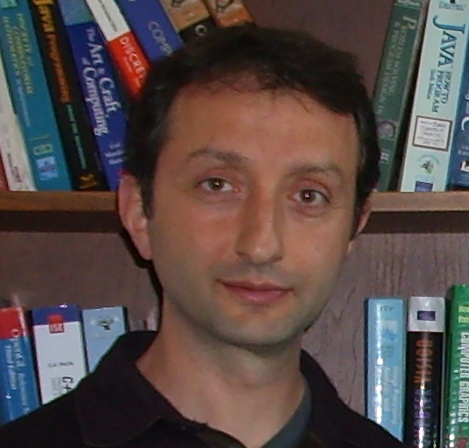}}]{Yucel Yemez} received the B.S. degree from Middle East Technical University, Ankara, in 1989, and the M.S. and Ph.D. degrees from Boğaziçi University, Istanbul, respectively, in 1992 and 1997, all in electrical engineering. From 1997 to 2000, he was a postdoctoral researcher in the Image and Signal Processing Department of Télécom Paris (ENST). Currently, he is a Professor in the Computer Engineering Department at Koç University, Istanbul and member of the KUIS AI Center at Koç University. His research interests include computer vision, deep learning, computer graphics, and multimedia signal processing. He has published extensively in leading journals and conferences and served  as an Associate Editor for Graphical Models (Elsevier) and The Visual Computer (Springer).\end{IEEEbiography}

\newpage

\begin{appendices}
\appendices
\section*{Appendix}

This supplementary material includes additional experimental details covering datasets, model configurations, and training procedures~(Section \ref{supp:sec:exp_detail}),
elaborates on the Invariant Slot Attention mechanism introduced in the main paper~(Section \ref{supp:sec:isa}),
and provides extended qualitative visualizations~(Section  \ref{supp:sec:vis}).

\section{Experimental Details}
\label{supp:sec:exp_detail}

\subsection{Dataset Details}
To ensure consistency across inputs, we removed the black borders present in all videos from the YTVIS19 dataset.
Given the self-supervised nature of our approach, we combine the standard dataset splits during training.
For evaluation, we use the publicly available validation sets for all datasets except YTVIS.
As the YTVIS dataset does not provide annotations for its validation split, we use the exact same subset of 300 videos from the training set that are selected by SOLV~\cite{Aydemir2023NeurIPS}.
During evaluation, we upsample the predicted segmentation masks to the original frame resolution using bilinear interpolation.

\subsection{Common Experimental Setup}
Unless stated otherwise, all experiments use the ViT-B/14 architecture pretrained with DINOv2~\cite{oquab2023dinov2} as the visual backbone, 7 slots with dimension 768, temporal window size $L=5$ consecutive frames~(corresponds to 2 past, 1 present, 2 future), and input resolution of $256 \times 256$ for Stable Diffusion VAE and $224 \times 224$ for the DINO encoder.

\boldparagraph{Training Schedule} We train models for 350K iterations on YTVIS19, then fine-tune for 50K iterations on DAVIS17. All models use AdamW optimizer~\cite{loshchilov2018adamw} with batch size 8, gradient clipping at 0.5, and linear warmup for the first 5K iterations on $2\times$A40 GPUs.

\subsection{Model Details} 

\boldparagraph{Feature Extractor}
We use the output of the final transformer block from DINOv2 ViT-B/14~\cite{oquab2023dinov2}, excluding the classification (CLS) token, with positional embeddings added after feature extraction.

\boldparagraph{Invariant Slot Attention} The input dimension $D_\text{slot}=768$ is used throughout the ISA architecture. After positional encoding addition to DINO tokens $\mathbf{F}_t$, slots and projected tokens are passed to ISA. Slots are updated with a GRU cell, followed by a residual MLP with layer normalization. All projection layers $(p, q, k, v, g)$ have dimension $D_\text{slot}$. GRU is iterated three times per frame. The scale parameter $\mathbf{s}_s$ is multiplied by $\delta = 5$ following~\cite{Aydemir2023NeurIPS}.

We initialize the components as follows: $\mathbf{G}_\text{abs}$ is initialized as a coordinate grid in $[-1, 1]$, slots $\mathbf{S}$ are initialized using Xavier initialization~\cite{glorot2010init}, and slot scale $\mathbf{s}_s$ and position $\mathbf{s}_p$ are initialized from a normal distribution~\cite{Aydemir2023NeurIPS}.

\boldparagraph{Temporal Aggregator}A 3-layer, 8-head transformer encoder~\cite{vaswani2017attention} with hidden dimension $4 \times D_\text{slot}$ and learnable temporal positional embeddings initialized from a normal distribution. Unavailable frame slots (indices $<$ 0 or $>$ frame count) are masked. Unlike SOLV~\cite{Aydemir2023NeurIPS}, all slots attend to one another across frames, rather than restricting same-index slots to interact only with their corresponding slots in other frames.

\boldparagraph{Temporal Register Aggregator}
A 1-layer, 8-head transformer encoder with hidden dimension $4 \times D_\text{slot}$. DINO features are spatially pooled before transformer input. Learnable temporal positional embeddings are initialized from a normal distribution, and unavailable frame tokens are masked.

\boldparagraph{Decoder} 
Adapter-injected Stable Diffusion 1.5~\cite{rombach2022ldm} with frozen pretrained parameters.

\subsection{Baselines}

All image-based baselines (LSD~\cite{jiang2023lsd}, SlotDiffusion~\cite{wu2023slotdiffusion}, SlotAdapt~\cite{akan2025slotadapt}) are trained on flattened video frames as independent images using publicly available code implementations. To ensure fair comparison, all baselines use DINOv2~\cite{oquab2023dinov2} as the image encoder, 7 slots with dimension 768, and follow the same training schedule of 350K iterations on YTVIS19 followed by 50K iterations fine-tuning on DAVIS17. For models requiring pretrained diffusion components, we use Stable Diffusion 1.5~\cite{rombach2022ldm}.

SOLV~\cite{Aydemir2023NeurIPS} is trained on 5-frame sequences following the same temporal setup as our method.

\section{Invariant Slot Attention}
\label{supp:sec:isa}
This section elaborates on the mechanics of invariant slot attention (ISA), originally introduced by Biza \etal~\cite{biza2023invariant}.
In our architecture, ISA is employed within the Object-Centric Temporal Encoding module~(Section \ref{sec:encoder}), making use of shared initialization as outlined in the main paper.
Starting from the common initialization $\cZ_t = \{\left(\mathbf{z}_t^j, \mathbf{s}^j_s, \mathbf{s}^j_p, \mathbf{G}_{\text{abs}, t}\right)\}_{j=1}^K$,
where $\mathbf{z}_t^j$ is the $j$-th slot representation, $\mathbf{s}^j_s$ denotes the scale parameters along $x$ and $y$ axes, likewise, $\mathbf{s}^j_p$ represents the position parameters for $x$ and $y$ axes, and $\mathbf{G}_{\text{abs}, t} \in \mathbb{R}^{N \times 2}$ is the absolute coordinate grid at time $t$,
our objective is to update the set of slots $\{\mathbf{z}_t^j\}_{j=1}^K$.
To clarify, we describe here the procedure for a single time step $t$ in the invariant slot attention mechanism:
\begin{align}
\label{eq:inv_slot_att_sup}
    &\mathbf{A}_t := \underset{j=1,\dots,K}{\mathrm{softmax}} \left(\mathbf{M}_t\right) \in \mathbb{R}^{K\times N}, \\
    &\mathbf{m}_t^j := \dfrac{1}{\sqrt{d}}\ \  p\left(k\left(\mathbf{F}_t\right) + g(\mathbf{G}^j_{\text{rel}, t})\right) q(\mathbf{z}^j_t) \in \mathbb{R}^{N}
\end{align}
Here, $q: \mathbb{R}^{D_\text{slot}} \to \mathbb{R}^{D_\text{slot}}$, $k: \mathbb{R}^{d} \to \mathbb{R}^{D_\text{slot}}$, $p: \mathbb{R}^{D_\text{slot}} \to \mathbb{R}^{D_\text{slot}}$, and $g: \mathbb{R}^{2} \to \mathbb{R}^{D_\text{slot}}$ are learnable linear transformations applied to each patch and slot vector independently. The vector $\mathbf{m}_t^j$ represents the $j$-th row of the unnormalized attention score matrix $\mathbf{M}_t \in \mathbb{R}^{K \times N}$, computing the affinity between all $N$ spatial locations and slot $j$. The softmax operation is applied column-wise over the $K$ slots, yielding the attention matrix $\mathbf{A}_t \in \mathbb{R}^{K \times N}$, where $\mathbf{a}_t^j$ denotes the $j$-th row containing the normalized attention weights for slot $j$ over all spatial locations.
The relative coordinate grid associated with each slot is computed as follows:
\begin{equation}
\mathbf{G}^j_{\text{rel}, t} := (\mathbf{G}_{\text{abs}, t} - \mathbf{s}^j_p) \oslash \mathbf{s}^j_s \in \mathbb{R}^{N \times 2}
\end{equation}
where $\oslash$ corresponds to element-wise division.
The attention weights $\mathbf{a}_t^j$ computed via \eqref{eq:inv_slot_att_sup} are used to infer both the position $\mathbf{s}_p^j$ and scale $\mathbf{s}_s^j$ of the slots, according to the formulation in Biza \etal~\cite{biza2023invariant}:
\begin{align}
\mathbf{s}^j_s &:= \sqrt{\dfrac{(\mathbf{G}_{\text{abs}, t}^T - \mathbf{s}^j_p \mathbf{1}_N)^2 \mathbf{a}_t^j}{\sum_{i=1}^N \mathbf{a}_t^j[i]}} \in \mathbb{R}^2, \label{eq:14}\\
\mathbf{s}^j_p &:= \dfrac{\mathbf{G}_{\text{abs}, t}^T \mathbf{a}_t^j}{\sum_{i=1}^N \mathbf{a}_t^j[i]} \in \mathbb{R}^2 \label{eq:15}
\end{align}
where the $\sqrt{\cdot}$ and $(\cdot)^2$ operations are performed element-wise, and $\mathbf{1}_N$ is the broadcast operator that replicates the vector to match the spatial dimension~(all ones row vector of dimension N). Following the attention computation, features are aggregated using a weighted combination guided by $\mathbf{w}^j$ and projections $v: \mathbb{R}^{d} \to \mathbb{R}^{D_\text{slot}}$ and $g: \mathbb{R}^{2} \to \mathbb{R}^{D_\text{slot}}$ applied to each patch vector independently, similarly to the original slot attention mechanism:
\begin{align}
\label{eq:slot_updates}
&\mathbf{u}_t^j := p\left(v\left(\mathbf{F}_t\right) + g(\mathbf{G}^j_{\text{rel}, t})\right)^T\!\! \mathbf{w}_t^j \in \mathbb{R}^{D_\text{slot}}, \\
&\mathbf{w}_t^j:= \dfrac{\mathbf{a}_t^j}{\sum_{i=1}^N \mathbf{a}_t^j[i]} \in \mathbb{R}^{N}
\end{align}
Here, $\mathbf{u}_t^j$ represents the aggregated features for updating slot $j$, and the vectors $\{\mathbf{u}_t^j\}_{j=1}^K$ form the columns of the full update matrix $\mathbf{U}_t \in \mathbb{R}^{D_\text{slot} \times K}$.
The aggregated representations $\mathbf{u}_t^j$ from \eqref{eq:slot_updates} are then used to update the slot vectors $\{\mathbf{z}_t^j\}_{j=1}^K$ via a GRU module, followed by an MLP-based residual pathway as described in \eqref{eq:last_eq}. This process is repeated over three iterative refinement steps:
\begin{align}
\label{eq:last_eq}
&\mathbf{z}_t^j := \mathrm{GRU}\left(\mathbf{z}_t^j, \mathbf{u}_t^j\right) \in \mathbb{R}^{D_\text{slot}}, \\
&\mathbf{z}_t^j := \mathbf{z}_t^j + \mathrm{MLP}\left(\mathrm{LayerNorm}(\mathbf{z}_t^j)\right)
\end{align}

\subsection*{Experimental Validation}
\begin{figure*}
    \centering
    \includegraphics[width=\linewidth]{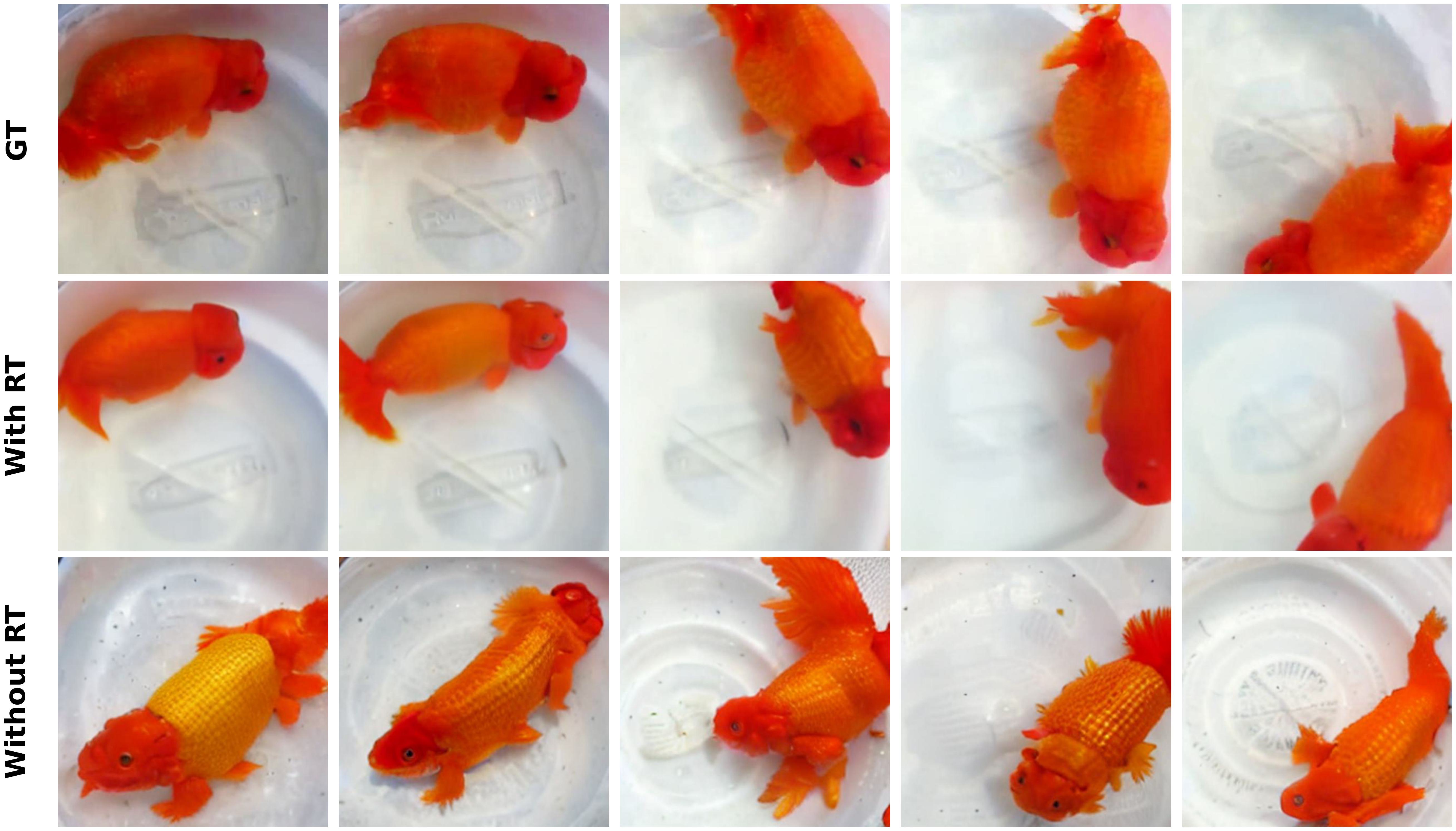}
    \caption{Temporal video generation results with and without register tokens on YTVIS dataset. \textbf{Without register tokens} (bottom), objects appear in incorrect positions and backgrounds deviate from ground truth (top). \textbf{With register tokens} (middle), generations accurately match ground truth positioning.}
    \label{fig:reg_token_vid_0}
\end{figure*}

\begin{figure*}
    \centering
    \includegraphics[width=\linewidth]{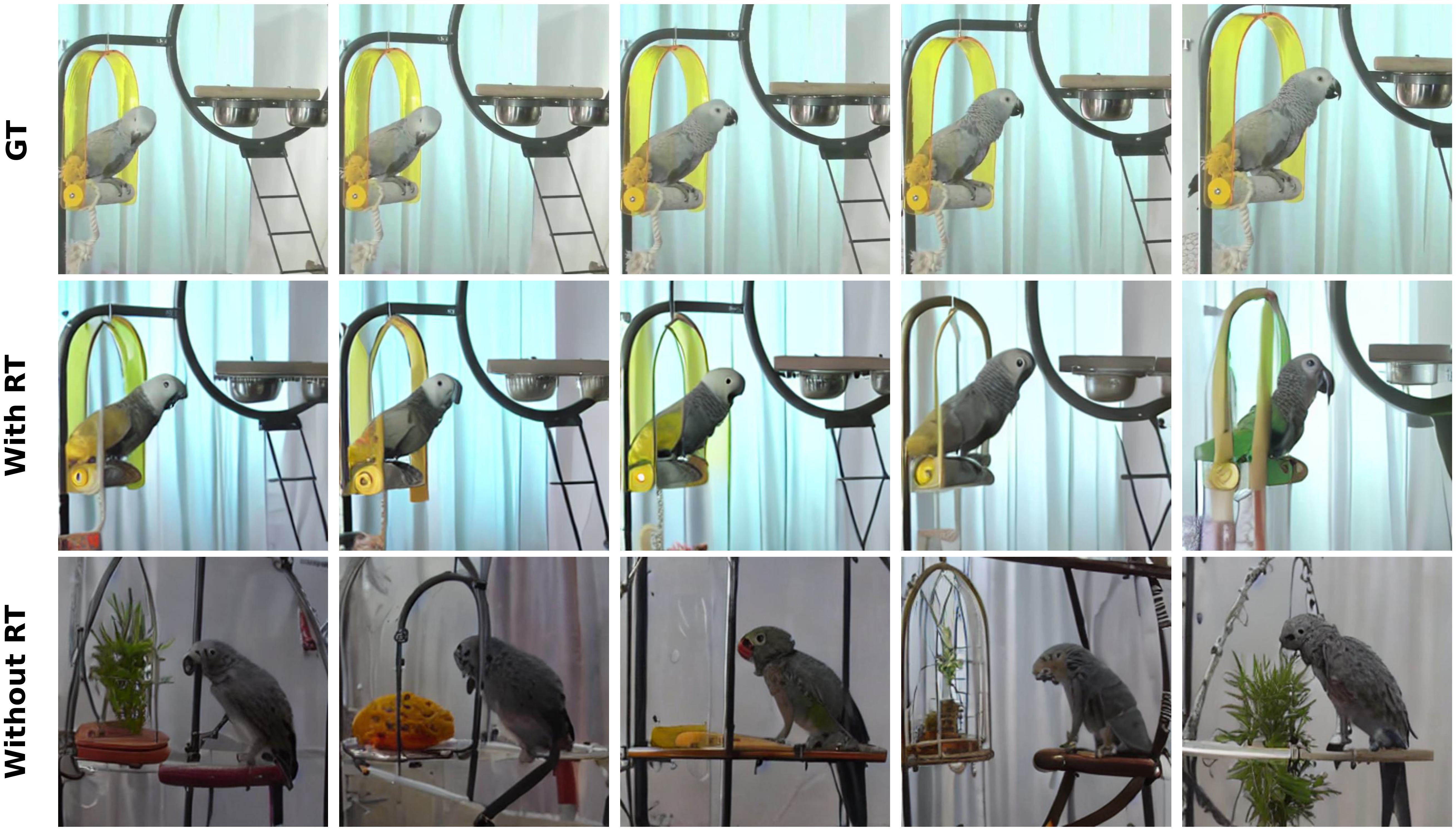}
    \caption{Temporal video generation results with and without register tokens on YTVIS dataset. Results without register tokens (bottom) lead to spatial inconsistencies across frames. Results with register tokens (middle) maintain consistent object positioning relative to ground truth (top).}
    \label{fig:reg_token_vid_22}
\end{figure*}

\begin{figure*}
    \centering
    \includegraphics[width=\linewidth]{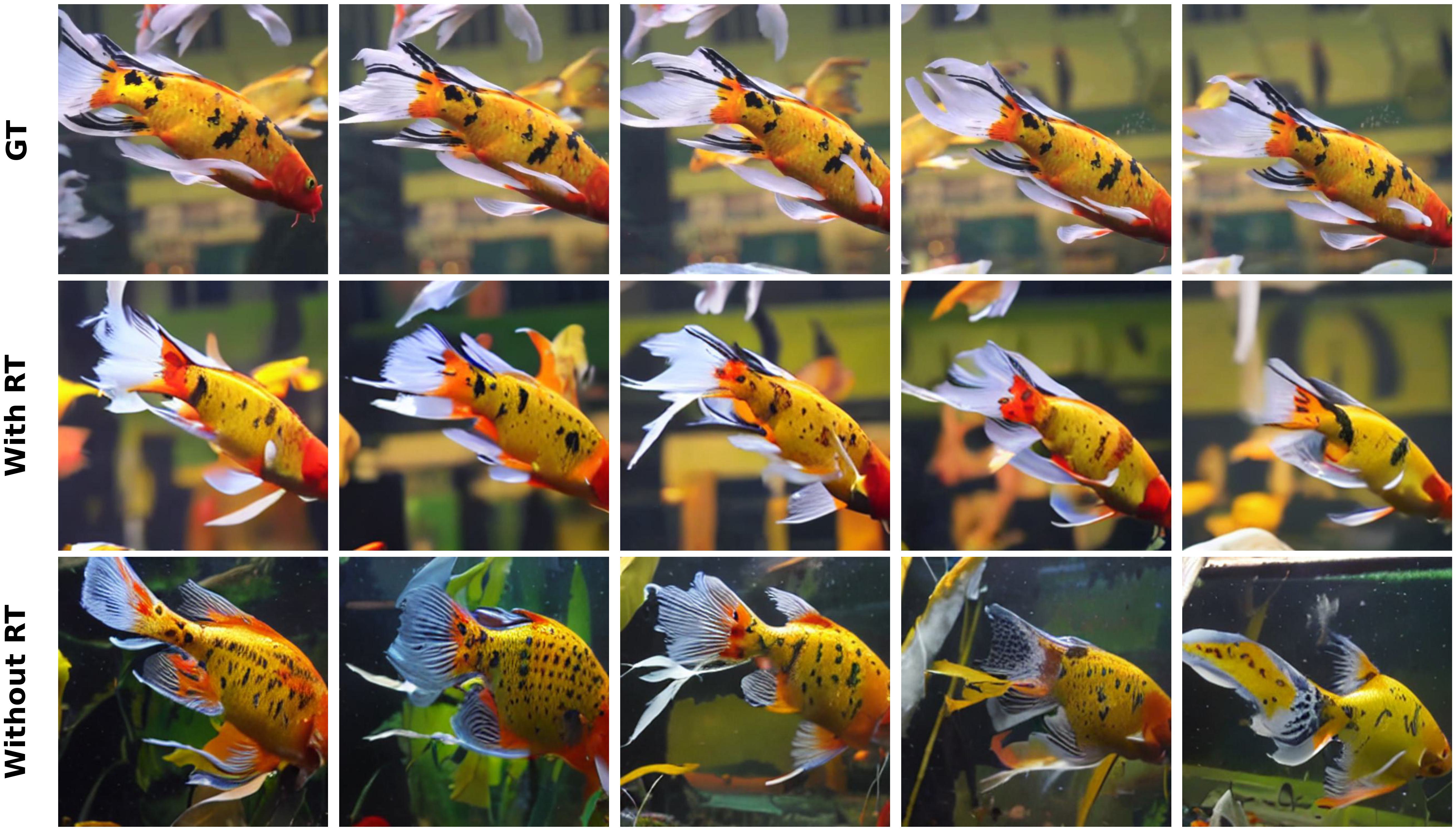}
    \caption{Temporal video generation results with and without register tokens on YTVIS dataset. Results without register tokens (bottom) lead to spatial inconsistencies across frames. Results with register tokens (middle) maintain consistent object positioning relative to ground truth (top).}
    \label{fig:reg_token_vid_24}
\end{figure*}

\begin{figure*}
    \centering
    \includegraphics[width=\linewidth]{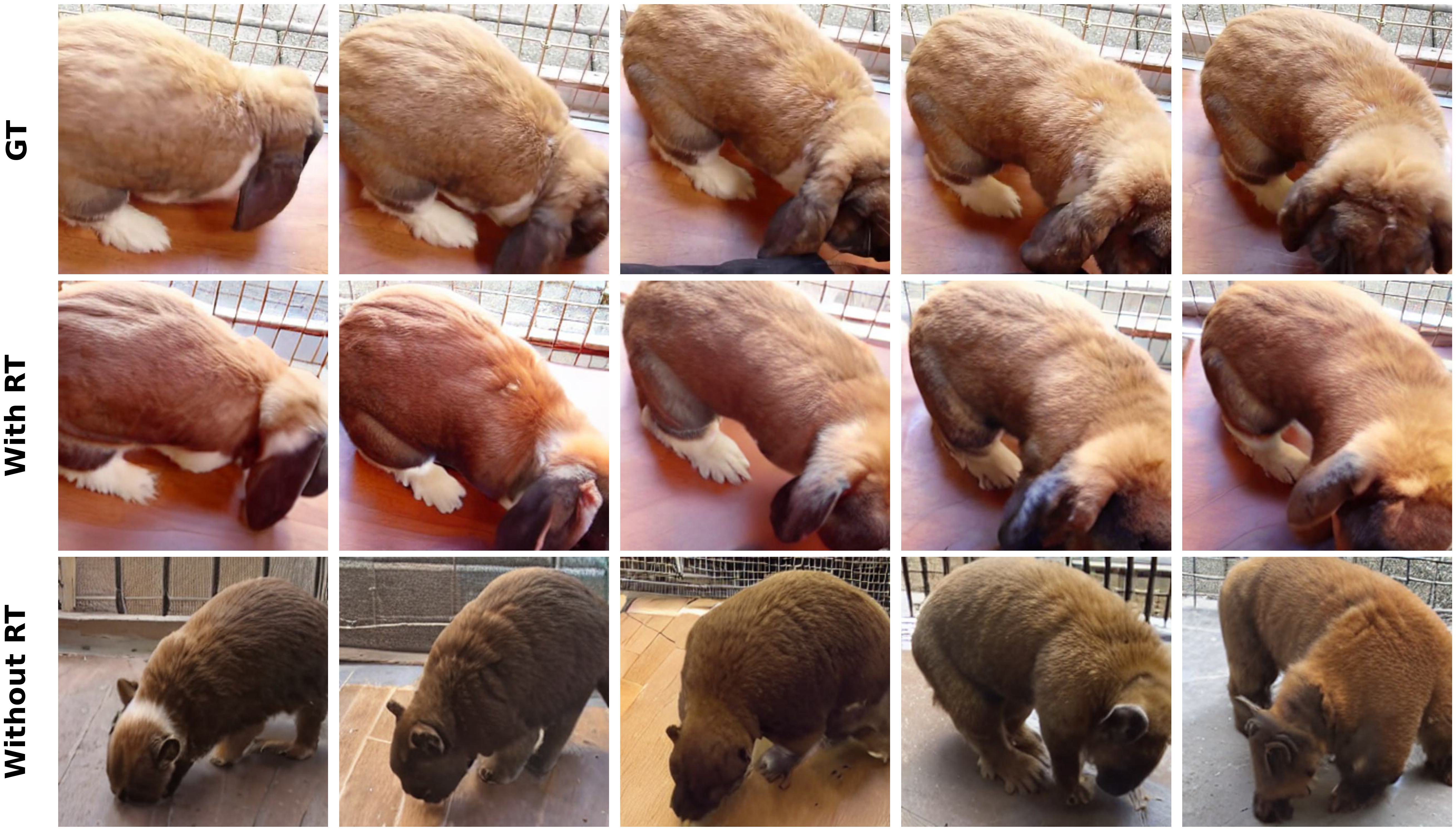}
    \caption{Temporal video generation results with and without register tokens on YTVIS dataset. Results without register tokens (bottom) lead to spatial inconsistencies across frames. Results with register tokens (middle) maintain consistent object positioning relative to ground truth (top).}
    \label{fig:reg_token_vid_29}
\end{figure*}

\begin{figure*}
    \centering
    \includegraphics[width=\linewidth]{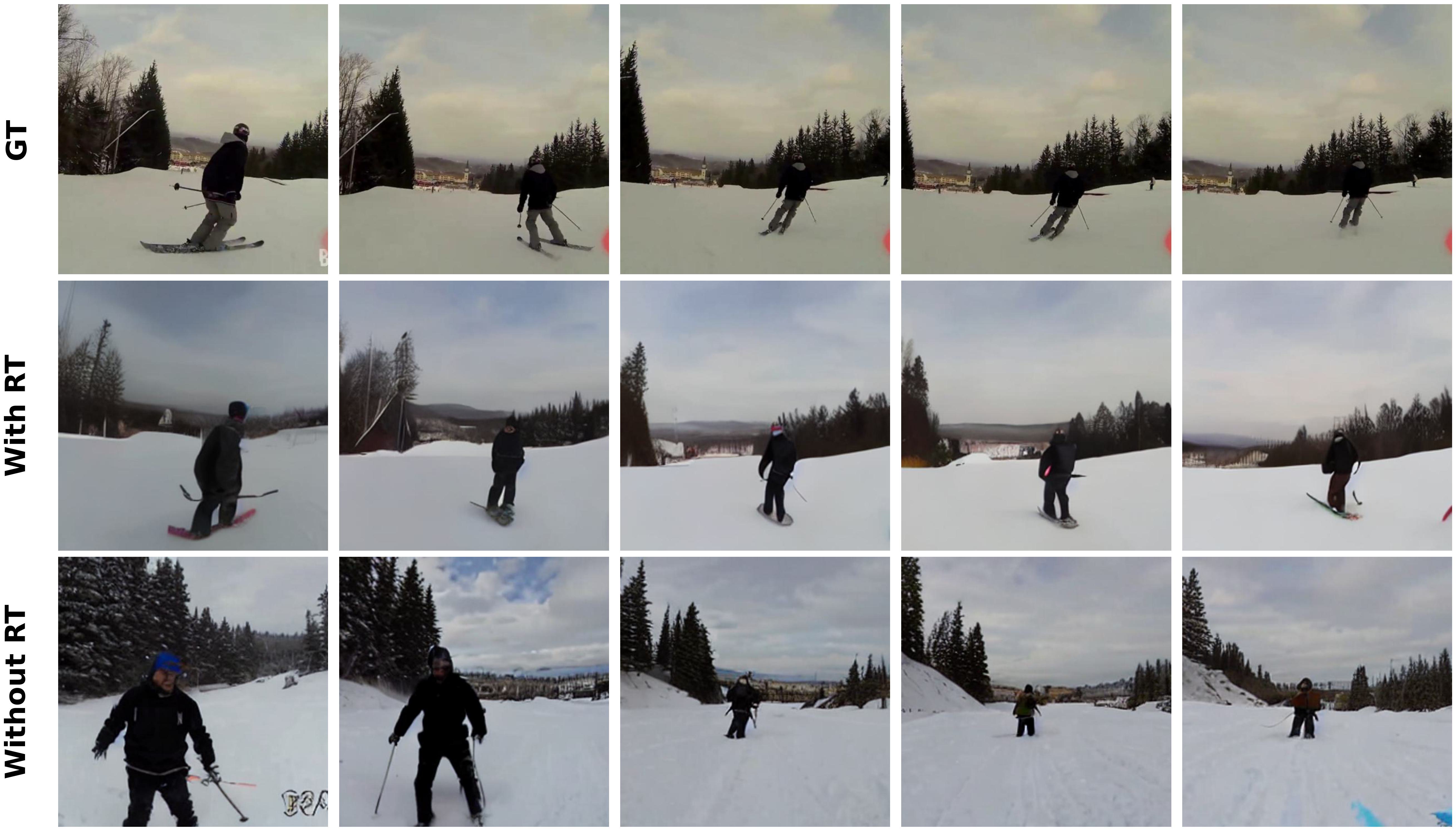}
    \caption{Temporal video generation results with and without register tokens on YTVIS dataset. Results without register tokens (bottom) lead to spatial inconsistencies across frames. Results with register tokens (middle) maintain consistent object positioning relative to ground truth (top).}
    \label{fig:reg_token_vid_69}
\end{figure*}

\begin{figure*}
    \centering
    \includegraphics[width=\linewidth]{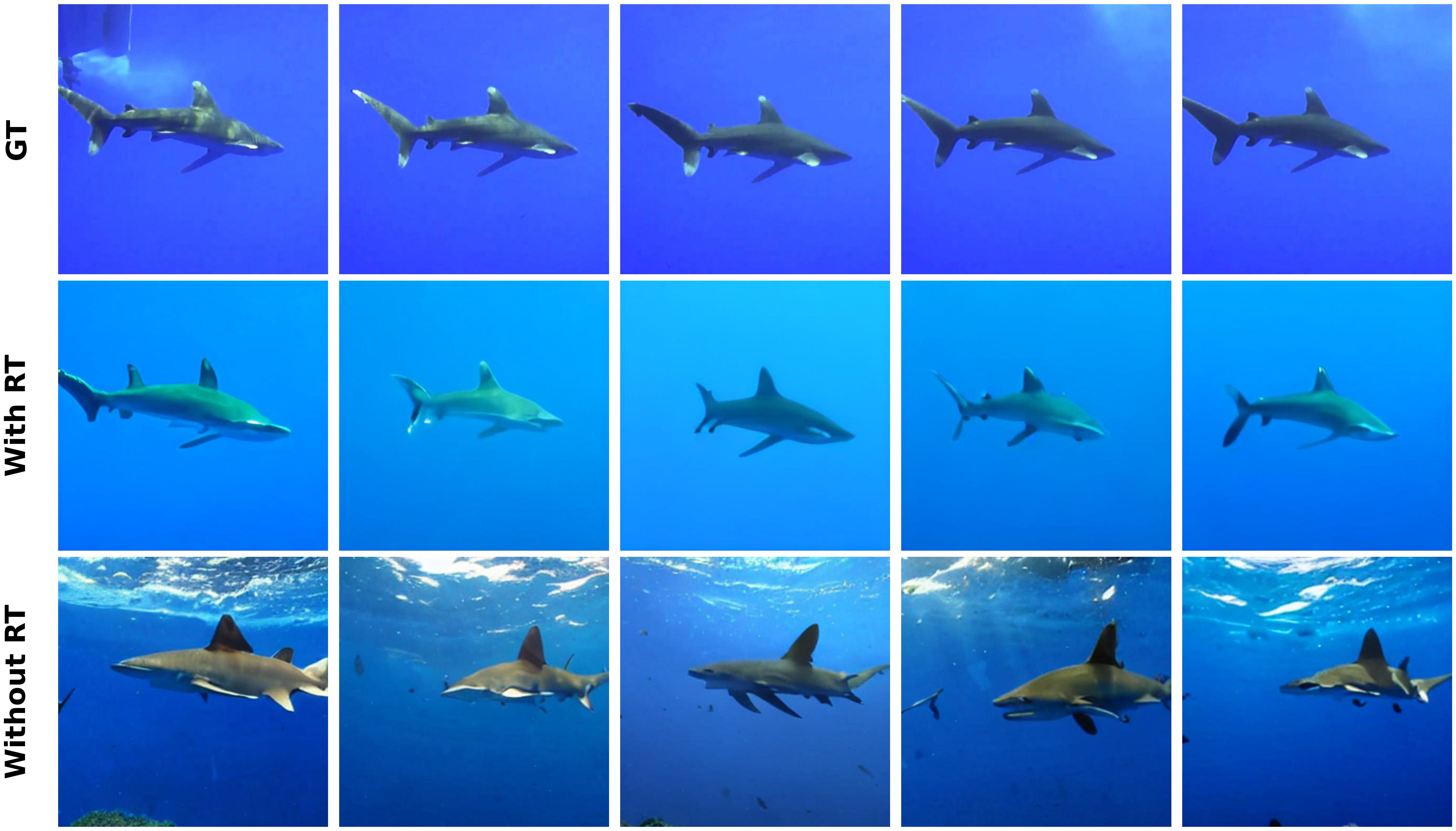}
    \caption{Temporal video generation results with and without register tokens on YTVIS dataset. Results without register tokens (bottom) lead to spatial inconsistencies across frames. Results with register tokens (middle) maintain consistent object positioning relative to ground truth (top).}
    \label{fig:reg_token_vid_101}
\end{figure*}

\begin{figure*}
    \centering
    \includegraphics[width=\linewidth]{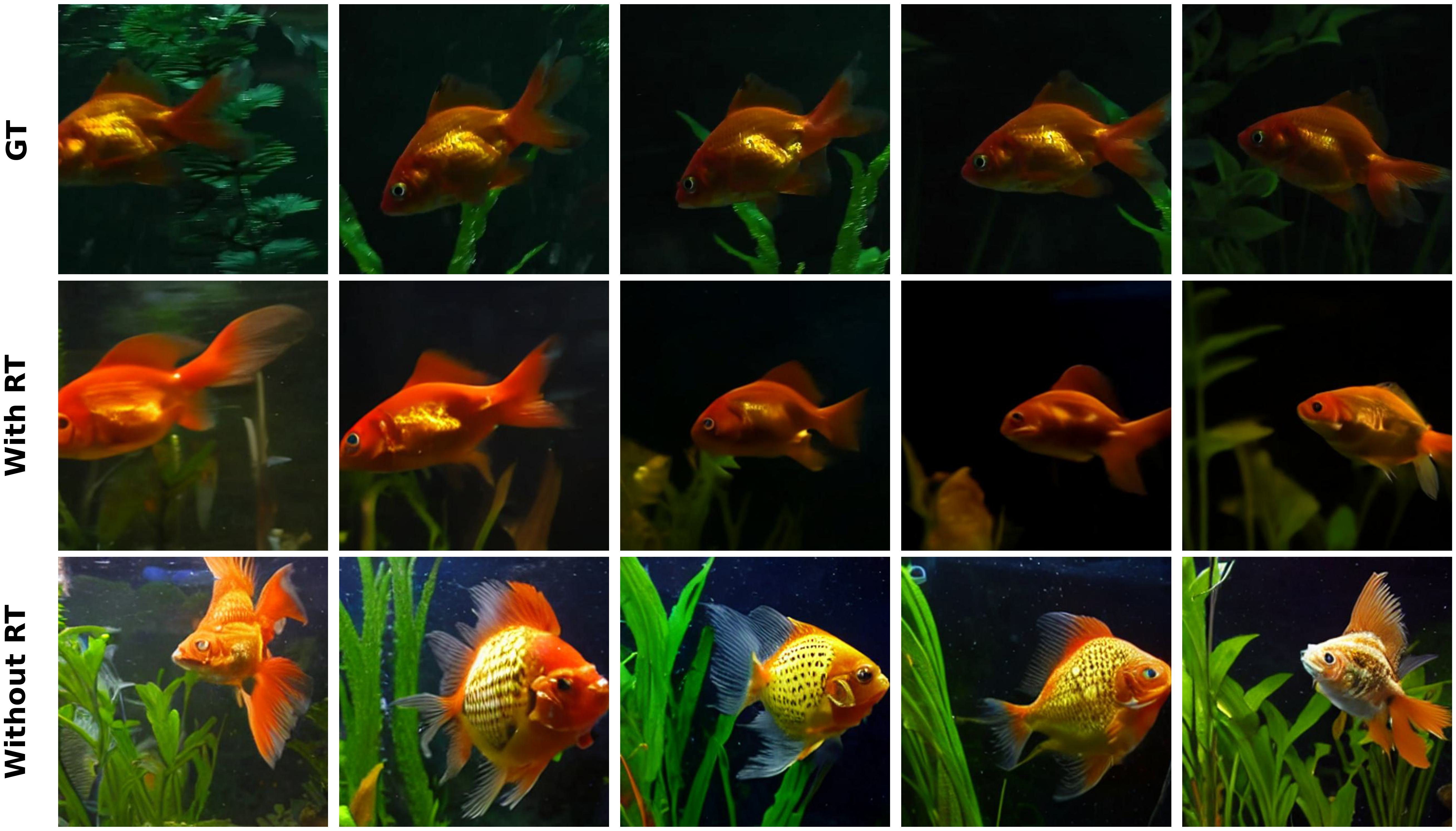}
    \caption{Temporal video generation results with and without register tokens on YTVIS dataset. Results without register tokens (bottom) lead to spatial inconsistencies across frames. Results with register tokens (middle) maintain consistent object positioning relative to ground truth (top).}
    \label{fig:reg_token_vid_126}
\end{figure*}

\begin{figure*}
    \centering
    \includegraphics[width=\linewidth]{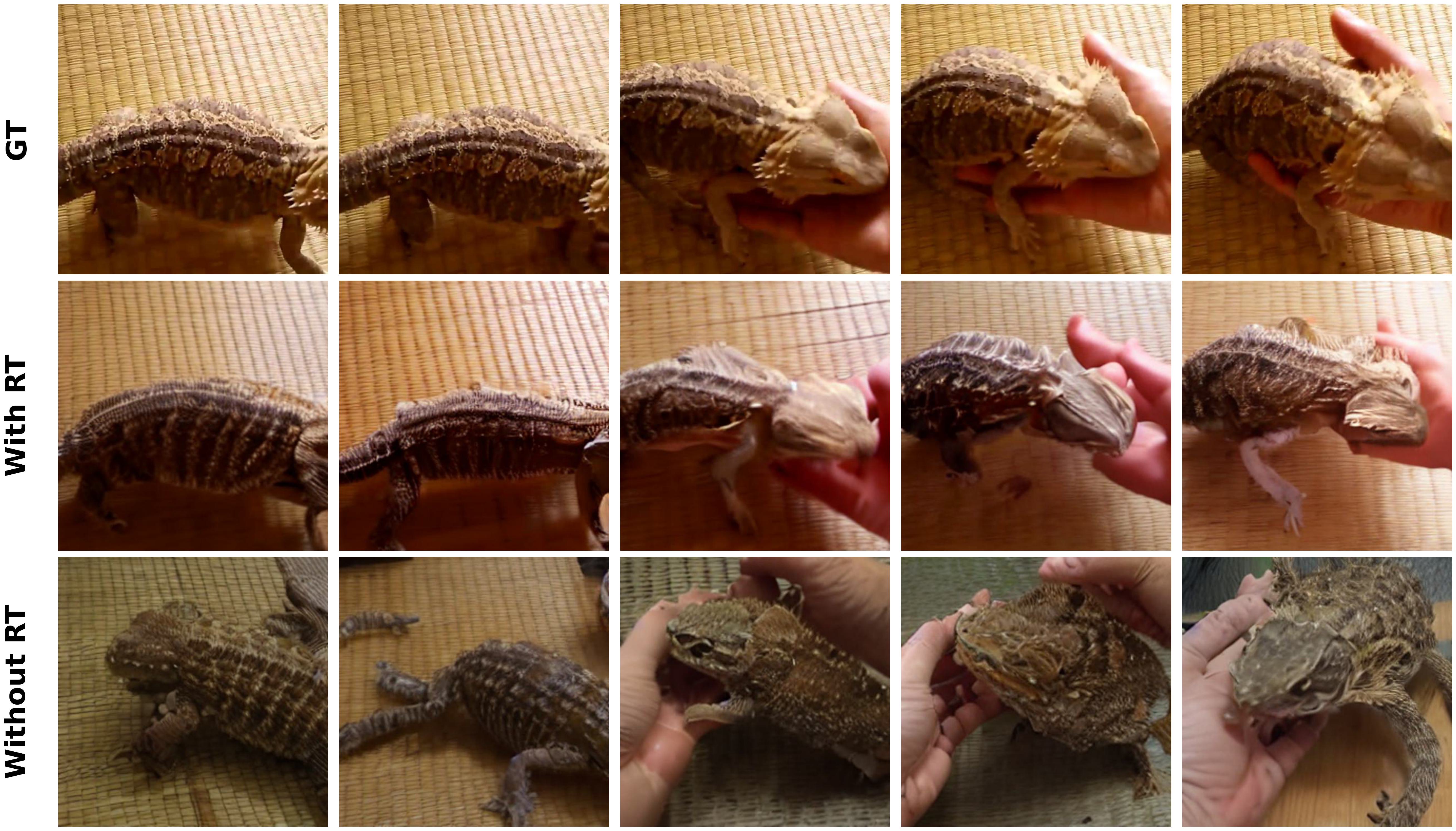}
    \caption{Temporal video generation results with and without register tokens on YTVIS dataset. Results without register tokens (bottom) lead to spatial inconsistencies across frames. Results with register tokens (middle) maintain consistent object positioning relative to ground truth (top).}
    \label{fig:reg_token_vid_153}
\end{figure*}

\begin{figure*}
    \centering
    \includegraphics[width=\linewidth]{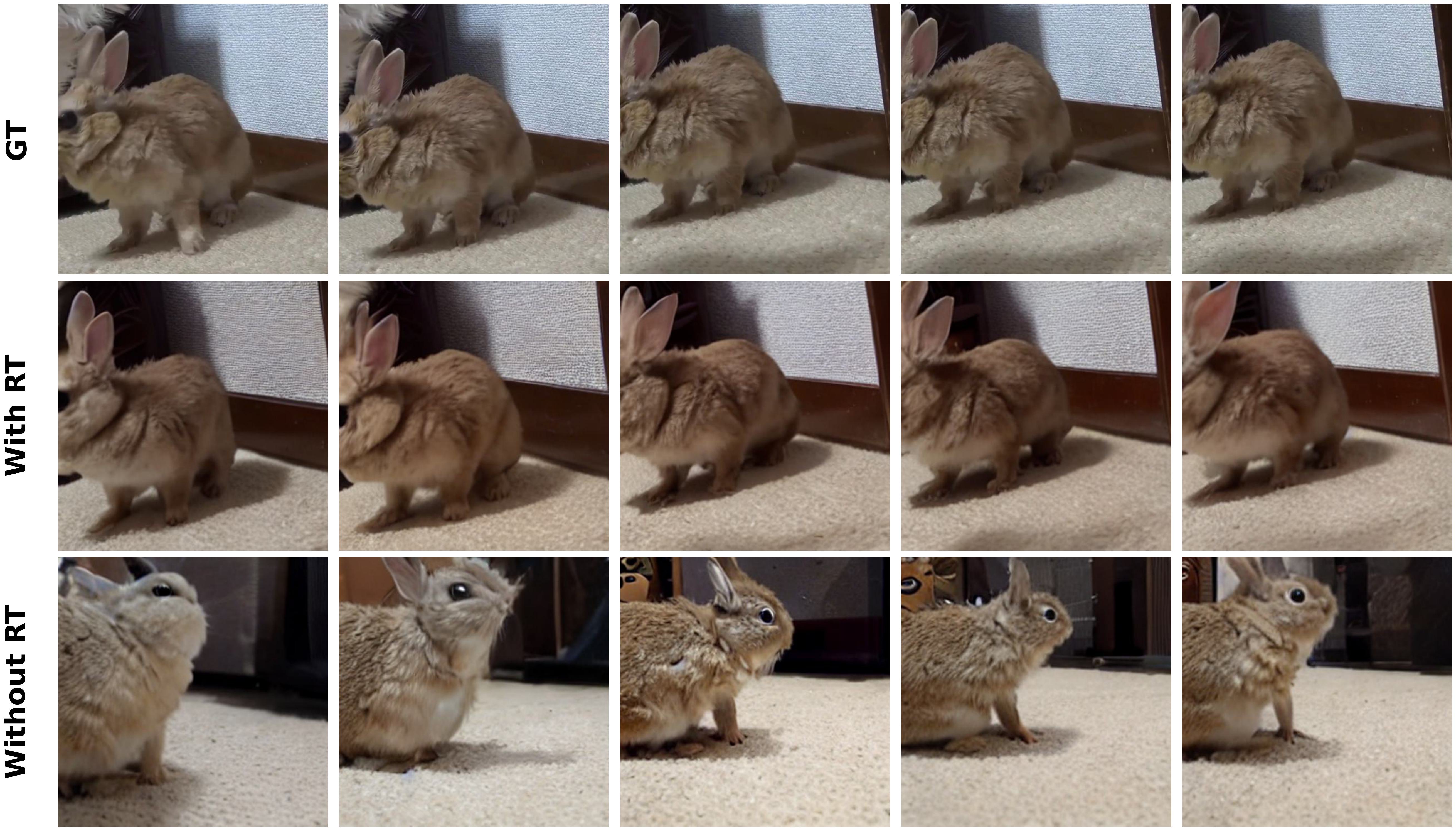}
    \caption{Temporal video generation results with and without register tokens on YTVIS dataset. Results without register tokens (bottom) lead to spatial inconsistencies across frames. Results with register tokens (middle) maintain consistent object positioning relative to ground truth (top).}
    \label{fig:reg_token_vid_168}
\end{figure*}

We validate our register token mechanism introduced in Section~\ref{subsec:isa_adaptation}. As explained in the main paper, register tokens provide spatial context to the diffusion decoder while keeping ISA slots free from spatial information, which maintains object-centric representations during generation.\\
\textbf{Evaluation Setup.} Table~\ref{table:video-gen-reg-token} compares our model with and without register tokens (RT) on YTVIS using the metrics from the main paper. We use the same trained model for both conditions, but replace register tokens with zero vectors during inference to isolate their effect on maintaining invariance properties.\\
\textbf{Quantitative Results.} Removing register tokens degrades performance across all metrics, confirming their role in preserving invariance. Without register tokens, reconstruction quality drops (PSNR, SSIM) because spatial information leaks into the slots, breaking their pose-invariant design. Perceptual quality also degrades (LPIPS, FID), and temporal consistency suffers significantly (FVD) as objects cannot maintain stable relationships across frames.\\
\textbf{Temporal Analysis.} The problem becomes more evident when viewing consecutive frames. Without register tokens, the same object shifts position, changes scale, and alters orientation between frames. This happens because slots now include spatial information, which violates the invariance principle of object-centric learning. With register tokens, spatial information remains separate, so slots focus on object identity.\\
\textbf{Visual Results.} Figures~\ref{fig:reg_token_vid_0} through~\ref{fig:reg_token_vid_168} show results across different scenarios. Each figure displays ground truth (top), our method with register tokens (middle), and without register tokens (bottom).

Without register tokens, objects appear in wrong positions, show incorrect scaling, have inconsistent poses, and create background artifacts. These issues occur consistently across various object types and scenes.

With register tokens, results maintain correct spatial placement, consistent scaling, proper poses, and clean backgrounds. The sequences flow smoothly with natural object motion and stable spatial relationships.

These results confirm that register tokens successfully preserve ISA's invariance properties within our diffusion framework, as detailed in Section~\ref{subsec:isa_adaptation}.

\begin{table}[t]
  \caption{\textbf{Video generation on YTVIS.} Register-token (RT) ablation.  
  The RT clearly boosts pixel accuracy (PSNR, SSIM), perceptual quality (LPIPS, FID) and temporal coherence (FVD) by carrying the pose information.}
  \label{table:video-gen-reg-token}
  \centering
  \small
  \setlength{\tabcolsep}{6pt}
  \begin{tabular}{lccccc}
    \toprule
    \textbf{Method} & PSNR$\uparrow$ & SSIM$\uparrow$ & LPIPS$\downarrow$ & FID$\downarrow$ & FVD$\downarrow$ \\
    \midrule
    w/o RT & 9.90 & 0.28 & 0.69 & 85.0 & 103.0 \\
    w/ RT  & \textbf{11.37} & \textbf{0.3933} & \textbf{0.5908} & \textbf{49.51} & \textbf{51.77} \\
    \bottomrule
  \end{tabular}
  \vspace{-0.5cm}
\end{table}

\clearpage

\section{Comparison with baselines}
\label{supp:sec:vis}
This section presents comprehensive video generation and segmentation results, comparing them against baseline methods across temporal sequences. While single-frame results are provided in the main paper, this supplementary material emphasizes temporal consistency and visual fidelity across consecutive frames to demonstrate the effectiveness of our unified slot-based framework. 

\newpage
\subsection{Generation Results}

\begin{figure*}[htbp]
    \centering
    \includegraphics[width=0.9\linewidth]{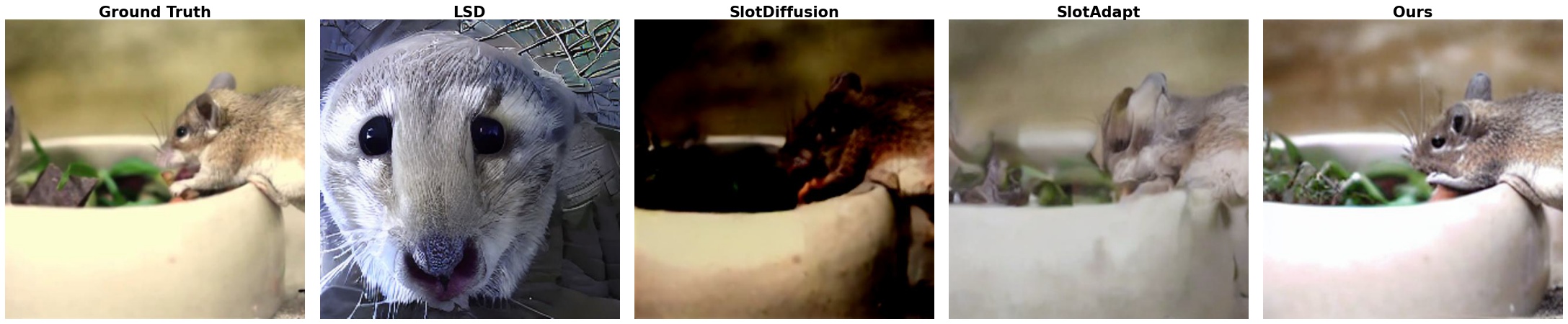}
    \includegraphics[width=0.9\linewidth]{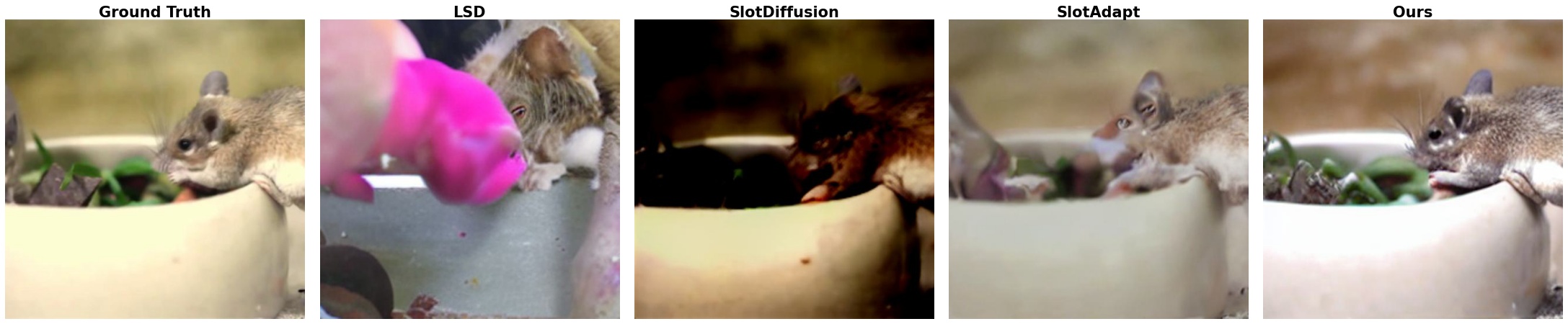}
    \includegraphics[width=0.9\linewidth]{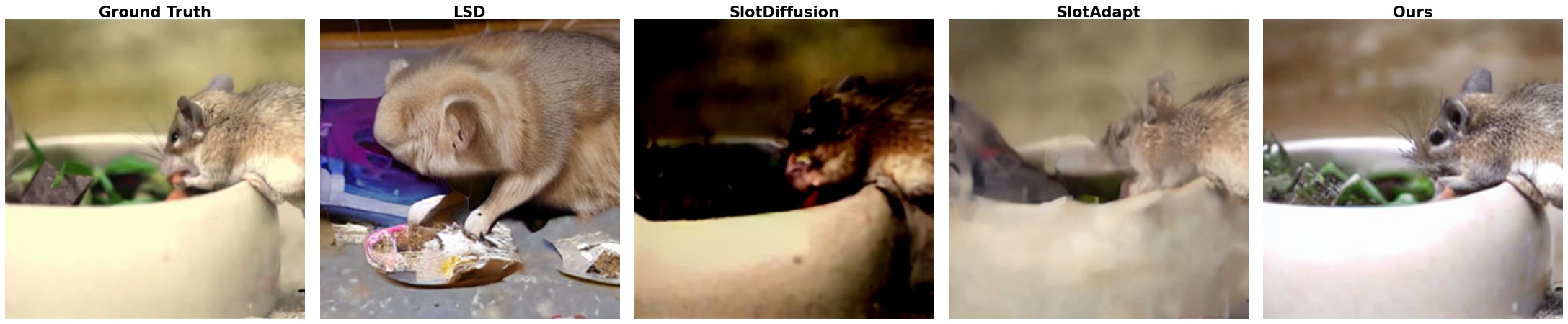}
    \includegraphics[width=0.9\linewidth]{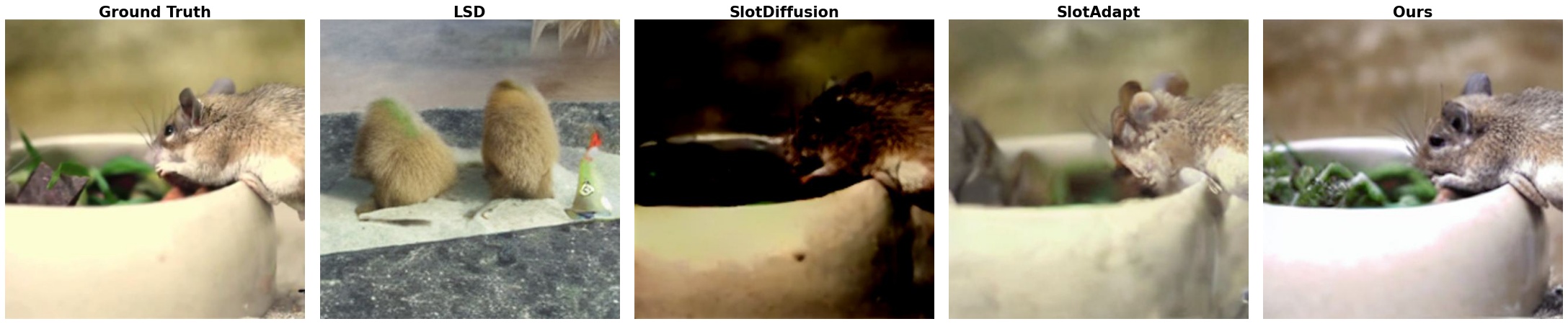}
    \includegraphics[width=0.9\linewidth]{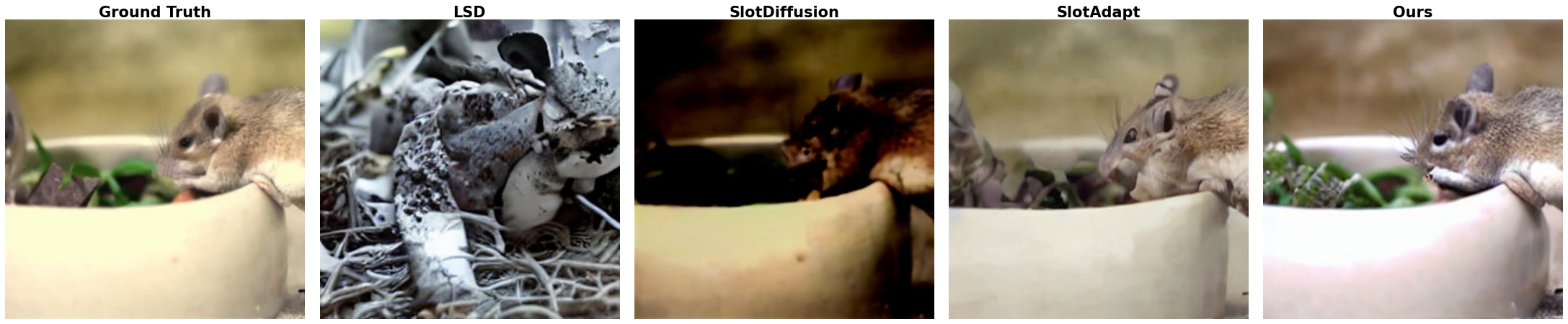}
    \caption{\textbf{Temporal Video Generation on YTVIS Dataset.} Multi-frame generation results showing object identity preservation and spatial coherence across five consecutive frames.}
    \label{fig:gen-comparison-ytvis-59}
\end{figure*}

\begin{figure*}[htbp]
    \centering
    \includegraphics[width=0.9\linewidth]{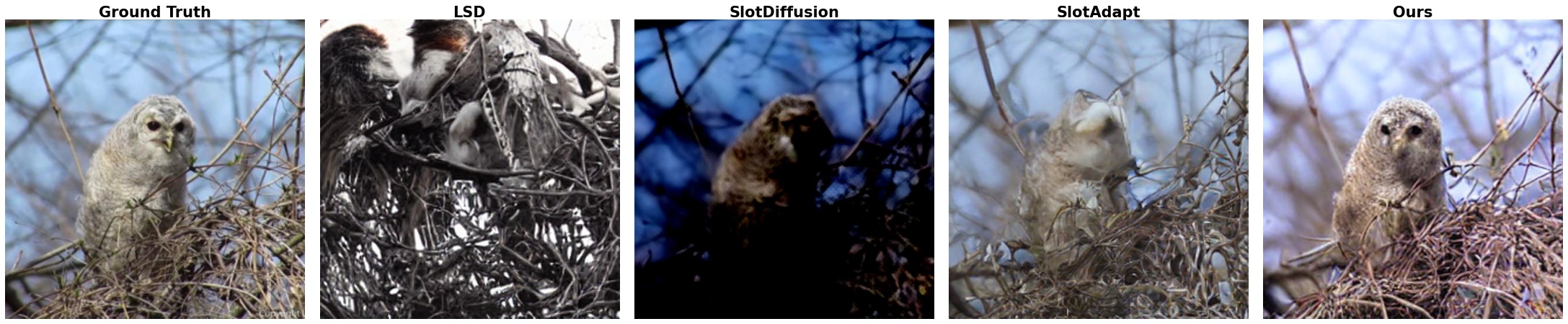}
    \includegraphics[width=0.9\linewidth]{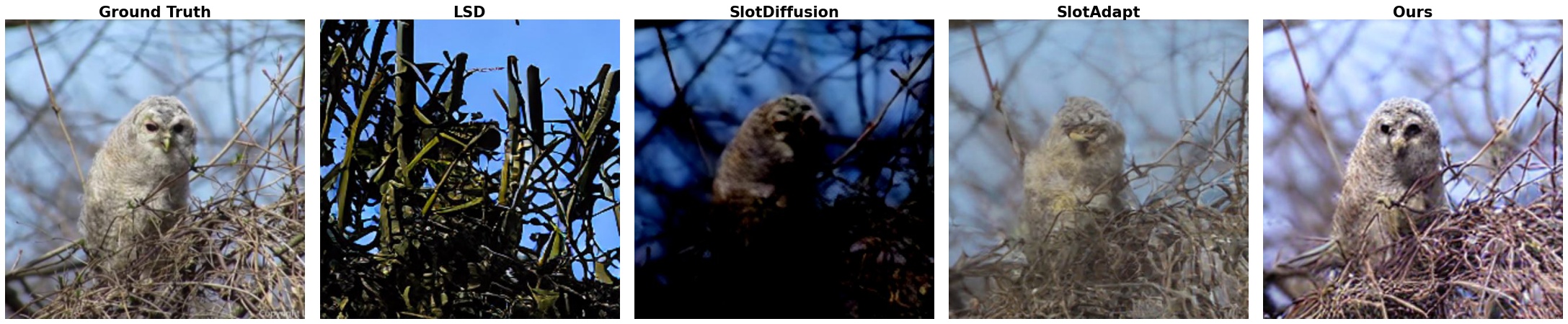}
    \includegraphics[width=0.9\linewidth]{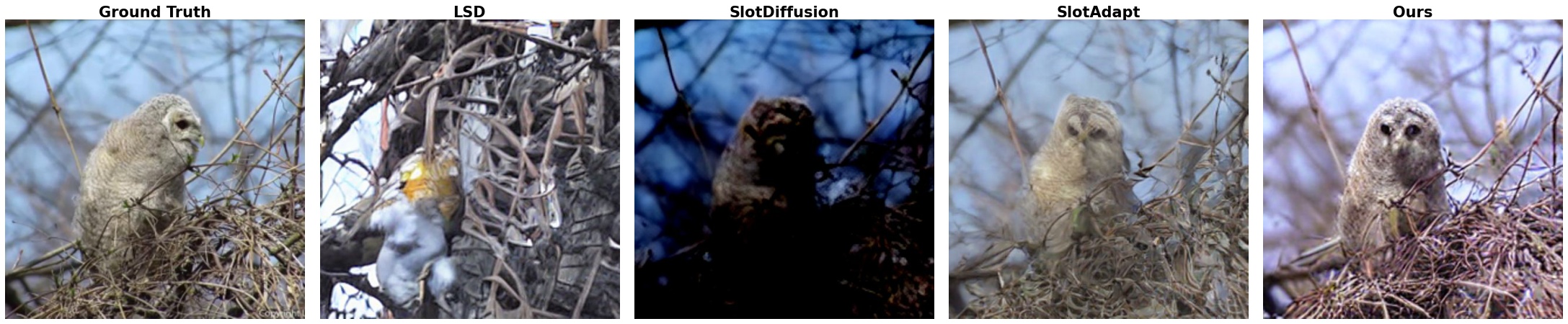}
    \includegraphics[width=0.9\linewidth]{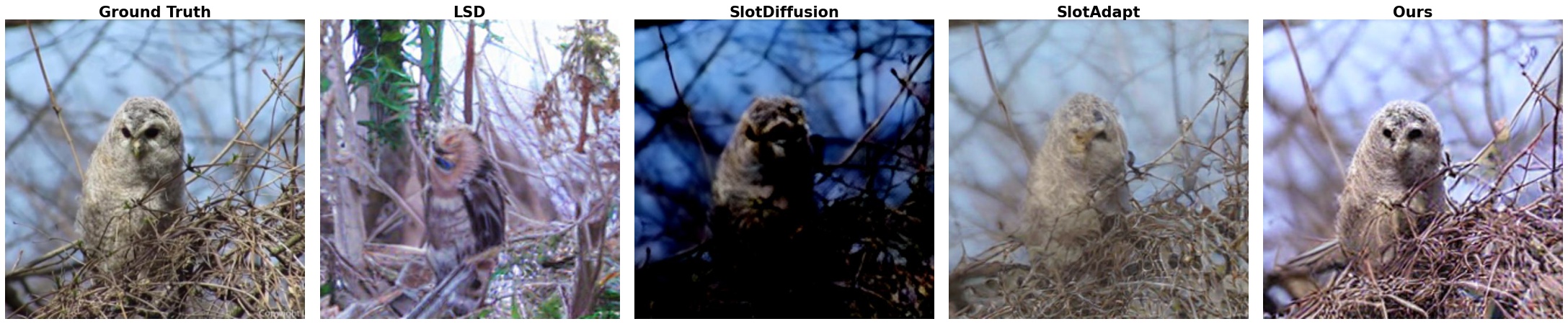}
    \includegraphics[width=0.9\linewidth]{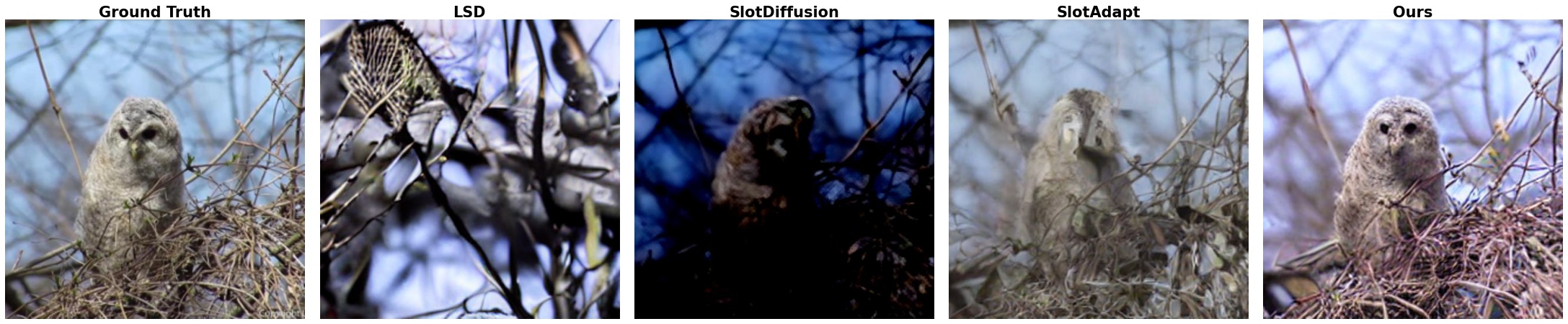}
    \caption{\textbf{Temporal Video Generation on YTVIS Dataset.} Video generation on challenging sequences with multiple objects and complex backgrounds.}
    \label{fig:gen-comparison-ytvis-78}
\end{figure*}

\begin{figure*}[htbp]
    \centering
    \includegraphics[width=0.9\linewidth]{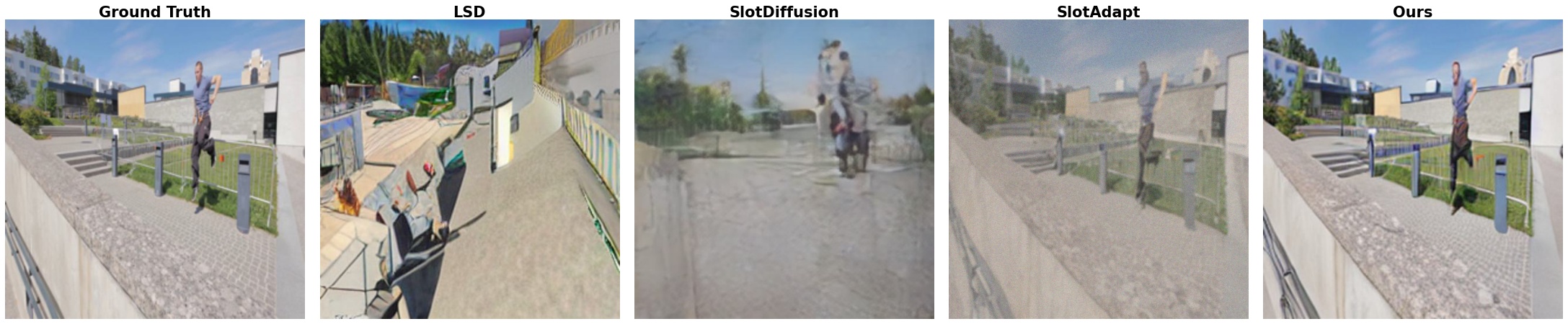}
    \includegraphics[width=0.9\linewidth]{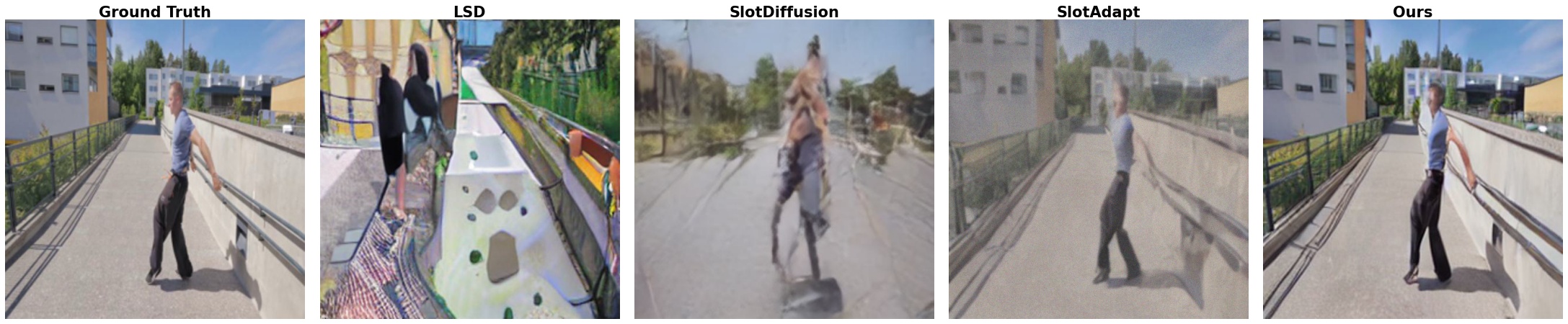}
    \includegraphics[width=0.9\linewidth]{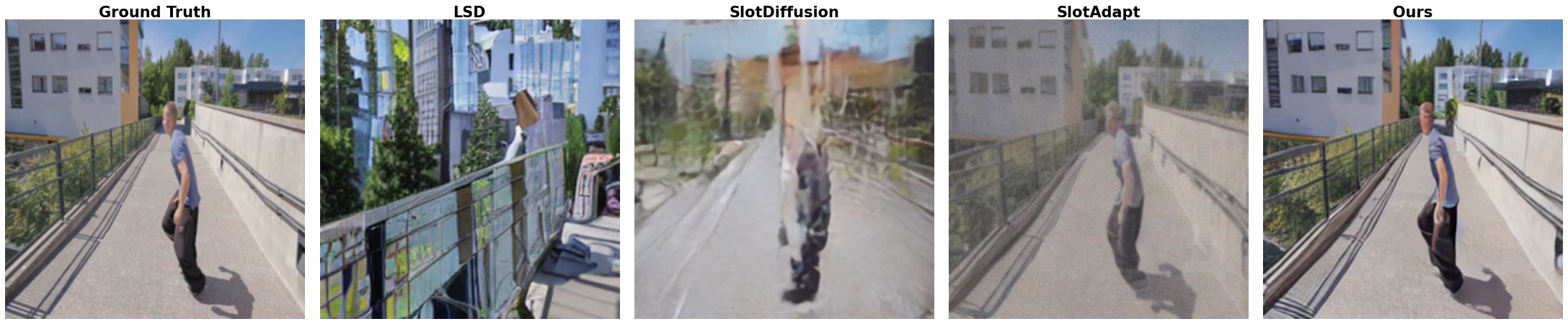}
    \includegraphics[width=0.9\linewidth]{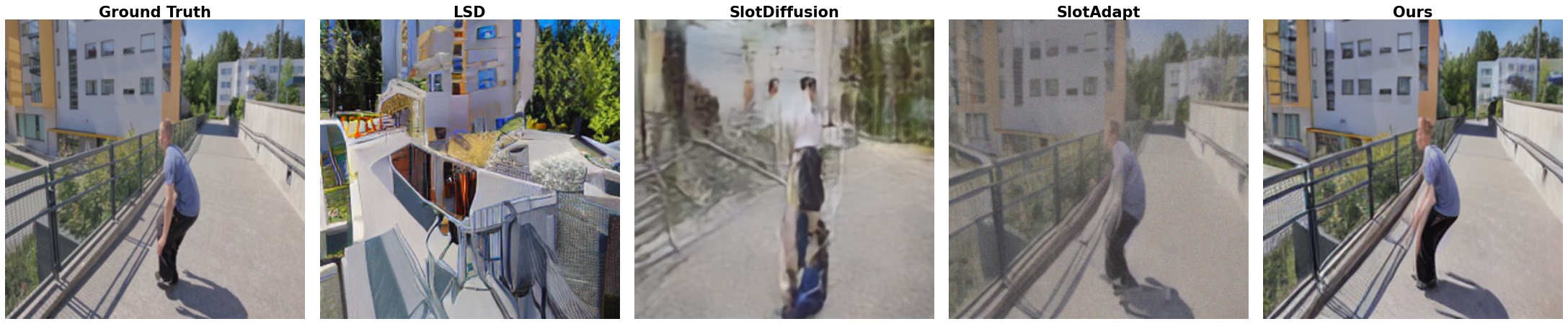}
    \includegraphics[width=0.9\linewidth]{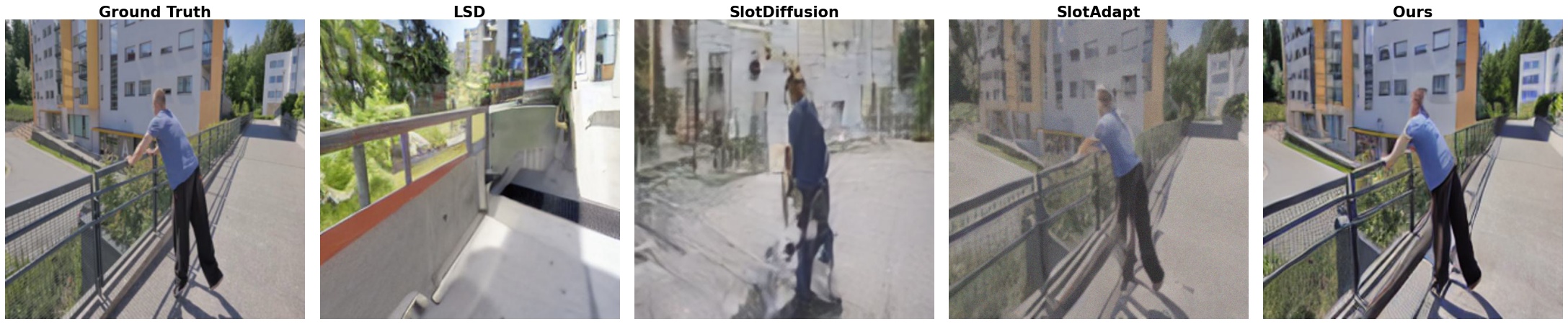}
    \caption{\textbf{Temporal Video Generation on DAVIS17 Dataset.} Results demonstrating object motion tracking and spatial layout consistency over time.}
    \label{fig:gen-comparison-davis-25}
\end{figure*}

\begin{figure*}[htbp]
    \centering
    \includegraphics[width=0.9\linewidth]{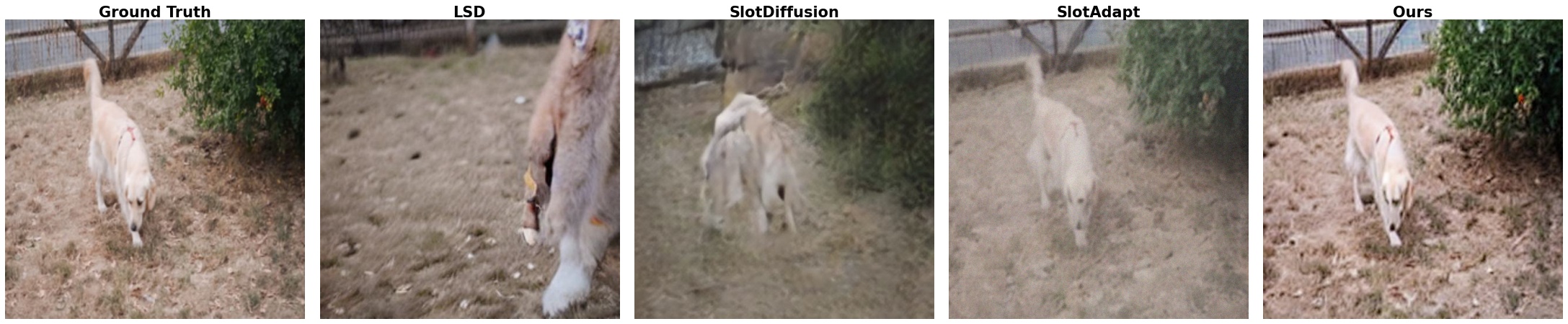}
    \includegraphics[width=0.9\linewidth]{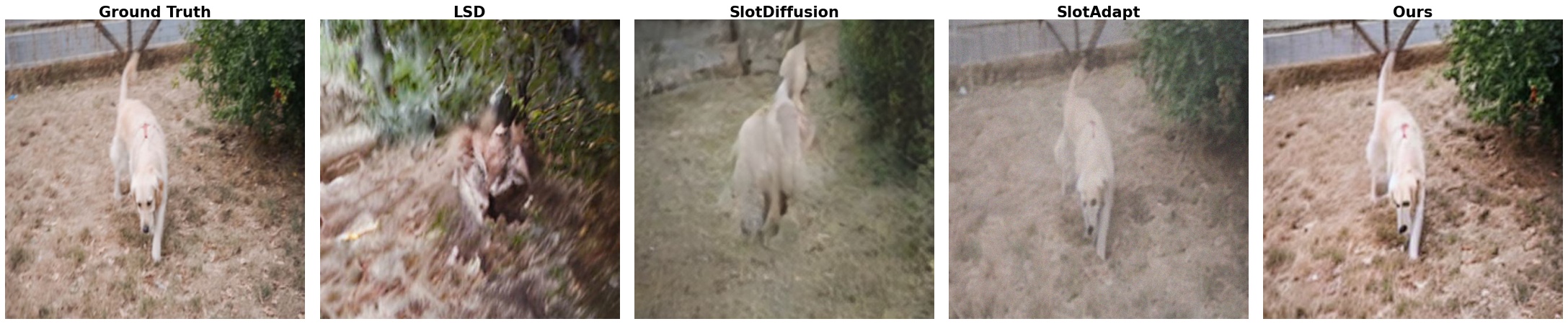}
    \includegraphics[width=0.9\linewidth]{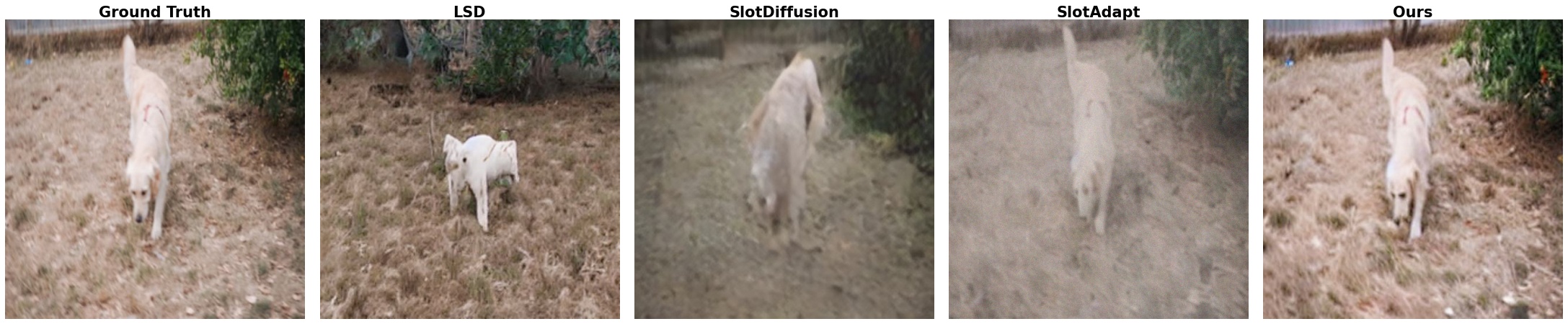}
    \includegraphics[width=0.9\linewidth]{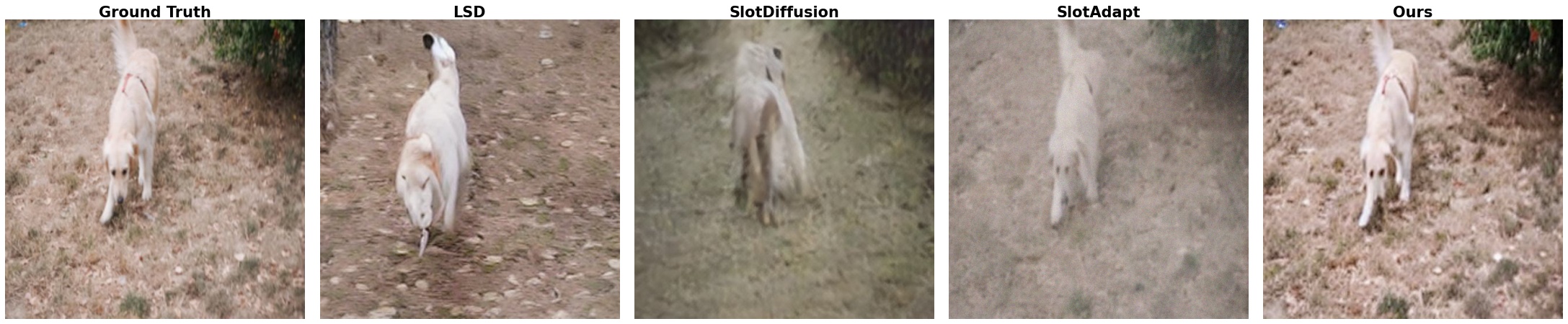}
    \includegraphics[width=0.9\linewidth]{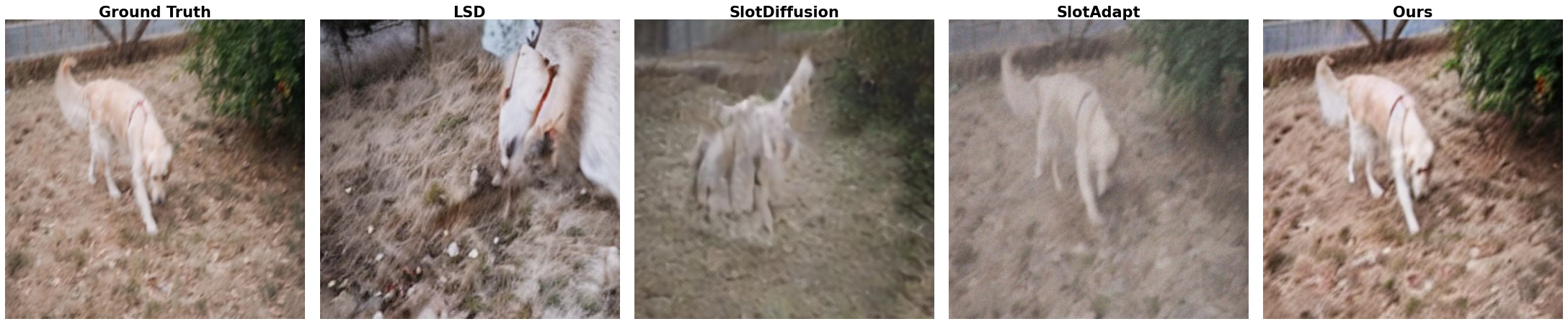}
    \caption{\textbf{Temporal Video Generation on DAVIS17 Dataset.} Generation results highlighting fine-detail preservation throughout temporal sequences.}
    \label{fig:gen-comparison-davis-9}
\end{figure*}

Figures~\ref{fig:gen-comparison-ytvis-59}, \ref{fig:gen-comparison-ytvis-78}, \ref{fig:gen-comparison-davis-25}, and \ref{fig:gen-comparison-davis-9} demonstrate our method's video generation capabilities across multiple consecutive frames on YTVIS and DAVIS17 datasets. Each figure shows five temporal frames from a single video sequence, with each row representing a different time step. The leftmost column shows ground truth frames, followed by results from baseline methods (LSD, SlotDiffusion, SlotAdapt), and our method in the rightmost column. \\
Our approach demonstrates improved temporal coherence, object identity preservation, and visual fidelity throughout the sequences. Key aspects to observe include: (1) structural stability of objects across time, (2) consistency of fine-grained details such as textures and colors, (3) natural motion dynamics, and (4) preservation of spatial relationships between objects and backgrounds. Baseline methods typically exhibit temporal artifacts, inconsistent object representations, and degradation in visual quality over time, while our unified framework successfully~handles these~challenging~scenarios through effective slot-basedtemporal binding. The results show our method's capability to generate high-quality, temporally consistent video content that maintains object coherence across complex motion patterns. 

\clearpage
\raggedbottom
\subsection{Segmentation Results}

\begin{figure*}[htbp]
    \centering
    \includegraphics[width=0.9\linewidth]{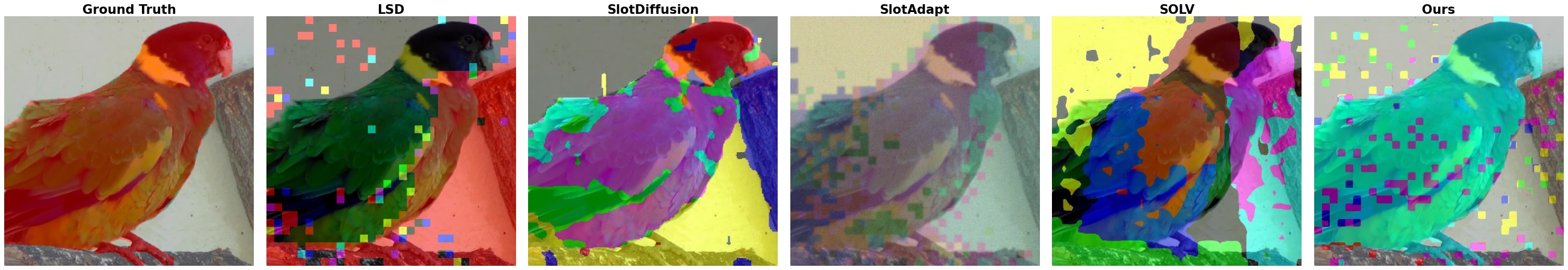}
    \includegraphics[width=0.9\linewidth]{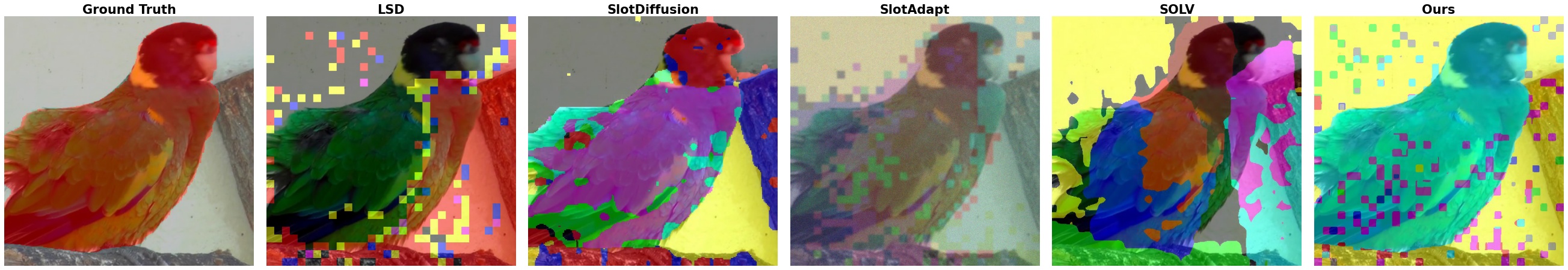}
    \includegraphics[width=0.9\linewidth]{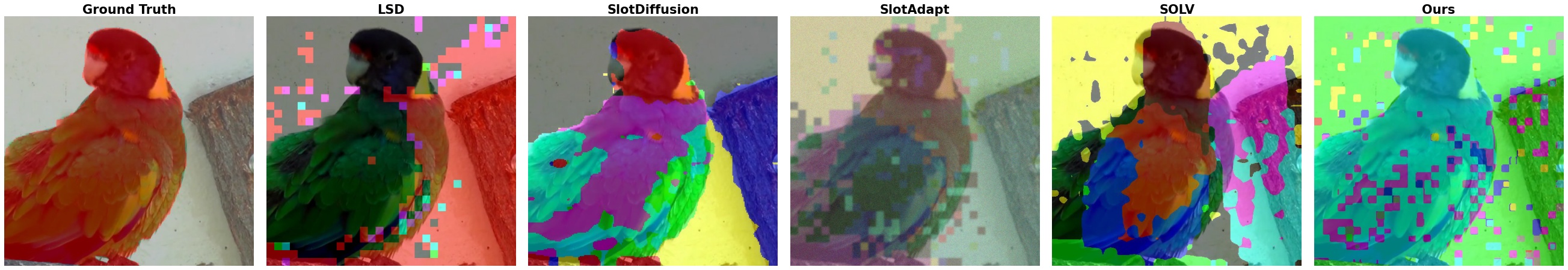}
    \includegraphics[width=0.9\linewidth]{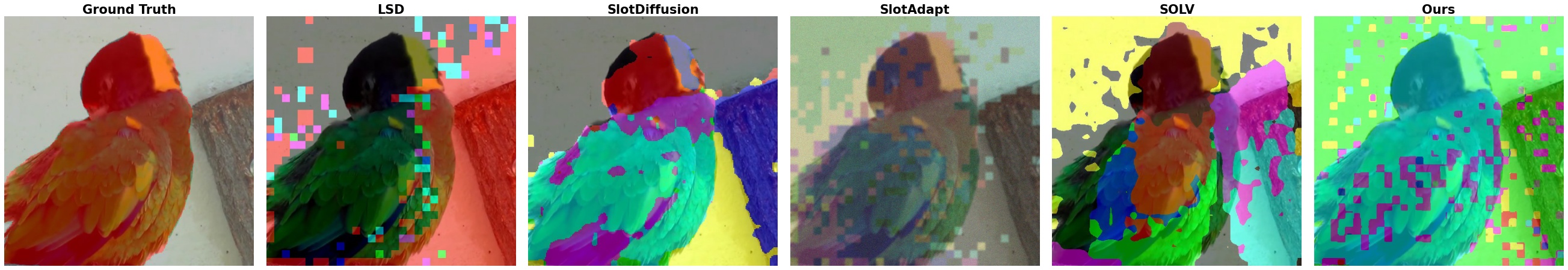}
    \includegraphics[width=0.9\linewidth]{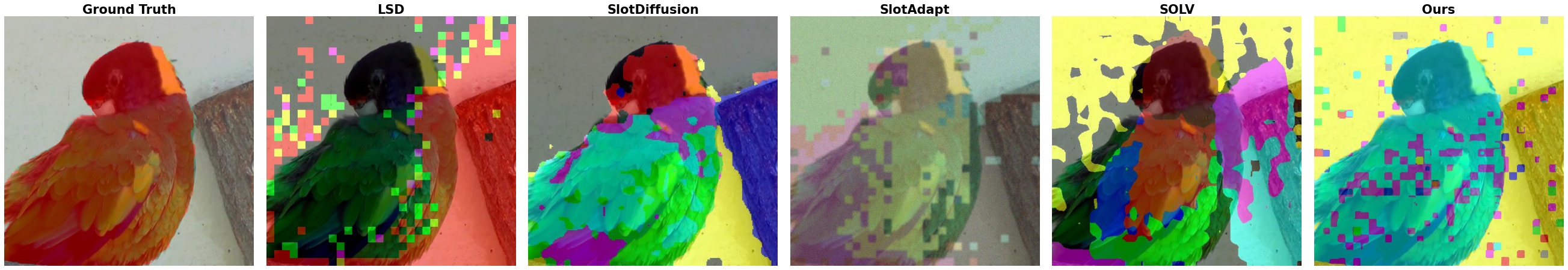}\vspace{-0.3cm}
    \caption{\textbf{Segmentation Results on YTVIS.} Consistent object boundary detection and identity preservation across dynamic motion and pose changes.}
    \label{fig:seg-comparison-ytvis-13}
\end{figure*}

\begin{figure*}[htbp]
    \centering
    \includegraphics[width=0.9\linewidth]{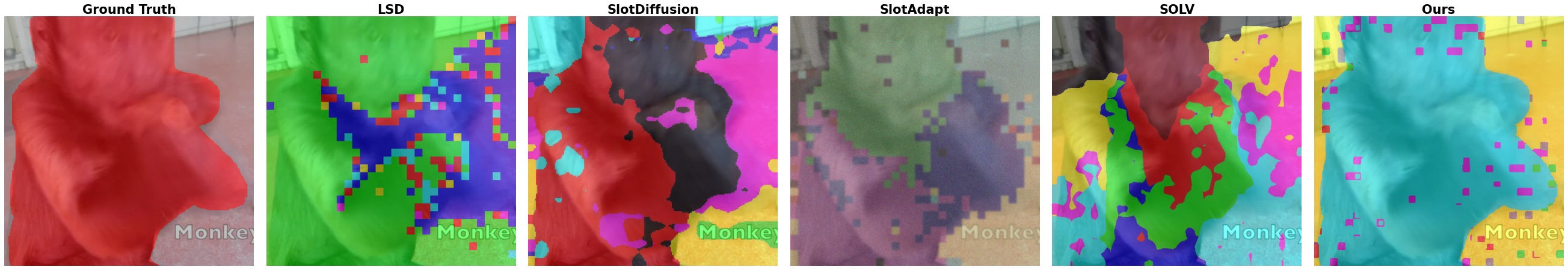}
    \includegraphics[width=0.9\linewidth]{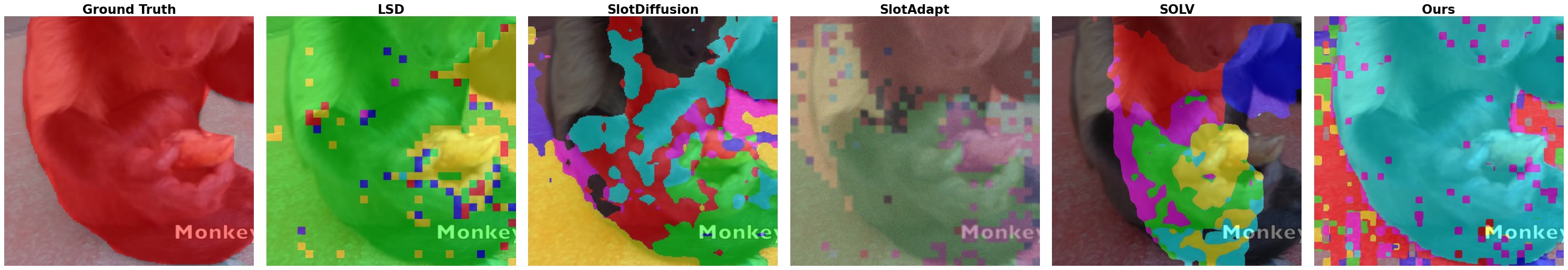}
    \includegraphics[width=0.9\linewidth]{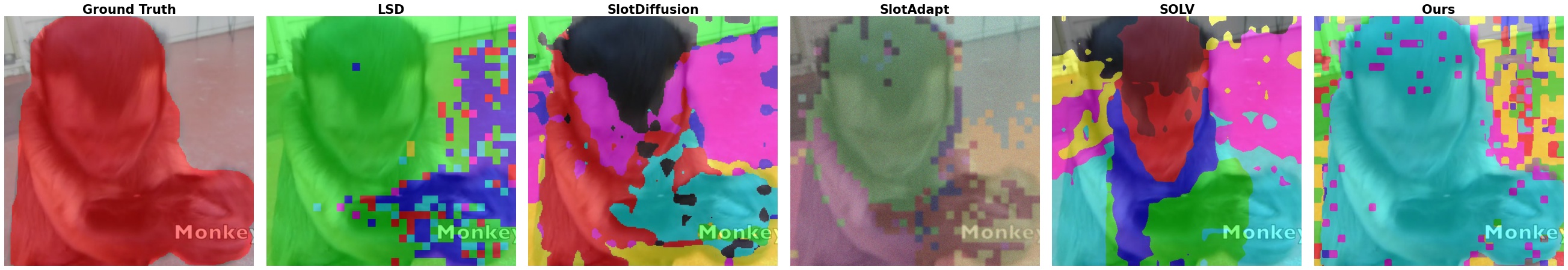}
    \includegraphics[width=0.9\linewidth]{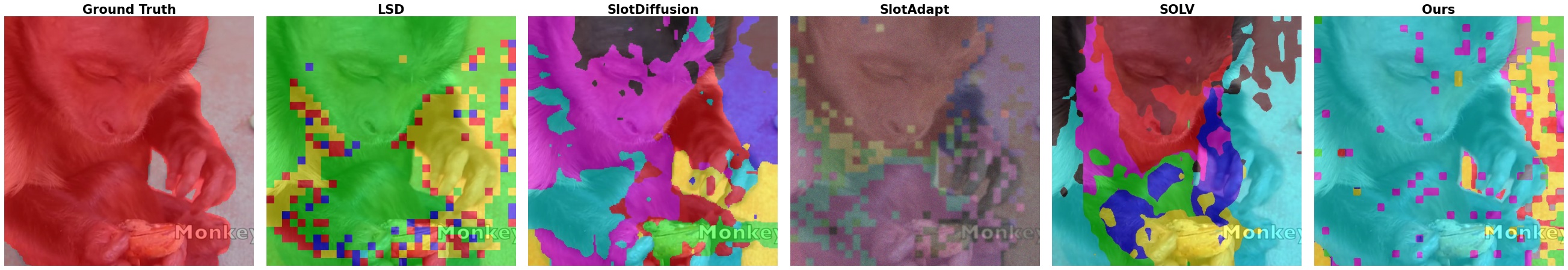}
    \includegraphics[width=0.9\linewidth]{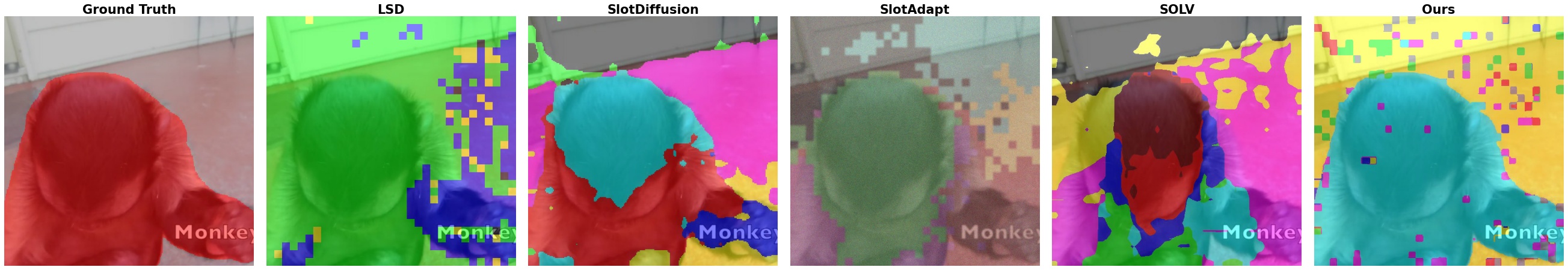}
    \caption{\textbf{Segmentation Results on YTVIS.} Handling of significant shape and appearance variations with temporal tracking.}
    \label{fig:seg-comparison-ytvis-26}
\end{figure*}

\begin{figure*}[htbp]
    \centering
    \includegraphics[width=0.9\linewidth]{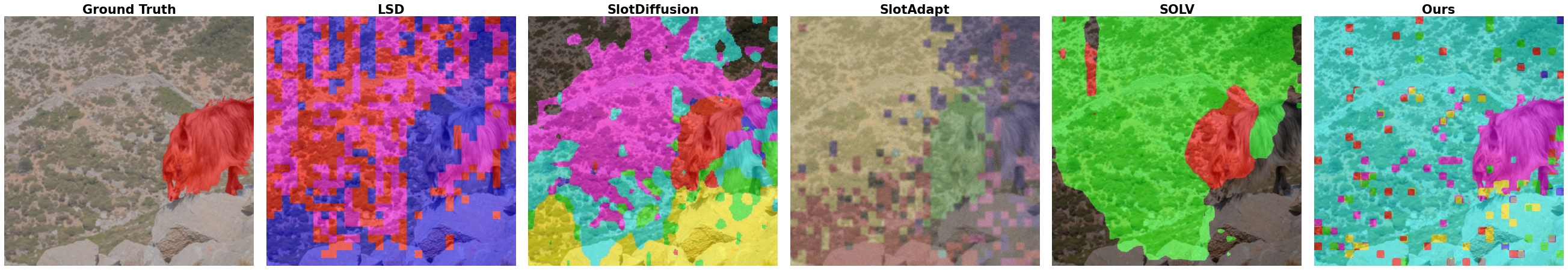}
    \includegraphics[width=0.9\linewidth]{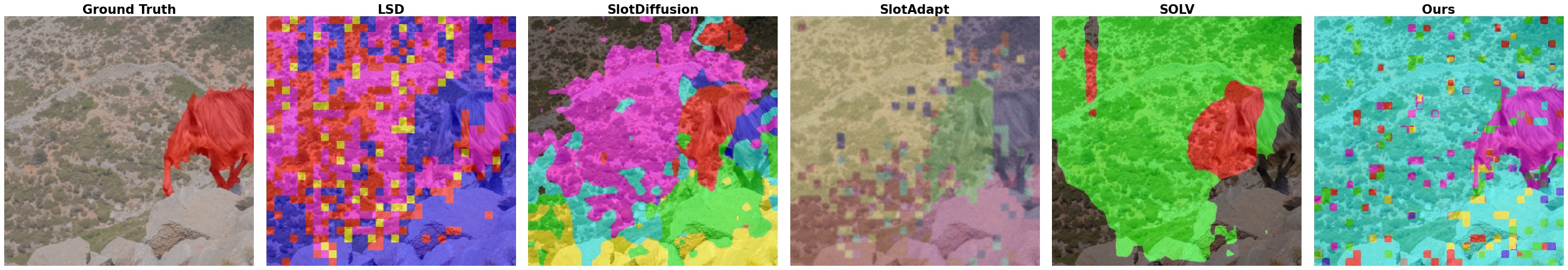}
    \includegraphics[width=0.9\linewidth]{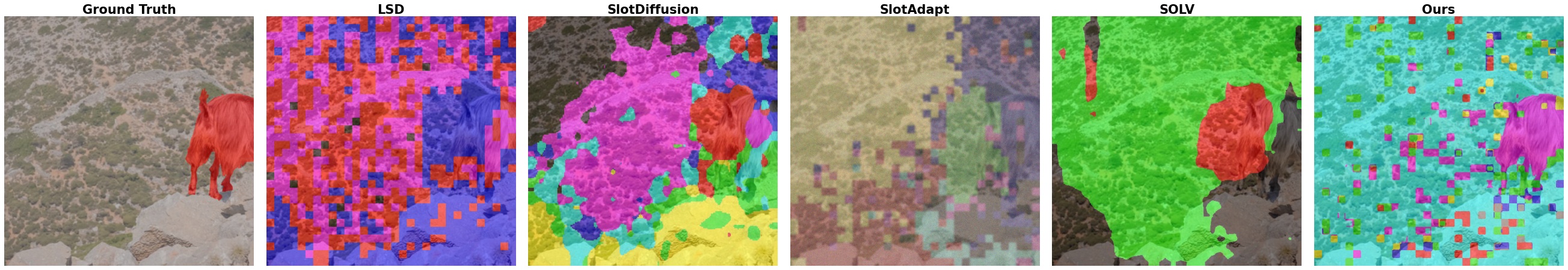}
    \includegraphics[width=0.9\linewidth]{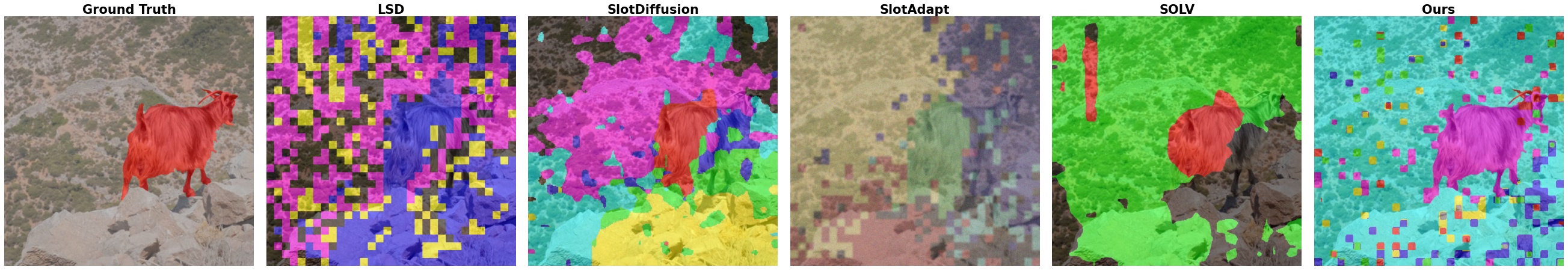}
    \includegraphics[width=0.9\linewidth]{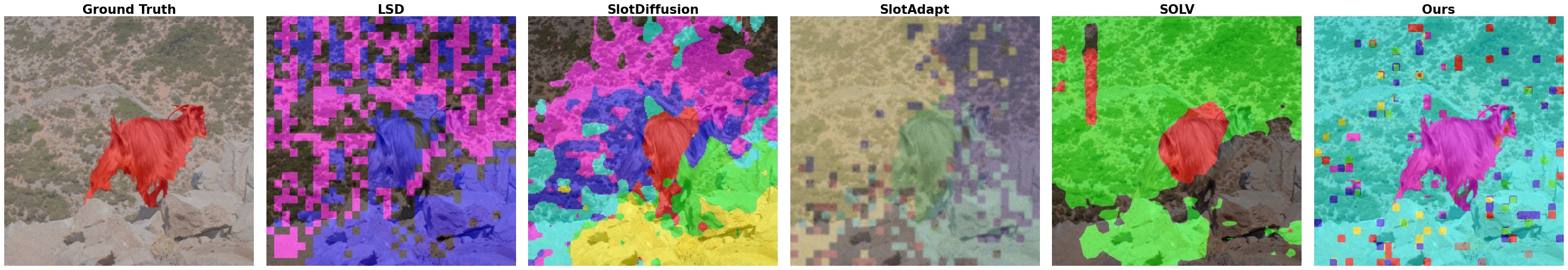}
    \caption{\textbf{Segmentation Results on DAVIS17.} Tracking and segmentation of small, fast-moving objects through rapid motion and scale changes.}
    \label{fig:seg-comparison-davis-13}
\end{figure*}

\begin{figure*}[htbp]
    \centering
    \includegraphics[width=0.9\linewidth]{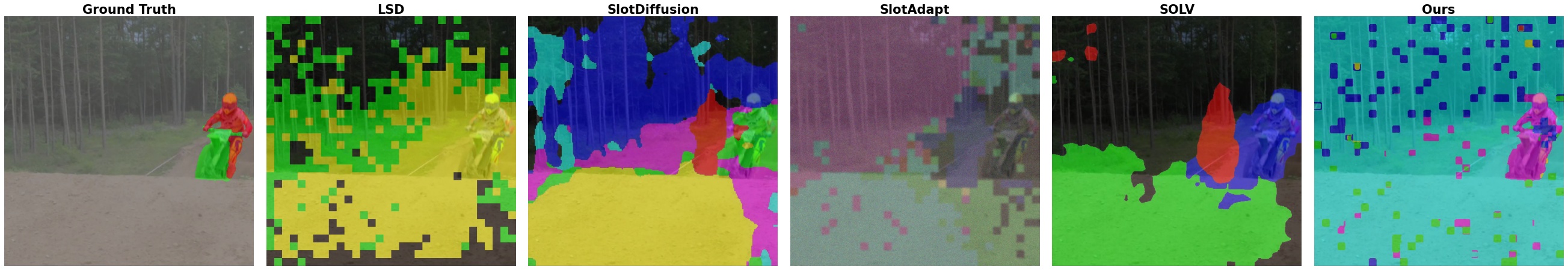}
    \includegraphics[width=0.9\linewidth]{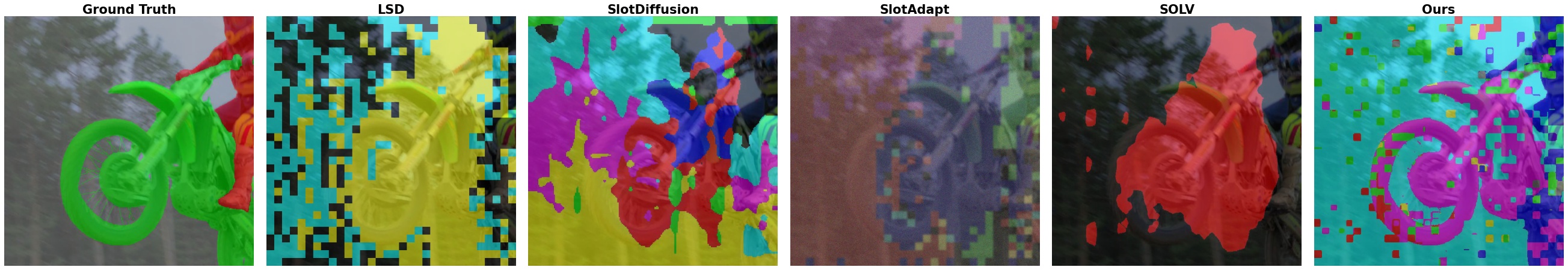}
    \includegraphics[width=0.9\linewidth]{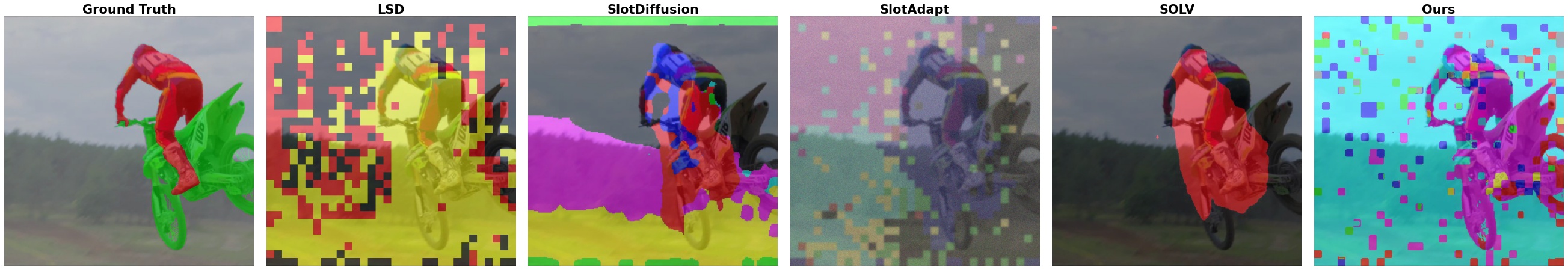}
    \includegraphics[width=0.9\linewidth]{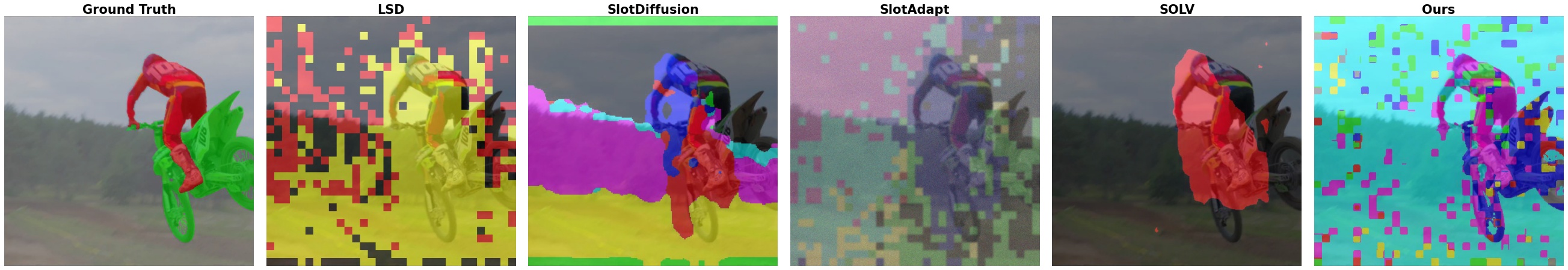}
    \includegraphics[width=0.9\linewidth]{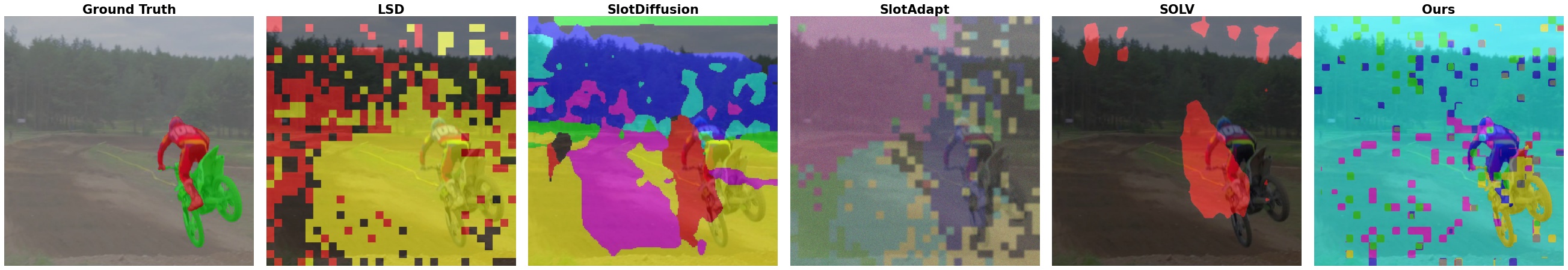}
    \caption{\textbf{Segmentation Results on DAVIS17.} Distinguishing and tracking multiple closely positioned objects with stable segmentation masks.}
    \label{fig:seg-comparison-davis-23}
\end{figure*}

Figures~\ref{fig:seg-comparison-ytvis-13}, \ref{fig:seg-comparison-ytvis-26}, \ref{fig:seg-comparison-davis-13}, and \ref{fig:seg-comparison-davis-23} demonstrate our method's unsupervised video object segmentation performance across temporal sequences on YTVIS and DAVIS17 datasets. Each figure displays five consecutive frames with segmentation masks overlaid, where different colors represent distinct object instances discovered by each method. The evaluation focuses on temporal binding consistency—the ability to maintain stable object identity and accurate boundaries across frames. 
\newpage
Our slot-based representations successfully handle various challenging scenarios including: (1) rapid object motion and deformation, (2) objects with similar appearances or spatial proximity, (3) scale changes and partial occlusions, and (4) complex multi-object interactions. The results demonstrate our framework's improved capability in preserving object identity and spatial coherence compared to baseline approaches, which often exhibit segmentation instability, identity confusion, and boundary degradation across temporal sequences. The consistent performance across diverse video content shows the effectiveness of our temporal object-centric learning approach for video understanding applications.

\clearpage

\subsection{Compositional Generations}
\begin{figure*}[htbp]
    \centering
    \includegraphics[width=0.9\linewidth]{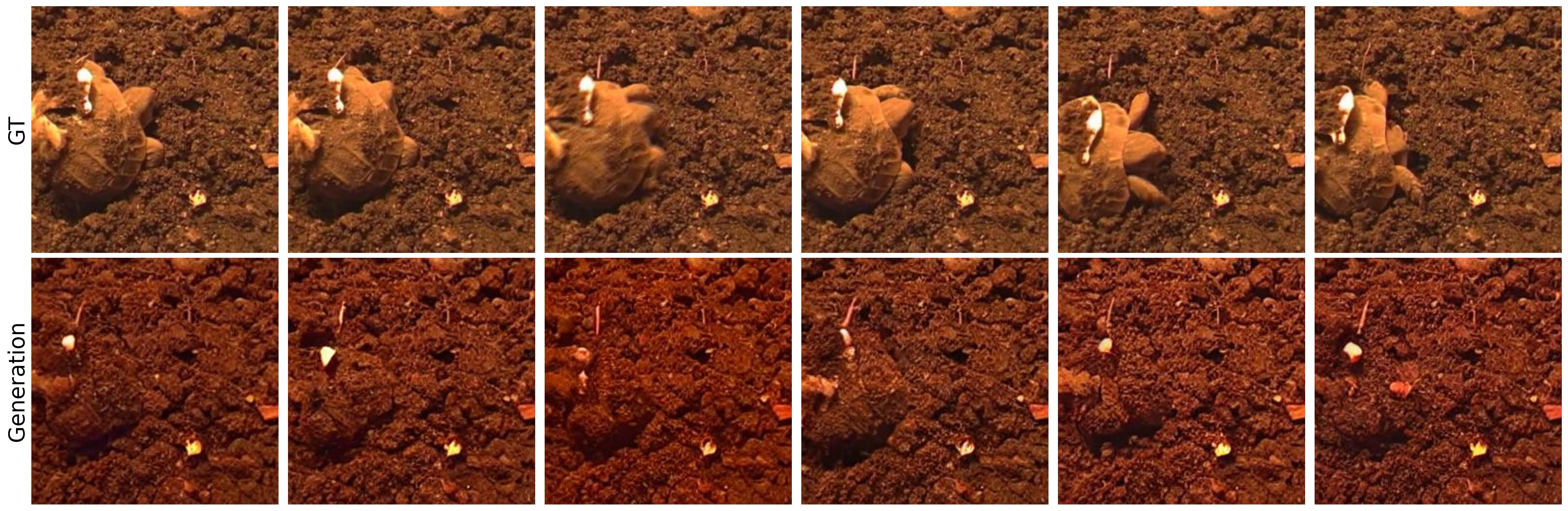}
    \includegraphics[width=0.9\linewidth]{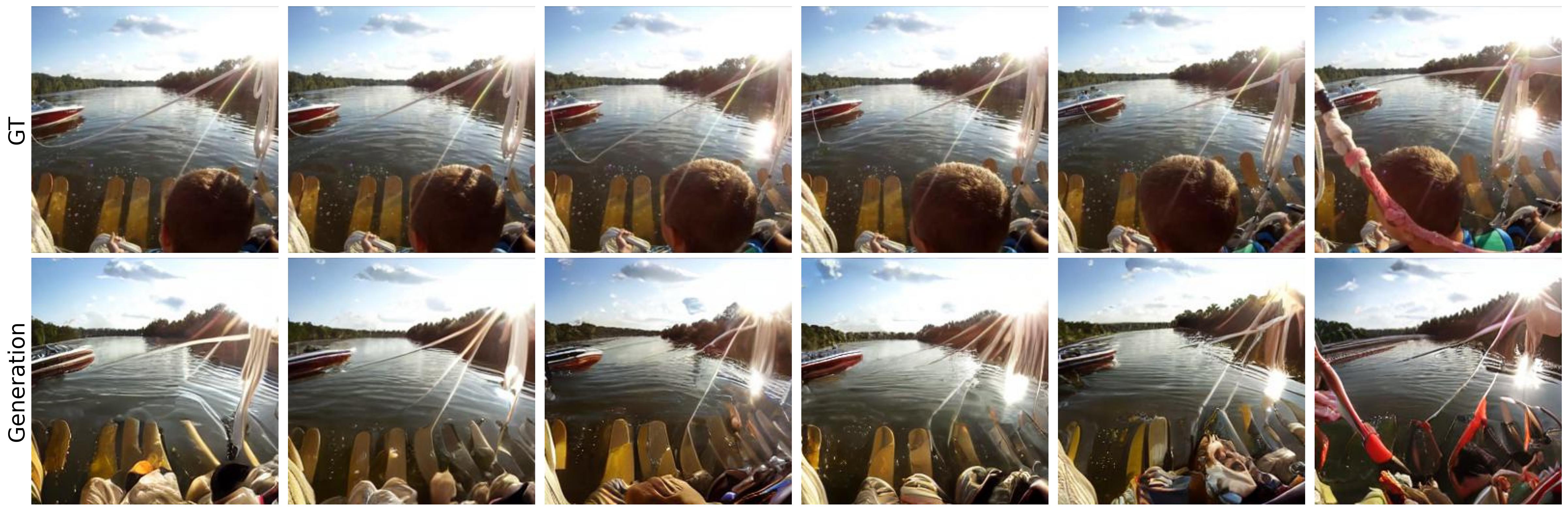}
    \includegraphics[width=0.9\linewidth]{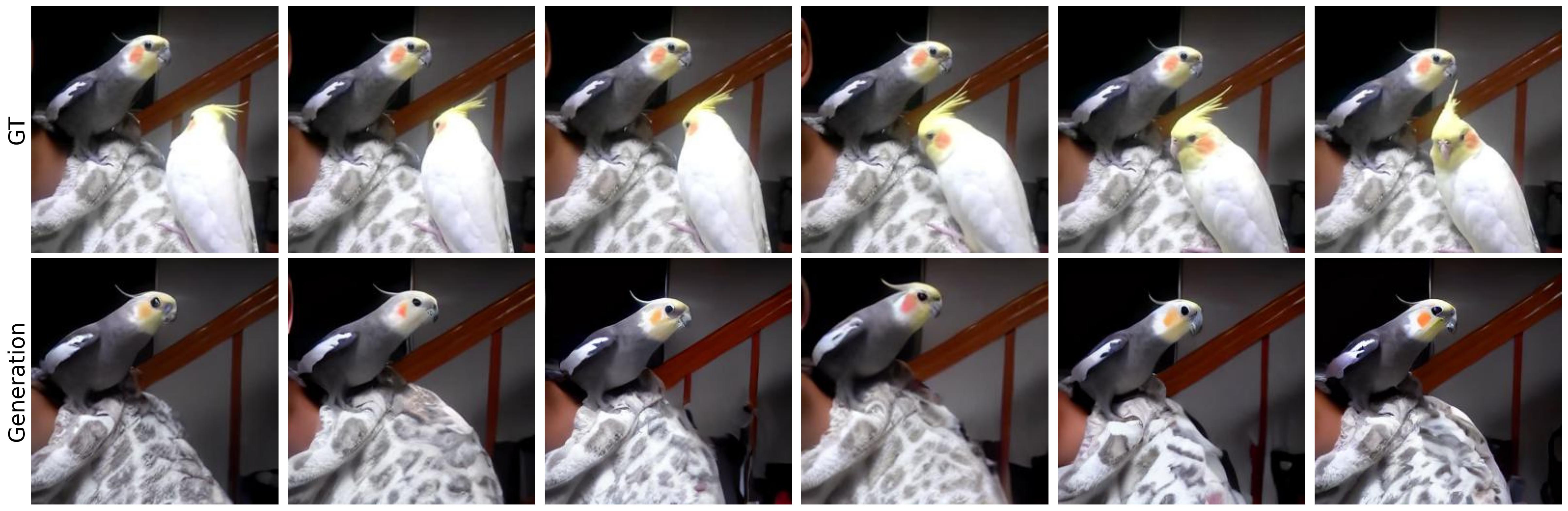}
    \caption{\textbf{Compositional Editing Examples.} Targeted object removal while maintaining scene coherence and temporal consistency.}
    \label{fig:comp-gen-multi-frame-1}
\end{figure*}

\begin{figure*}[htbp]
    \centering
    \includegraphics[width=0.9\linewidth]{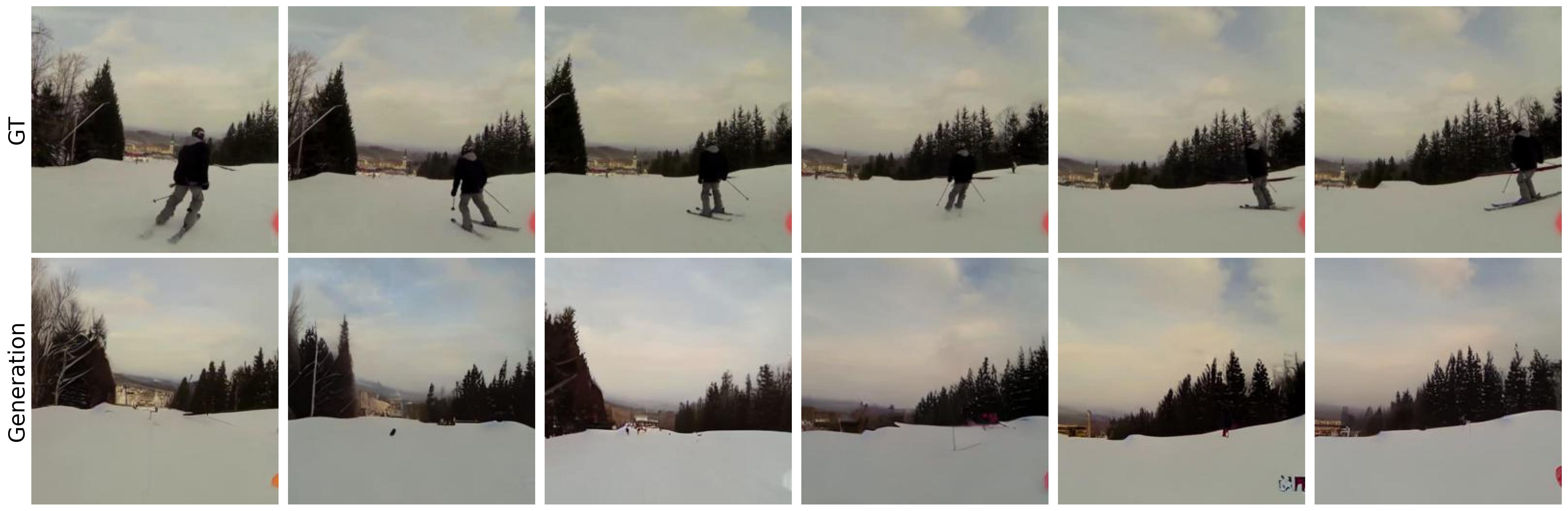}
    \includegraphics[width=0.9\linewidth]{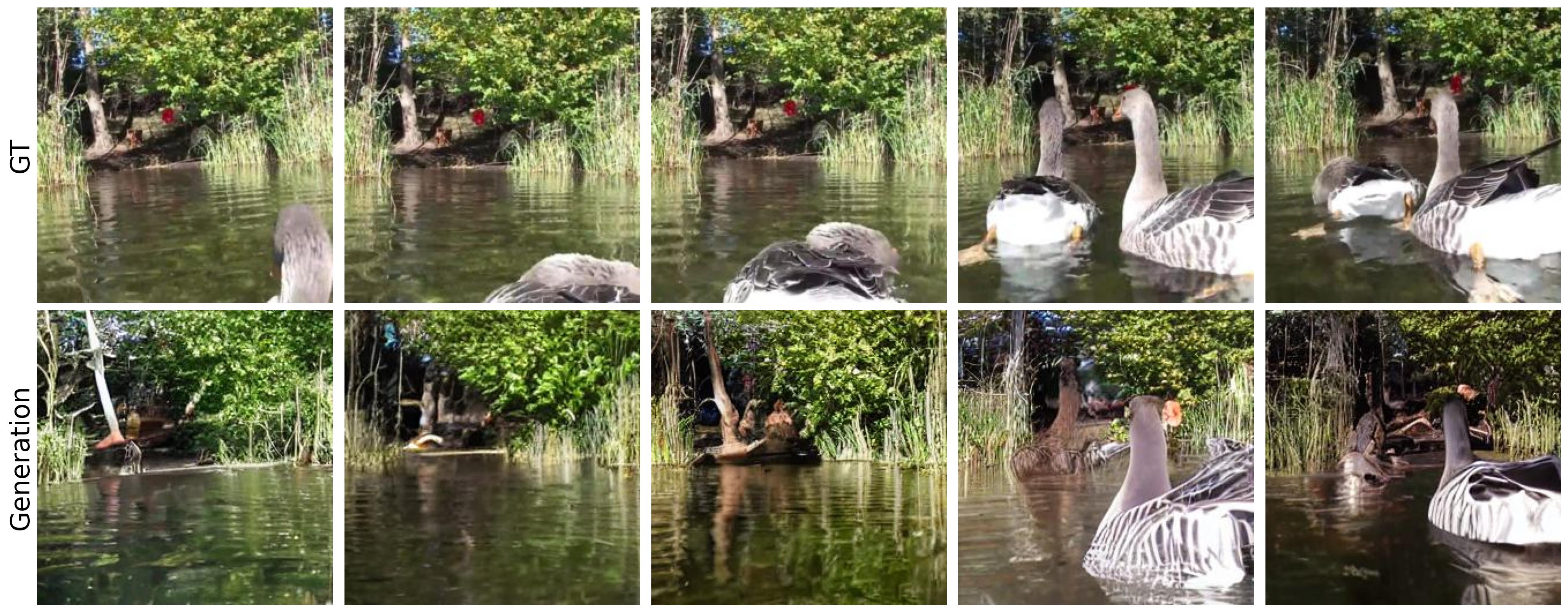}
    \includegraphics[width=0.9\linewidth]{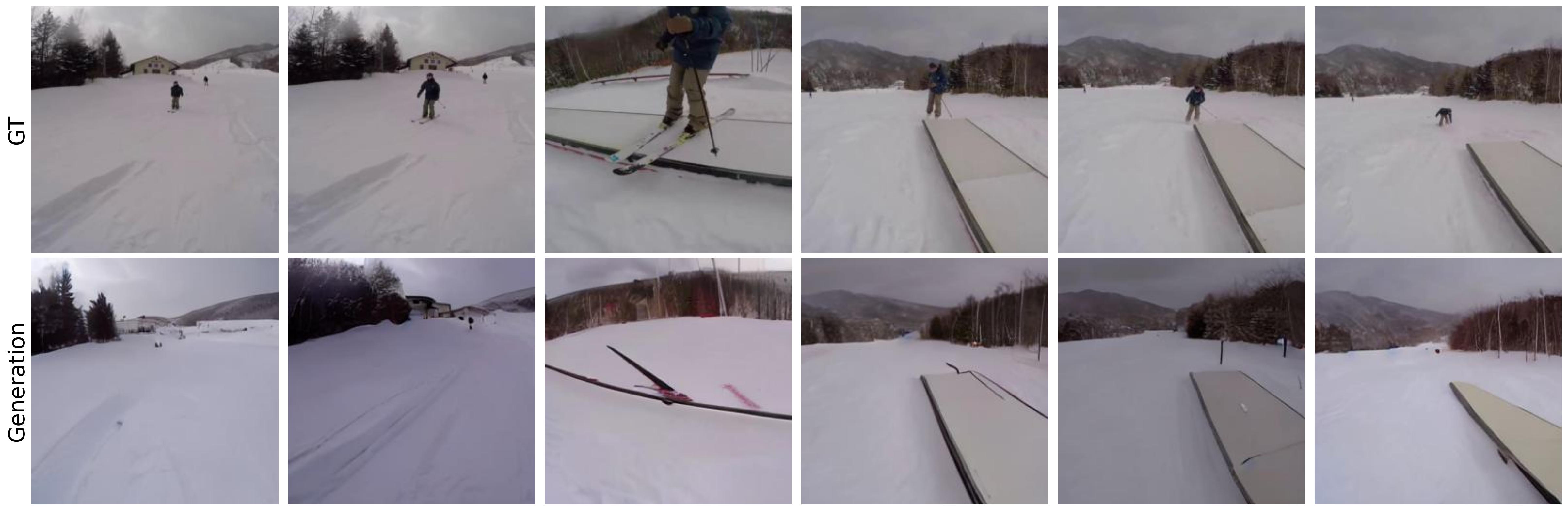}
    \caption{\textbf{Compositional Editing Examples.} Object deletion across diverse video sequences with realistic scene completion.}
    \label{fig:comp-gen-multi-frame-2}
\end{figure*}

\begin{figure*}[htbp]
    \centering
    \includegraphics[width=0.9\linewidth]{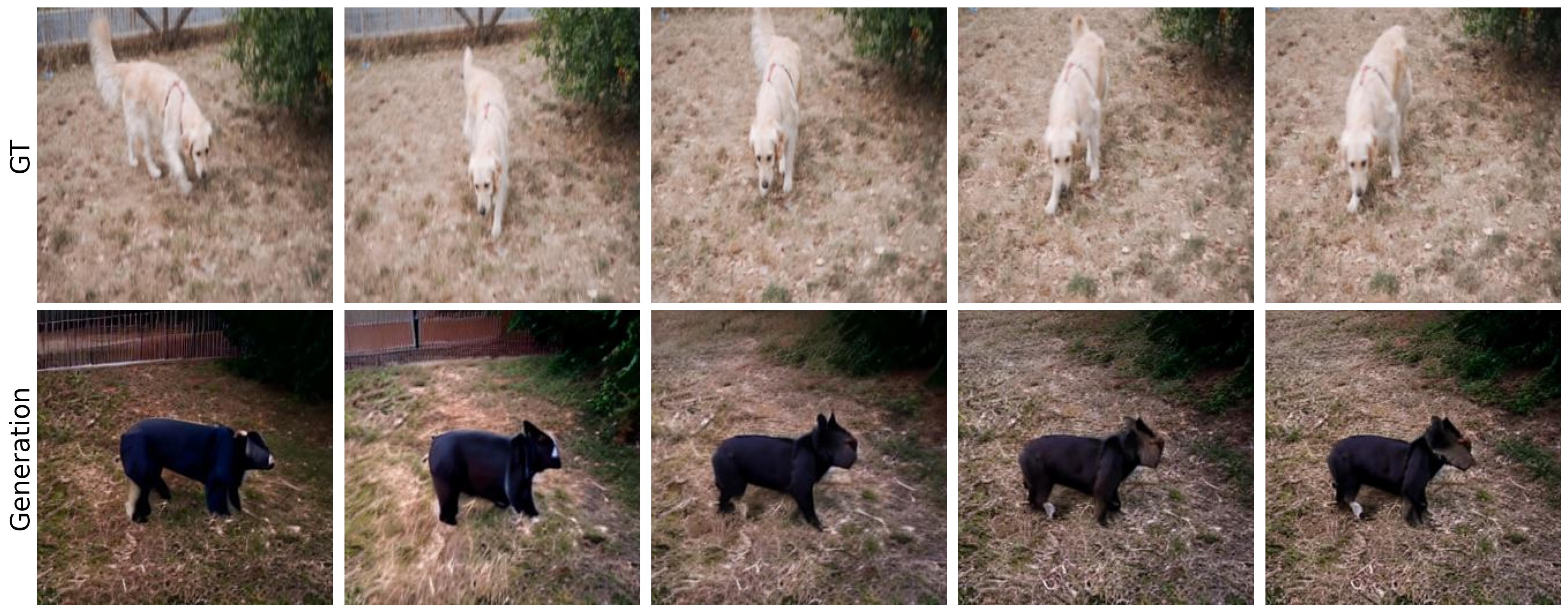}
    \includegraphics[width=0.9\linewidth]{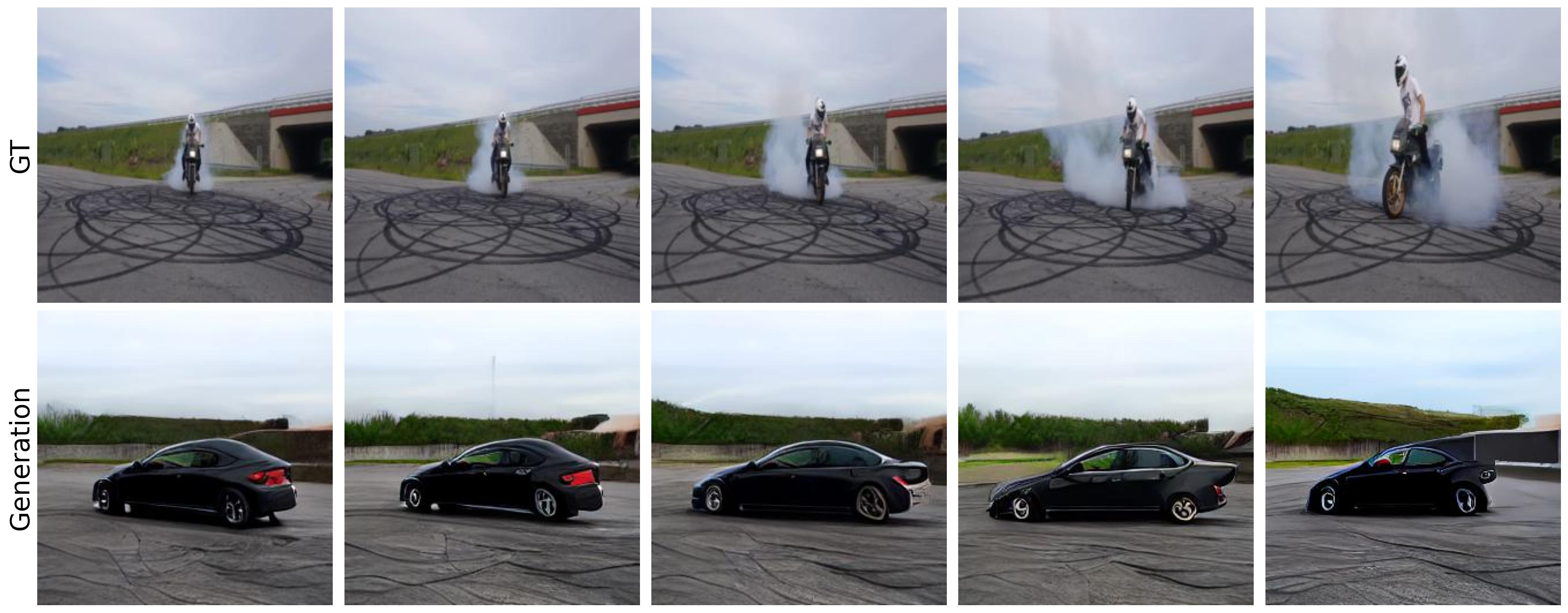}
    \includegraphics[width=0.9\linewidth]{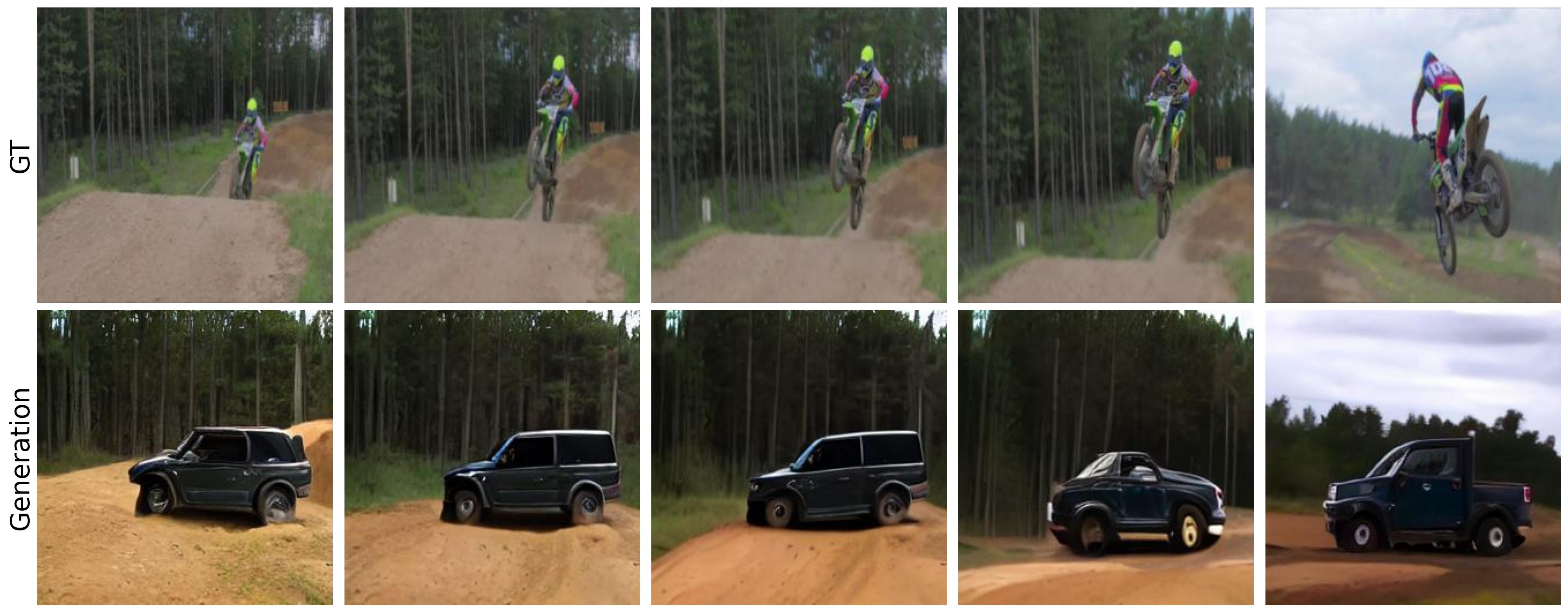}
    \caption{\textbf{Compositional Editing Examples.} Targeted object replacement is performed while preserving scene coherence and temporal consistency. \textit{Top}: the white dog is replaced with a black dog. \textit{Middle} and \textit{bottom}: the motorcycle is replaced with a black car.}    \label{fig:comp-gen-multi-frame-3}
\end{figure*}

Figures~\ref{fig:comp-gen-multi-frame-1}, \ref{fig:comp-gen-multi-frame-2}  and \ref{fig:comp-gen-multi-frame-3} demonstrate our framework's compositional editing capabilities through object deletion and replacement across temporal sequences. Our slot-based representation enables targeted removal or replacement of specific objects while maintaining scene coherence and temporal consistency. The top row (GT) shows original ground truth frames, while the bottom row (Gen) displays generated results after object removal.

The model addresses several technical challenges: (1) background inpainting where objects were removed, (2) preservation of natural motion dynamics in remaining scene elements, (3) maintenance of lighting and shadow consistency, and (4) integration of filled regions with surrounding context. These results show our approach's effectiveness in enabling object-level control for video editing applications, demonstrating the practical utility of our unified slot-based framework for compositional video manipulation tasks.

\end{appendices}

\end{document}